\documentclass[journal]{IEEEtran}
\usepackage{multirow}
\usepackage{makecell}
\usepackage{soul}
\usepackage{color}
\usepackage{booktabs}
\usepackage[export]{adjustbox}

\newcommand{\etal}{\emph{et al.} }

\usepackage{amssymb}

\usepackage{cite}

\ifCLASSINFOpdf
\else
\fi
\usepackage{amsmath}
\usepackage{amsfonts}
\usepackage{mathrsfs}

\usepackage{array}

 \usepackage[caption=false,font=footnotesize]{subfig}
\usepackage{url}

\begin{document}
%
\title{Unified Image Aesthetic Assessment through the Recognition of the Image Style and Composition}
\title{A Deep Multi-Network Approach for Image Aesthetic Assessment}
\title{Composition and Style Attributes Guided Image Aesthetic Assessment}
%
%
%

\author{
\IEEEauthorblockN{Luigi~Celona\IEEEauthorrefmark{1}, Marco~Leonardi\IEEEauthorrefmark{1}\IEEEauthorrefmark{2}, Paolo~Napoletano\IEEEauthorrefmark{1}, and Alessandro~Rozza\IEEEauthorrefmark{2}}\\
\IEEEauthorblockA{\IEEEauthorrefmark{1}Department of Informatics, Systems and Communication, University of Milano-Bicocca, viale Sarca, 336 Milano, Italy}\\
\IEEEauthorblockA{\IEEEauthorrefmark{2}lastminute.com, Corso S. Gottardo, 30 Chiasso, Switzerland}%
\thanks{Corresponding author: Luigi Celona (luigi.celona@unimib.it).}
\thanks{Manuscript received December 1, 2012; revised August 26, 2015.}
}

%
%

\markboth{Journal of \LaTeX\ Class Files,~Vol.~14, No.~8, August~2015}%
{Shell \MakeLowercase{\textit{et al.}}: Bare Demo of IEEEtran.cls for IEEE Journals}
%


\maketitle

\begin{abstract}
The aesthetic quality of an image is defined as the measure or appreciation of the beauty of an image. Aesthetics is inherently a subjective property but there are certain factors that influence it such as, the semantic content of the image, the attributes describing the artistic aspect, the photographic setup used for the shot, etc. In this paper we propose a method for the automatic prediction of the aesthetics of an image that is based on the analysis of the semantic content, the artistic style and the composition of the image. The proposed network includes: a pre-trained network for semantic features extraction (the  Backbone); a Multi Layer Perceptron (MLP) network that relies on the  Backbone features for the prediction of  image attributes (the AttributeNet); a self-adaptive Hypernetwork that  exploits the  attributes  prior  encoded  into  the  embedding  generated by  the  AttributeNet  to  predict  the  parameters  of  the  target network dedicated to aesthetic estimation (the AestheticNet). Given an image, the proposed multi-network is able to predict: style and composition attributes, and aesthetic score distribution. Results on three benchmark datasets demonstrate the effectiveness of the proposed method, while the ablation study gives a better understanding of the proposed network.
\end{abstract}


\IEEEpeerreviewmaketitle

\section{Introduction}
\IEEEPARstart{A}{esthetics} of an image is defined as the measure or appreciation of the beauty of an image. Image aesthetics is a subjective property that depends on the viewer's preferences, experiences, and photography skills. Despite this, the occurrence of specific factors or patterns objectively makes an image more appealing than others. Researchers have in fact found that aesthetics can be influenced by several factors including lighting \cite{freeman2007complete}, color scheme \cite{shamoi2020modeling}, contrast \cite{itten1975design}, composition \cite{liu2020composition}, semantic photo content \cite{celona2021genetic,luo2011content}, and image styles~\cite{obrador2010role,kodak1981how}.

The semantic content of a photo is a key aspect in the evaluation of aesthetic quality: (i) psychology research shows that certain kinds of content are more attractive than others \cite{freeman2017photographer}; (ii) professional photographers choose different photographic techniques and have different aesthetic criteria in mind when shooting different types of contents \cite{datta2006studying,luo2011content}. 

\begin{figure}
  \centering
  \includegraphics[width=.8\columnwidth]{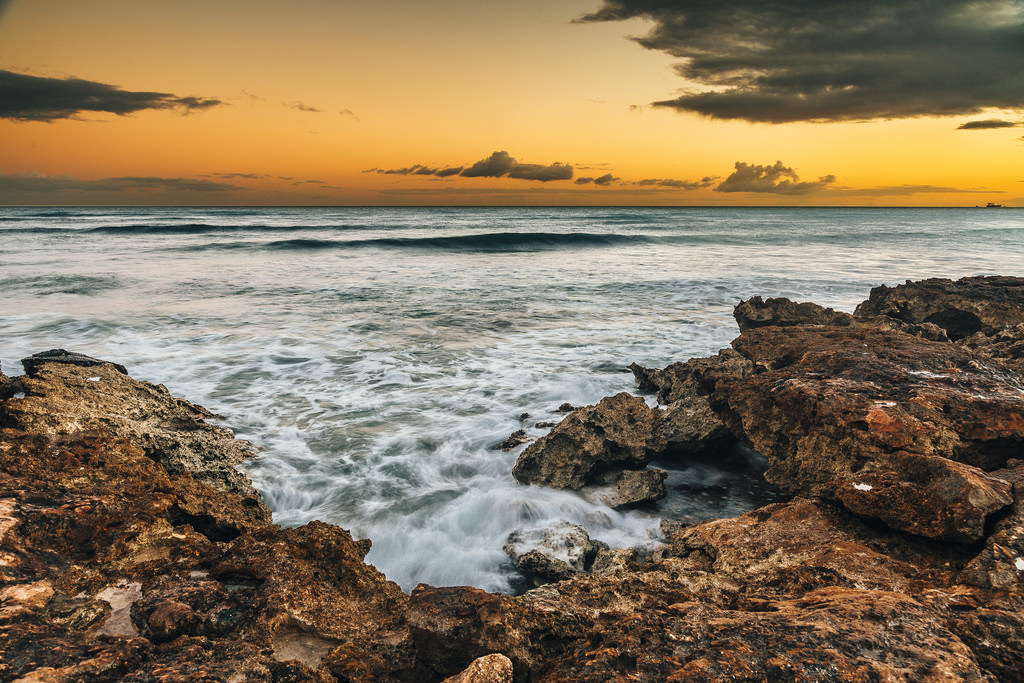} \\ \vspace{1em}
  \includegraphics[width=.8\columnwidth]{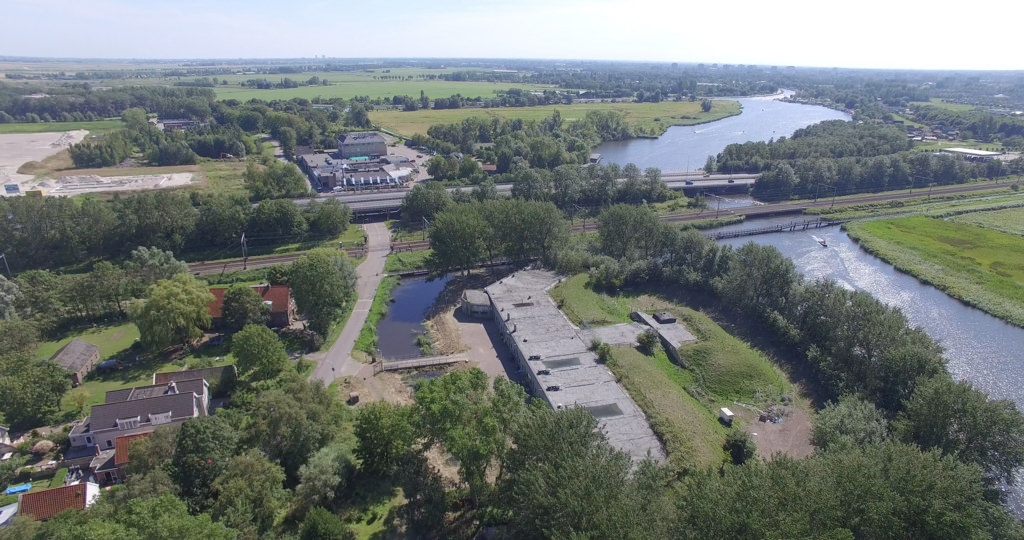}
  \caption{Two images with high and low esthetics from the AADB database \cite{kong2016aesthetics}. The top image has a high aesthetic likely thanks to good lighting and harmonious color combinations, while the image with a low aesthetic has low light and dull colors. 
  }
  \label{fig:teaser}
\end{figure}

In the same way, image styles such as ``Long Exposure'', ``Macro'', ``Bokeh'' and others, or image geometric composition rules such as ``Rule of Thirds'', ``Curved'' and others, influence the aesthetic quality of an image~\cite{obrador2010role,kodak1981how}. Figure \ref{fig:teaser} shows an example of a high-quality image (top)  and a low-quality image (bottom). In this example, the image on the top has been rated on average by a group of humans with high level of aesthetics. This is likely due to nice attributes such as, good lighting and harmonious color combinations, which make the image attractive. In contrast, the image below has a low aesthetic level rate which likely is due to low light and dull colors.

Literature reports mostly on three different aesthetics recognition tasks: high vs. low aesthetic quality~\cite{lu2014rapid,kao2017deep,ma2017lamp,liu2020composition}, aesthetic score regression~\cite{bianco2016predicting,hosu2019effective,shu2020learning,pan2019image}, and aesthetic score distribution prediction \cite{talebi2018nima,zhang2019gated,chen2020adaptive}. Whatever is the recognition task, most of researchers do not explicitly model the aforementioned factors that influence image aesthetics and indeed they prefer holistic approaches~\cite{datta2006studying,luo2011content,marchesotti2011assessing,hosu2019effective}. Besides, methods explicitly modeling aesthetic attributes try to learn a universal model to be applied to images with different aesthetic attributes \cite{pan2019image,shu2020learning,leonardi2021model}. However, images that share the same aesthetic attribute or particular combinations of attributes can also have different levels of aesthetics. It is therefore necessary to learn aesthetic prediction models that specialize for each type of attribute and at the same time are able to consider the correlation between several attributes. 

In this work, we describe a new method that aims to resolve the aforementioned limitations. The proposed method models the aesthetics of the image by explicitly taking into account image semantic content, style and composition.
In particular, we exploit side information related to aesthetic attributes in a way that each attribute is suitably employed to build an \emph{ad-hoc} image aesthetics estimator.

To better exploit side information given by image style and composition attributes, the training stage of the proposed method is multi-stage. The first training stage involves a Multi Layer Perceptron (MLP) network (the AttributeNet) which is specially trained to recognize image style and composition. This network takes as input semantic features extracted by a pre-trained network (the Backbone). The second training stage concerns a hypernetwork (HyperNet).
The latter exploits the attributes prior encoded into the embedding generated by the AttributeNet to predict the parameters of the target network dedicated to aesthetic estimation (AestheticNet). The adoption of the attribute-conditioned hypernetwork, therefore, determines attribute-specific aesthetic estimators. 
The HyperNet is trained using the Earth Mover’s Distance (EMD) as a loss function to better learn the distribution of user judgments attributed to each image. This strategy allows to model the consensus and the diversity of opinions among the annotators and consequently to improve the effectiveness of the proposed method.

Given a test image, the proposed method predicts image style and composition as well as the aesthetic score distribution.

To summarize, the contribution of this work are the following.
\begin{itemize}
  \item We present a deep learning-based method that not only it estimates aesthetics in terms of score distribution but it also determines the style and composition of the input image.
   \item We propose a hypernetwork which adaptively generates the aesthetic quality prediction parameters basing on aesthetic attributes. The proposed method predicts image aesthetics in a content- and attribute- aware manner, therefore it is not limited to a holistic evaluation of the aesthetic quality.
  \item We conduct comprehensive experiments for unified aesthetic prediction tasks: aesthetic classification, aesthetic regression, and aesthetic label distribution. For all of these tasks, the proposed method achieves higher performance than state-of-the-art approaches on three common benchmark datasets (AADB \cite{kong2016aesthetics}, AVA \cite{murray2012ava}, and Photo.net \cite{datta2006studying}).
\end{itemize}

The reminder of this article is organized as follows. In Section \ref{sec:related-works}, related works are summarized. Section \ref{sec:problem} defines the proposed attribute-guided aesthetic assessment method, that is then detailed in Section \ref{sec:method}. Section \ref{sec:datasets} presents the datasets used for aesthetic-related attribute recognition as well as those for aesthetic assessment. Section \ref{sec:experiments} describes the evaluation and the training procedures. Section \ref{sec:results} reports the quantitative evaluation of the proposed method and the comparison with previous methods. Finally, in Section \ref{sec:conclusion} conclusions and future work are drawn.

\section{Related works}
\label{sec:related-works}
In this section, we review  relevant literature related to image aesthetic quality assessment and we highlight the differences between the proposed method and similar existing methods.
\subsection{Image aesthetic quality assessment}
From the seminal work of Datta \etal \cite{datta2006studying} many research efforts have been made, and various methods have been proposed for estimating the aesthetics of images \cite{deng2017image}. Several papers proposed the use of hand-crafted features to encapsulate both aspects of human perception and photographic rules.
For example, Datta \etal \cite{datta2006studying} carefully selected 56 hand-crafted visual features based on standard photographic rules (such as rule of thirds, colorfulness, or saturation) to discriminate between aesthetically pleasing and displeasing images. Luo \etal \cite{luo2011content} extracted features encoding photographic rules, e.g. composition, lighting, and color arrangement, to evaluate aesthetics in different ways based on the photo content. Zhang \etal \cite{zhang2014fusion} modeled image aesthetics by focusing on the image composition which is modeled using graphlets small-sized connected graphs.

However, methods based on hand-crafted features can only achieve limited success~\cite{deng2017image}: (i) hand-crafted features can not exhaustively model the variations of photographic rules between different categories of images; (ii) hand-crafted features are heuristics, and so it is challenging to mathematically model some photographic rules. Based on the previous considerations and thanks to the availability of more labeled data, the trend has shifted from hand-crafted feature-based methods to deep learning methods \cite{bianco2016predicting,hii2017multigap,zhou2016joint}. 

RAPID \cite{lu2014rapid} is a double-column network that captures both local and global information of images for discriminating low and high aesthetics. Given that one patch may not well represent the fine-grained information in the entire image, Lu \etal \cite{lu2015deep} extended RAPID by proposing Deep Multi-patch Aggregation Network (DMA-Net). In DMA-Net, an input image is represented by a bag of random cropped patches. The proposed layers, namely the statistics and sorting layers, enabled the integration of multiple input patches. Given that DMA-Net failed to encode the global layout of the image, Ma \etal presented the Adaptive Layout-Aware Multi-Patch Convolutional Neural Network (A-Lamp CNN) \cite{ma2017lamp}. This method is able to accept arbitrary sized images and learn from both fined grained details and holistic image layout simultaneously. It consists of two subnets, i.e. a Multi-Patch subnet which is very similar to DMA-Net and a Layout-Aware subnet consisting of an object-based attribute graph. Multi-Net Adaptive Spatial Pooling ConvNet (MNA-CNN) \cite{mai2016composition} is trained and tested on images at their original sizes and aspect ratios. It computed aesthetics by combining multi-level features and scene information. Chen \etal \cite{chen2020adaptive} designed the Adaptive Fractional Dilated Convolution (AFDC) that, similarly to MNA-CNN, avoids altering original image aspect ratio and composition. RGNet \cite{liu2020composition} builds a region graph to represent the visual elements and their spatial layout in the image, and then performs reasoning on the graph to uncover the mutual dependencies of the local regions. Gated Peripheral-Foveal Convolutional Neural Network (GPF-CNN) \cite{zhang2019gated} is a deep architecture designed to: encode the global image composition; extract the fine-grained details from aesthetic-relevant regions.

Professional photographers adopt different photographic techniques and have various aesthetic criteria depending on the portrait content. Therefore, Kao \etal \cite{kao2017deep} proposed a Multi-task Convolutional Neural Network (MTCNN). This model aims to simultaneously estimate the semantic content and the aesthetic class of an image.

Neural IMage Assessment (NIMA) \cite{talebi2018nima} replaced the classification layer of a pre-trained ImageNet CNN with a fully-connected regression head that predicts the distributions of ratings per image. The squared Earth Mover's Distance (EMD) has been employed as the loss function. In this paper, similar with \cite{talebi2018nima}, we optimize our network for aesthetic score distribution by minimizing EMD loss. Multi-level Spatially Pooled activation blocks (MLSP) \cite{hosu2019effective} exploited a transfer learning strategy that uses features extracted from a pre-trained ImageNet CNN.

Some recent works regard multi-modal aesthetic evaluation models that leverage visual information along with user comments. The latter encodes high-level semantic information and are relevant for aesthetic decisions \cite{zhou2016joint,hii2017multigap,zhang2021mscan}.
\subsection{Correlation with existing methods}
There are several state-of-the-art methods similar to the one proposed in this article as they use multiple attributes describing the aesthetic or artistic aspect of a photo for aesthetic evaluation \cite{kong2016aesthetics,leonardi2021model,gao2020style,shu2020learning,pan2019image,lee2019property}.

Leonardi \etal \cite{leonardi2021model} and Gao \etal \cite{gao2020style} used attributes as mid-level representation to estimate the aesthetics of the image. Thus, the errors on the attributes might be propagated to the assessed aesthetics. Likewise, Lee \etal \cite{lee2019property} presented the Property-Specific Aesthetic Assessment (PSAA) algorithm. The PSAA algorithm first classifies an image into a specific aesthetic property and then discriminates the image into a high or low aesthetic quality through a specific-property network. Our work differs from the PSAA in three main aspects. First, PSAA discriminates high or low aesthetics while we face the more challenging task of estimating the distribution of aesthetic scores. Second, the aesthetic categorization of PSAA depends exclusively on the aesthetic classifier for the specific property with the highest score, while our aesthetic estimate relies on the representation of the attributes and their interactions. Third, PSAA’s attribute estimation errors might be propagated to the assessed aesthetic, similarly as in the methods of Leonardi \etal and Gao \emph{et al}.
Pan \etal \cite{pan2019image} proposed a multi-task deep network to learn the aesthetic score and aesthetic attributes simultaneously. Attributes were used as additional information for the learning paradigm called Learning Using Privileged Information (LUPI) \cite{vapnik2009new}. Besides, adversarial learning was introduced to capture the correlation between the aesthetic score and attributes. Shu \etal \cite{shu2020learning} also exploited LUPI by proposing Deep Convolutional Neural Network with Privileged Information (PI-DCNN): a novel method exploring photo attributes as privileged information for photo aesthetic assessment.

There are three major differences between the previous methods and the one proposed in this article which are summarized a follows:

\begin{itemize}
  \item First, previous methods are limited to the datasets annotated with aesthetic attributes, namely AVA or AADB. In contrast, in the proposed method, side information about composition and style is learned from specially designed datasets. The proposed method can therefore generalize to a larger number of aesthetic attributes.
  \item Second, the previous methods learn a universal model of aesthetics that depends indiscriminately on the aesthetic attributes. In contrast, the proposed method learns aesthetic models that are dependent on the different aesthetic attributes present within the image and their correlation.
  \item Finally, we use Earth Mover's Distance (EMD) as a loss function to better learn the distribution of user judgments attributed to each image. As demonstrated in previous works \cite{talebi2018nima,chen2020adaptive}, this strategy allows to model the consensus and the diversity of opinions among the annotators and consequently to improve the effectiveness of the proposed method.
\end{itemize}

\section{Proposed method}
\label{sec:problem}
In the following we introduce the mathematical formulation of the proposed method. Given an input image $\mathbf{X}$, the goal of the proposed method is to estimate both the aesthetic score distribution $\mathbf{\hat{q}}$ and the presence of a set of aesthetic-related attributes $\mathbf{\hat{y}}$ by using the network $f$ parametrized with $\theta^*$:
\begin{equation}
    \mathbf{X} \xrightarrow{f(\theta^{*})} (\mathbf{\hat{q}},\hat{\mathbf{y}}).
\end{equation}

More specifically, $f \leftarrow (f_{s},f_{t})$ consists of two networks. The network $f_{s}$ handles the \textit{side} information regarding the aesthetic-related attributes and produces the final output $\mathbf{\hat{y}}$ and the embedding $\mathbf{e}_s$. The latter is exploited by an \textit{attribute-conditioned} hypernetwork $\hat{\theta}_t = h(\mathbf{e}_s; \theta^*_h)$ that adaptively generate the parameters $\hat{\theta}_t$ of the network $f_{t}$. Such a network $f_{t}$ carries out the \textit{main} task of aesthetic assessment. Hence, by using the attribute-conditioned hypernetwork we subordinate the aesthetic assessment task to that of attribute estimation.

$\theta^{*} \leftarrow (\theta^{*}_{h},\theta^{*}_{s},\theta^{*}_{b})$ is the set of learned parameters of the proposed method. Unlike the $\theta^{*}_{b}$ parameters, which belong to a pre-trained backbone, the others are learned for our tasks. Similar to \cite{benjamin2018measuring}, we adopt a two-step optimization procedure to introduce attribute-constraint into the hypernetwork.

The first step regards the training of the parameters $\theta_s$ of the network $f_s$. Let $\mathcal{D}_{s} = \{(\mathbf{X}_s^{(i)},\mathbf{y}^{(i)})\}_{i=1}^N$ denotes the training set of $N$ training samples. Each training sample consists of a color image $\mathbf{X}_s^{(i)} \in \mathbb{R}^d$ and $K$ aesthetic-related attributes $\mathbf{y}^{(i)} \in \mathbb{R}^K$. Given the training set $\mathcal{D}_{s}$, our goal is to learn a network $f_{s}:\mathbb{R}^d \rightarrow \mathbb{R}^K$, which predicts whether an attribute occurs or not into the input image:
\begin{equation}
    \mathcal{L}_{side}\{f_{s}(\mathbf{e}_{b};\theta_{s},\mathbf{y})\mid \theta_{s}\in \Theta_{s}\},
    \label{eq:side_net}
\end{equation}
where $\theta_s \in \Theta_{s}$ are the learnable real-valued parameters and $\mathbf{e}_{b} = b(\mathbf{X}_{s};\theta^{*(L)}_{b})$ is the embedding corresponding to the activations of layer $L$ of the pre-trained backbone $b$ given the input $\mathbf{X}_{s}$.

The second training concerns the hypernetwork. Instead of directly learning the $\theta_{t}$ parameters of the network $f_{t}$, the hypernetwork is trained to learn the parameters $\theta_h$ of a metamodel $h$. The output of this metamodel is $\hat{\theta}_{t}$. The network $h$ can therefore be thought of as a generator of parameters to obtain attribute-specific aesthetic estimators. Let $\mathcal{D}_{t} = {(\mathbf{X}_t^{(i)},\mathbf{q}^{(i)})}_{i=1}^N$ denotes the training set of $N$ training samples. Each training sample consists of a color image $\mathbf{X}_t^{(i)} \in \mathbb{R}^d$ and a distribution of aesthetics ratings $\mathbf{q}^{(i)}=[q_{s_1},q_{s_2},...,q_{s_B}]$. Where ${s_j}$ is the $j$-th score bucket, $B$ is the total number of score buckets, and $q_{s_j}$ denotes the number of voters that give the discrete score $s_j$ to the image. Given the training set $\mathcal{D}_t$, our goal is to learn the parameters $\theta_h$ of the metamodel to generate the parameters for the network $f_t:\mathbb{R}^d \rightarrow \mathbb{R}^B$, which predicts the aesthetic score distribution $\hat{\mathbf{q}}$:

\begin{equation}
    \mathcal{L}_{task}\{f_{t}(\mathbf{e}_{b};h(\mathbf{e}_{s};\theta_{h}),\mathbf{q}) \mid \theta_{h}\in \Theta_{h}\},
\end{equation}

where $\mathbf{e}_{b}$ is the same embedding as in Eq. \ref{eq:side_net}, $\mathbf{e}_{s} =f_s(\mathbf{e}_{b};\theta_{s}^{*(M)})$ is the attribute-conditioned embedding obtained from the $M$-th layer of the pre-trained $f_s$ given the input $\mathbf{e}_{b}$, and $\theta_{h}\in \Theta_{h}$ are the learnable real-valued parameters.

\begin{figure*}[tb]
  \centering
  \includegraphics[width=2\columnwidth]{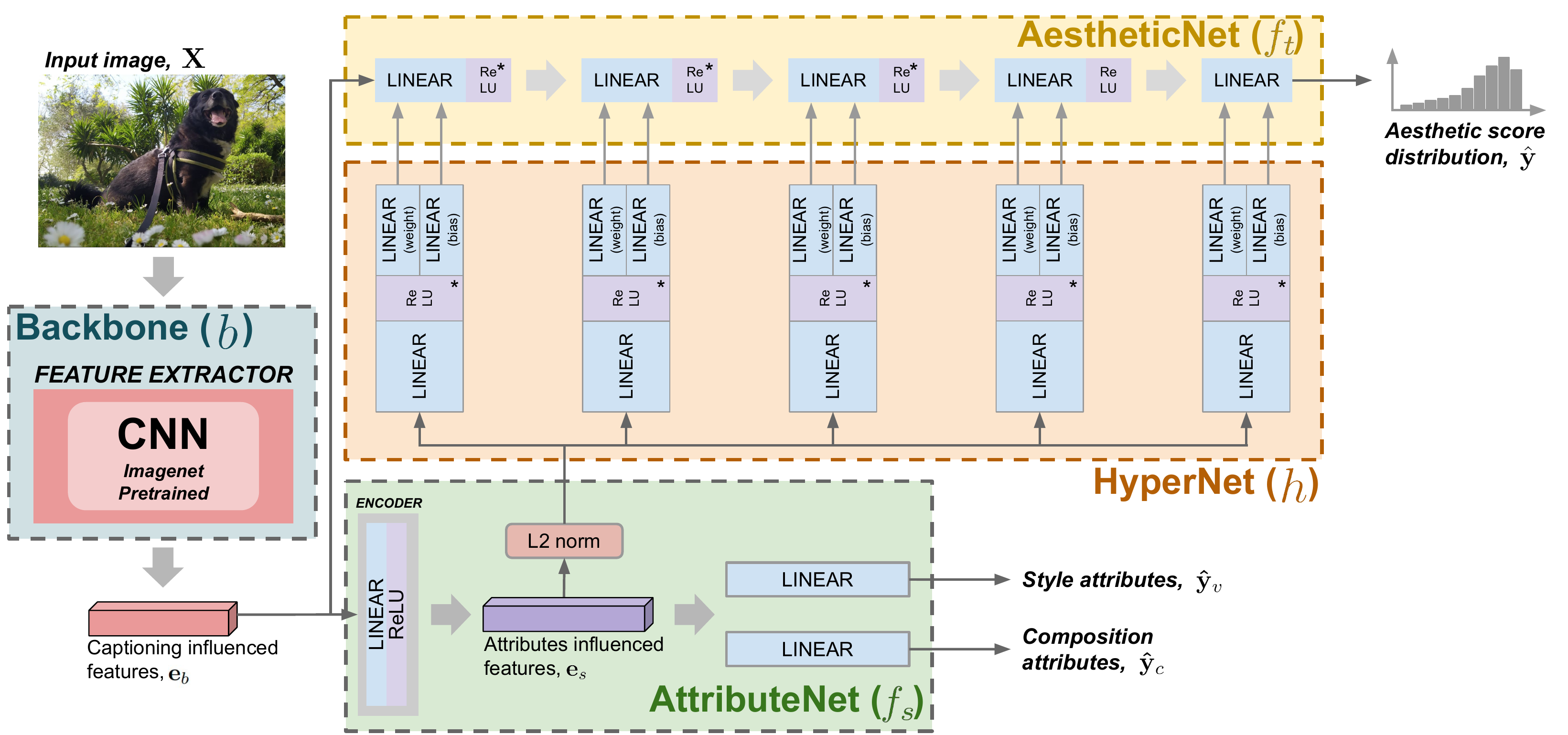}
  \caption{The proposed method is composed of four main parts: the Backbone ($b$), the AttributeNet ($f_s$), the HyperNet ($h$) and the AestheticNet ($f_t$). The input image $\mathbf{X}$ is first fed to the Backbone to extract a feature set ($\mathbf{e}_b$) that encodes the content of the image. Then, this feature set is fed to AttributeNet. The goal of the AttributeNet is to predict aesthetics-related attributes (i.e. $\hat{\mathbf{y}}_v$ and $\hat{\mathbf{y}}_c$) and influence the input of the HyperNet. The HyperNet aims to predict the weights and the biases of the AestheticNet. Finally, the AestheticNet infers the aesthetic score distribution, $\hat{\mathbf{y}}$, of the input image over the content related feature set with the weights and the biases predicted by the HyperNet. *Trained with dropout}
  \label{fig:HyperNetwork}
\end{figure*}

\section{Proposed network architectures}
\label{sec:method}

The proposed architecture includes four different networks trained using a multi-stage approach: the Backbone, the AttributeNet, the HyperNet and the AestheticNet. The overall architecture of the model is shown in Figure \ref{fig:HyperNetwork}.

The Backbone is an ImageNet \cite{deng2009imagenet} pre-trained neural network that outputs multi-level features used as inputs of the AestheticNet and the AttributeNet. The latter is a Multi Layer Perceptron (MLP) network specially trained for image style and composition recognition. The features obtained from the previously trained AttributeNet are then used as HyperNet inputs. It is a metamodel dedicated to calculating the weights and distortions of the AestheticNet. Therefore, the HyperNet is trained to allow the AestheticNet to predict aesthetics scores for input images in close agreement with human judgments.

The following sections detail each of the aforementioned components of the proposed method.
\subsection{Backbone}
Many of the earlier approaches to image aesthetics assessment extract features from warped and/or cropped input images \cite{bianco2016predicting,talebi2018nima,shu2020learning}. A shortcoming of these methods is that they alter the composition of the image and the aspect ratio of the objects. Thus, they may harm the task of aesthetics assessment.

On the other hand, previous works have shown the effectiveness of multi-level features to predict perceptual judgements, either for image quality assessment \cite{lin2020deepfl} or image aesthetics assessment \cite{hosu2019effective,reddy2020measuring,leonardi2021model}.

For the previous reasons, the Backbone network $b : \mathbf{X} \rightarrow \mathbf{e}_b$ encodes an input image, at the original size, $\mathbf{X} \in \mathbb{R}^d$ into a Multi-Level Spatially Pooled (MLSP) embedding vector $\mathbf{e}_b \in \mathbb{R}^D$. The resulting embedding vector encodes information at multiple levels of abstraction: from low- to high-level features. This goal is achieved by stacking activations from $L$ layers of a given pre-trained CNN. As the spatial resolution of the different activation maps varies, a Global Average Pooling (GAP) is adopted to squeeze the spatial dimensions into a channel activation vector. Therefore, the size of the MLSP embedding vector depends solely on the number of channels in each layer. This last aspect allows processing images at full resolution without the need to resize or crop them.

To summarize, the Backbone network $b(\mathbf{X};\theta^{*(L)}_{b})$ outputs the embedding $\mathbf{e}_{b}$ corresponding to the activations of $L$ layers given the input $\mathbf{X}$.

We exploit an ImageNet-pretrained EfficientNet-B4 \cite{tan2019efficientnet} as Backbone network $b$: a very efficient yet effective model not only on ImageNet but also on transfer learning datasets. Recently, it demonstrates its effectiveness for image aesthetic assessment \cite{reddy2020measuring,leonardi2021model}. Following \cite{reddy2020measuring}, the activations of the MBConv blocks \cite{tan2019efficientnet} having numbers $L = \{15, 21, 25, 29, 31\}$ are considered. Given the input image $\mathbf{X}$ with shape $ h \times w \times 3$, the resulting activation maps have the following shapes: $\frac{h}{16} \times \frac{w}{16} \times 112$, $\frac{h}{16} \times \frac{w}{16} \times 160$, $\frac{h}{32} \times \frac{w}{32} \times 272$, $\frac{h}{32} \times \frac{w}{32} \times 272$, and $\frac{h}{32} \times \frac{w}{32} \times 448$. The previous 5 activation maps are spatially narrowed using the GAP and stacked on the channel dimension, thus obtaining a fixed sized narrow MLSP embedding vector of shape $1\times 1\times 1264$.
\subsection{AttributeNet}
The AttributeNet $f_s : \mathbf{e}_b \rightarrow \mathbf{\hat{y}}$ with $\mathbf{\hat{y}} \in \mathbb{R}^K$ is a Multi Layer Perceptron (MLP) that aims to categorize the backbone embedding $\mathbf{e}_b$ with respect to $K$ aesthetic-related attributes. More specifically, $\mathbf{\hat{y}} \leftarrow (\mathbf{\hat{y}}_v, \mathbf{\hat{y}}_c)$, where $\mathbf{\hat{y}}_v \in \mathbb{R}^{K_v}$ is the set of image styles and $\mathbf{\hat{y}}_c \in \mathbb{R}^{K_c}$ is the set of composition rules. Therefore, the MLP performs two tasks simultaneously and consists of three linear blocks. The first block is a linear layer with ReLU that transforms the embedding vector $\mathbf{e}_b$ into an embedding vector, $\mathbf{e}_s$:
\begin{equation}
    \mathbf{e}_{s} = \mathrm{ReLU}(\mathbf{W}_s^\top\mathbf{e}_b + \mathbf{b}_s).
\end{equation}
Given that $\mathbf{e}_{s}$ is shared between the two tasks, it intrinsically encodes the style and composition as well as their relationships. The second and third blocks are independent linear layers categorizing the input image into style and composition:
\begin{align}
    \mathbf{\hat{y}}_v &= \mathbf{W}_v^\top\mathbf{e}_s + \mathbf{b}_v, \\
    \mathbf{\hat{y}}_c &= \mathbf{W}_c^\top\mathbf{e}_s + \mathbf{b}_c,
\end{align}
where $\mathbf{W}_v$ and $\mathbf{b}_v$ are the parameters for predicting the style $\mathbf{\hat{y}}_v$, and $\mathbf{W}_c$ and $\mathbf{b}_c$ are the parameters for predicting the composition $\mathbf{\hat{y}}_c$.
\subsection{AestheticNet}
The AestheticNet $f_t(\mathbf{e}_b;\hat{\theta}_t)$ aims to predict the aesthetic score distribution $\hat{\mathbf{q}}$ given the embedding vector $\mathbf{e}_b$ produced by the Backbone. It is a MLP composed of $M$ linear layers whose parameters 
$\hat{\theta}_t = \{(\hat{\mathbf{W}}_1,\hat{\mathbf{b}}_1),(\hat{\mathbf{W}}_2, \hat{\mathbf{b}}_2),...,(\hat{\mathbf{W}}_M, \hat{\mathbf{b}}_M)\}$ are computed by the HyperNet $h$:
\begin{align}
    \mathbf{x}_1 &= \mathrm{ReLU}(\hat{\mathbf{W}}^\top_1\mathbf{e}_b + \hat{\mathbf{b}}_1), \\
    \mathbf{x}_i &= \mathrm{ReLU}(\hat{\mathbf{W}}^\top_i\mathbf{x}_{i-1} + \hat{\mathbf{b}}_i), \quad \mathrm{with} \quad i=2,...,M-1\\
    \hat{\mathbf{q}} &= \hat{\mathbf{W}}^\top_M\mathbf{x}_{i-1} + \hat{\mathbf{b}}_M.
\end{align}

Since the output of a linear layer is of the same size of the input, thus in correspondence of the following input $\mathbf{x}_{in} \in \mathbb{R}^{N_{in}}$ we have the following output $\mathbf{x}_{out} \in \mathbb{R}^{N_{out}}$. As a consequence, the weights of the layer corresponds to $\mathbf{W} \in \mathbb{R}^{N_{in} \times N_{out}}$ and the bias are equal to $\mathbf{b} \in \mathbb{R}^{N_{out}}$. The number of parameters in a linear layer is $N_{in}N_{out}$ for $\mathbf{w}$ and $N_{out}$ for $\mathbf{b}$.
\subsection{HyperNet}
The HyperNet $h(\mathbf{e}_s;\theta^*_h)$ is the metamodel  that generates the parameters $\hat{\mathbf{\theta}}_t$ for the $M$ layers of the AestheticNet. Therefore, the HyperNet is composed of $M$ HyperNet Blocks (HBs), $h = \{HB_1, HB_2,...,HB_M\}$.

Each HB is composed of a linear layer that reduces the size of the \textit{l2}-normalized embedding $\mathbf{e}_s \in \mathbb{R}^D$ to a size of $d \mid d \ll D$:
\begin{equation}
    \mathbf{e}_r = \mathrm{ReLU}(\mathbf{W}_i^{r\top}\mathbf{e}_s + \mathbf{b}^r_i),
\end{equation}
where $\mathrm{ReLU}$ is the activation function, $\mathbf{W}_i^{r} \in \mathbb{R}^{D\times d}$ are the learned weights, and $\mathbf{b}^r_i \in \mathbb{R}^d$ are the learned bias. Two linear layers are then dedicated to the estimation of the parameters, $\mathbf{W}_i$ and $\mathbf{b}_i$ of the $i$-th layer of the AestheticNet:
\begin{align}
    \hat{\mathbf{W}}_i &= \mathbf{W}_i^{w\top}\mathbf{e}_r + \mathbf{b}_i^w, \\
    \hat{\mathbf{b}}_i &= \mathbf{W}_i^{b\top}\mathbf{e}_r + \mathbf{b}_i^b.
\end{align}
Given the generated weights $\hat{\mathbf{W}}_i \in \mathbb{R}^{N_{in}\times N_{out}}$, the learned parameters of the linear layer are $\mathbf{W}_i^{w} \in \mathbb{R}^{d \times N_{in}N_{out}}$ and $\mathbf{b}_i^{w} \in \mathbb{R}^{N_{in}N_{out}}$, respectively. Instead, for the generated bias $\hat{\mathbf{b}}_i \in \mathbb{R}^{N_{out}}$, the learned parameters of the linear layer are $\mathbf{W}_i^{b} \in \mathbb{R}^{d \times N_{out}}$ and $\mathbf{b}_i^{b} \in \mathbb{R}^{N_{out}}$.

\section{Datasets}
\label{sec:datasets}
\subsection{Datasets for aesthetic-related attribute  recognition}
Most previous methods that exploit the relationship between attributes and aesthetic quality rely on the use of the aesthetic attributes provided for the AVA and AADB datasets \cite{gao2020style,shu2020learning}. Although the two sets of attributes are valuable as they span the traditional photographic principles of color, lighting, focus, and composition, they have some drawbacks. First, they are not exchangeable or can be merged because their annotation are different: in the AVA dataset, annotations are binary values that indicate the occurrence of the attribute in the image; for AADB, the annotation of each attribute can assume continuous values between -1 and 1, where ``positive'' and ``negative'' indicates that the occurrence of the attribute improves or degrades the aesthetic level of the image, respectively. Second, only a subset of the AVA images (about 4.44\%) provides the style annotations. Third, the number of attributes in the two sets is limited. Few attributes categorize images for composition style (e.g., Rule of Thirds and symmetry). Although the role of emotion in the aesthetic experience is proven \cite{cupchik1995emotion}, no attributes are specifying what emotions the image content conveys.

Based on the above considerations, and experimental results (see Section \ref{sec:ablation-data}), the set of aesthetic attributes provided by AADB and AVA is not exploited for the training of the proposed method. Instead, a different and broader set of attributes is considered in this work. Attributes were chosen by taking into account not only the composition but also the style of an image. Among these, there are the optical techniques used during the shot (such as bokeh effect and depth-of-Field), the genre of the image content (e.g., horror or romantic), the atmospheric light conditions (such as hazy or sunny), finally, the mood aroused by the image (e.g., serene).

In this paper we use the KU-PCP \cite{lee2018photographic} and FlickrStyle \cite{karayev2014recognizing} dataset. The set $\mathcal{D}_{t}$ is used to train the AttributeNet and it is composed by the two datasets mentioned above. Table \ref{tab:attributes-list} compares the lists of attributes present in the AADB, the AVA datasets and the list of the 29 attributes taken from the KU-PCP and FlickrStyle datasets. We highlight that the set of selected attributes is twice as large as those of AADB (11 attributes) and AVA (14 attributes). Most of the attributes present in the previous sets are also available in the proposed list. We lack attributes such as silhouettes, negative photo, or light on white which are more related to the type of content and not to style or composition.
%
\begin{table}
  \centering
  \caption{Image attributes available for AADB, AVA, and in our selection.}
  \label{tab:attributes-list}
  \begin{tabular}{lccc}
  \toprule
    Attribute Name & AADB \cite{lee2018photographic} & AVA & FlickrStyle + KU-PCP \\ \midrule
    Balancing elements   & \checkmark & & \\
    Bright         & & & \checkmark \\
    Bokeh          & & & \checkmark \\
    Center         & & & \checkmark \\
    Color harmony      & \checkmark & & \\
    Complementary      & & \checkmark & \\
    Content         & \checkmark & & \\
    Curved         & & & \checkmark \\
    Depth-of-Field     & \checkmark & \checkmark & \checkmark \\
    Detailed        & & & \checkmark \\
    Diagonal        & & & \checkmark \\
    Duotones        & & \checkmark & \\
    Ethereal        & & & \checkmark \\
    Geometric        & & & \checkmark \\
    Hazy          & & & \checkmark \\
    HDR           & & \checkmark & \checkmark \\
    Horizontal       & & & \checkmark \\
    Horror         & & & \checkmark \\
    Light          & \checkmark & & \\
    Light on White     & & \checkmark & \\
    Long Exposure      & & \checkmark & \checkmark \\
    Macro          & & \checkmark & \checkmark \\
    Melancholy       & & & \checkmark \\
    Minimal         & & & \checkmark \\
    Motion Blur       & \checkmark & \checkmark & \\
    Negative Photo     & & \checkmark & \\
    Noir          & & & \checkmark \\
    Object         & \checkmark & & \\
    Pastel         & & & \checkmark \\
    Photo Grain       & & \checkmark & \\
    Pattern/Repetition       & \checkmark & & \checkmark \\
    Romantic        & & & \checkmark \\
    Rule of Thirds     & \checkmark & \checkmark & \checkmark \\
    Serene         & & & \checkmark \\
    Silhouettes       & & \checkmark & \\
    Soft Focus       & & \checkmark & \\
    Sunny          & & & \checkmark \\
    Symmetry        & \checkmark & & \checkmark \\
    Texture         & & & \checkmark \\
    Triangle        & & & \checkmark \\
    Vanishing Point     & & \checkmark & \\
    Vertical        & & & \checkmark \\
    Vintage         & & & \checkmark \\
    Vivid Color       & \checkmark & & \\ \bottomrule
  \end{tabular}
\end{table}

\textbf{FlickrStyle:} The FlickrStyle dataset \cite{karayev2014recognizing} is a collection of 80,000 photographs gathered from the Flickr website annotated with 20 curated style labels. These can be categorized into:
\begin{itemize}
  \item \textit{Atmosphere:} Hazy, Sunny
  \item \textit{Color:} Bright, Pastel
  \item \textit{Composition styles:} Detailed, Geometric, Minimal, Texture
  \item \textit{Genre:} Horror, Noir, Romantic, Vintage
  \item \textit{Mood:} Ethereal, Melancholy, Serene
  \item \textit{Optical techniques:} Bokeh, Depth-of-Field, HDR, Long Exposure, Macro
\end{itemize}
\begin{figure*}
\setlength{\tabcolsep}{2pt}
  \centering
  {\footnotesize
  \subfloat[]{\label{fig:flickr-samples}
    \begin{tabular}{ccccccccc}
      Bokeh & DoF & Hazy & HDR & Long Exposure & Macro & Minimal & Noir & Pastel \\
      \includegraphics[width=.1\textwidth,valign=c]{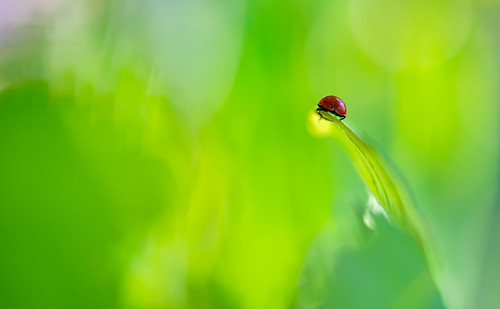} & \includegraphics[valign=c,width=.1\textwidth]{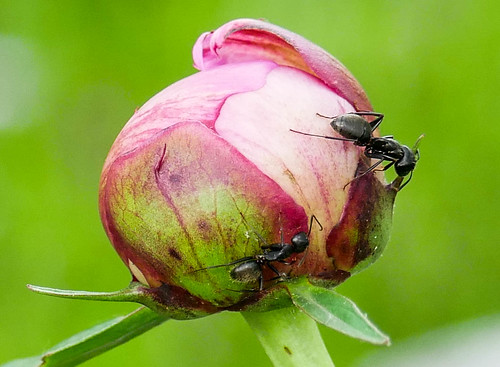} & \includegraphics[valign=c,width=.1\textwidth]{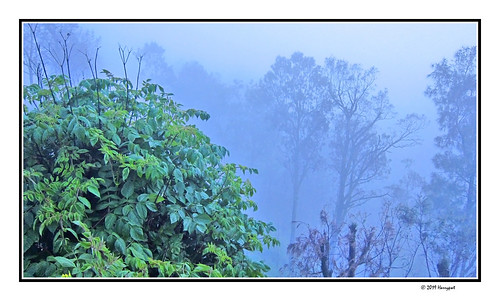} & \includegraphics[valign=c,width=.1\textwidth]{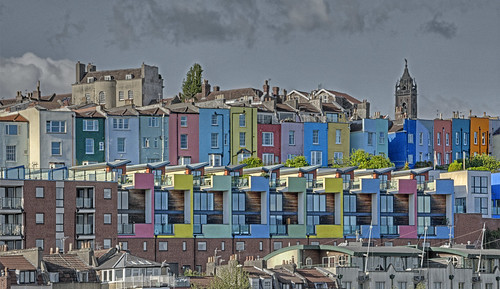} & \includegraphics[valign=c,width=.1\textwidth]{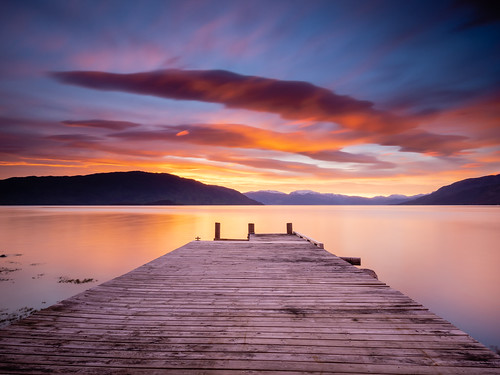} & \includegraphics[valign=c,width=.1\textwidth]{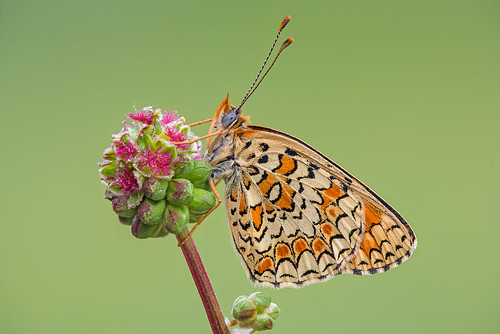} & \includegraphics[valign=c,width=.1\textwidth]{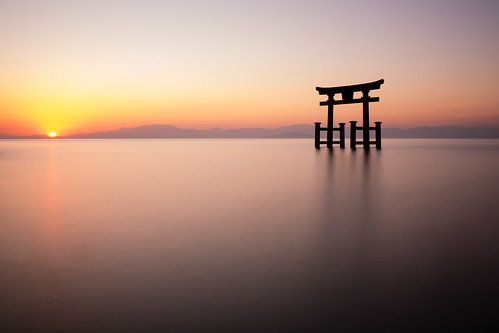} & \includegraphics[valign=c,width=.1\textwidth]{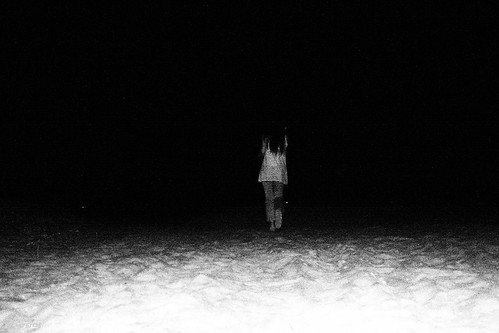} & \includegraphics[valign=c,width=.1\textwidth]{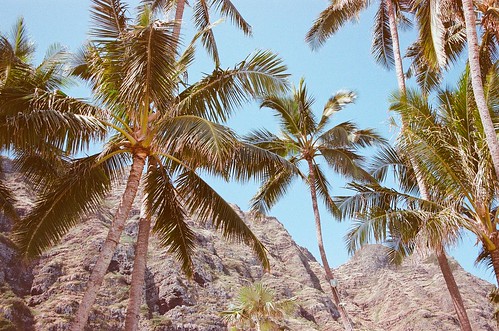}
    \end{tabular}
  }\\
  \subfloat[]{\label{fig:kupcp-samples}
    \begin{tabular}{ccccccccc}
    Center & Curved & Diagonal & Horizontal & Pattern & RoT & Symmetric & Triangle & Vertical \\
    \includegraphics[valign=c,width=.1\textwidth]{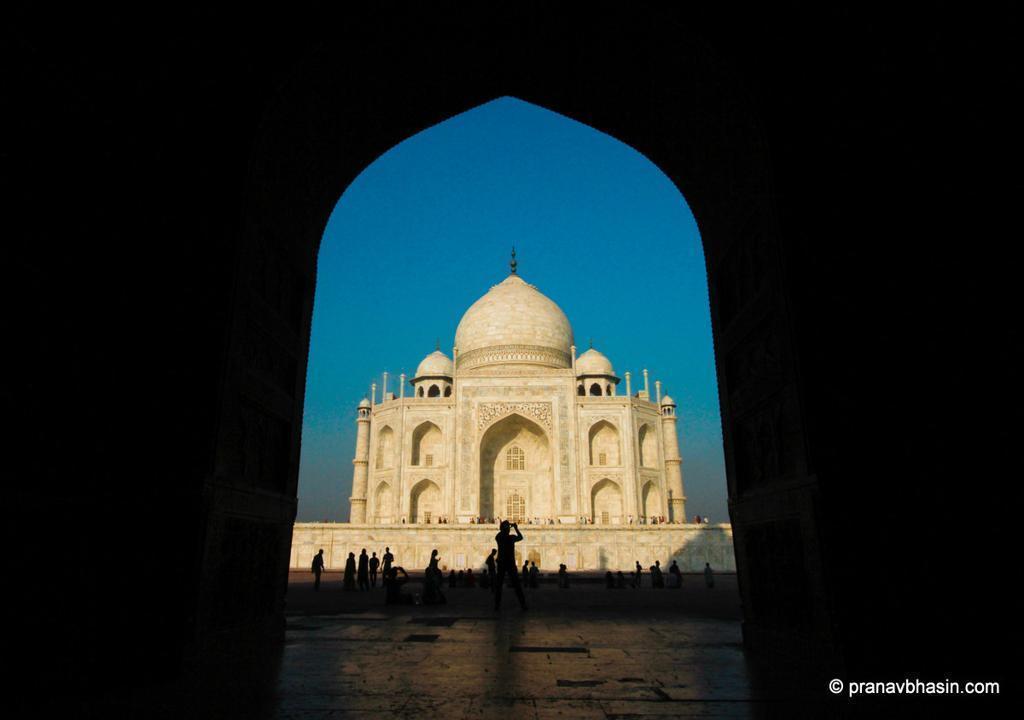} & \includegraphics[valign=c,width=.1\textwidth]{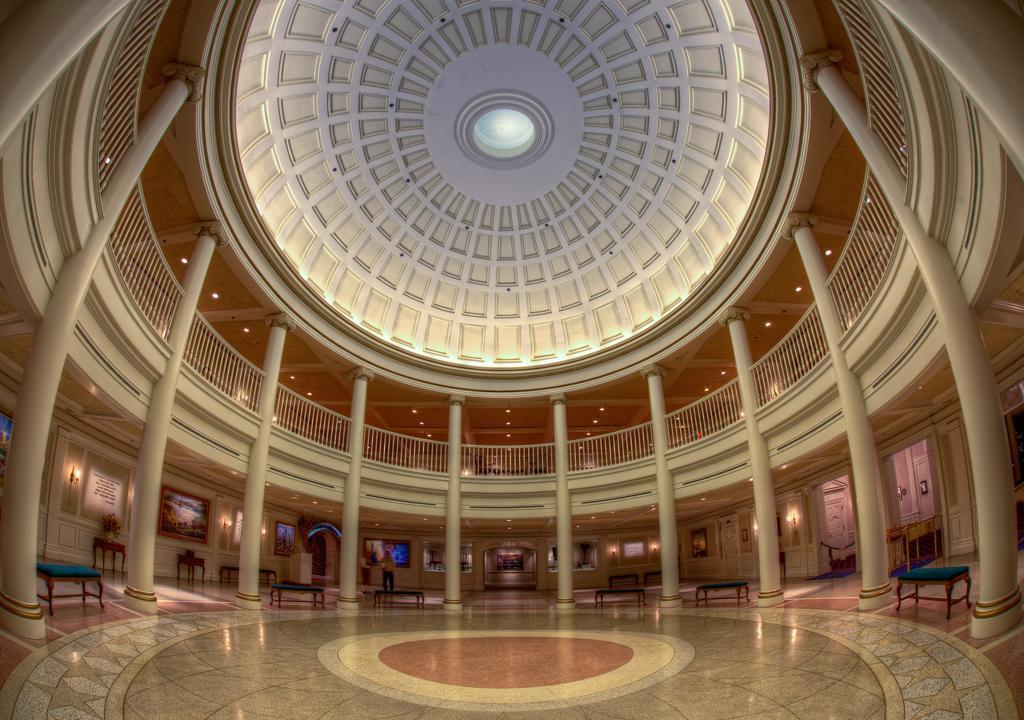} & \includegraphics[valign=c,width=.1\textwidth]{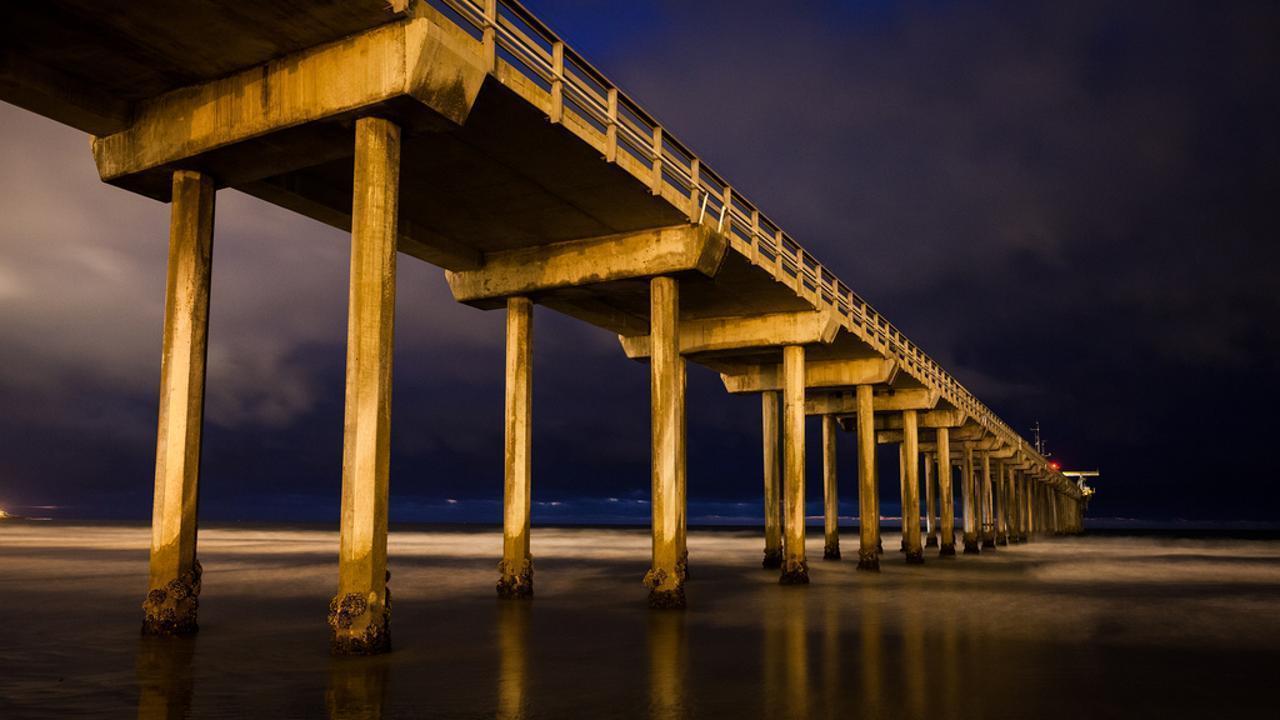} & \includegraphics[valign=c,width=.1\textwidth]{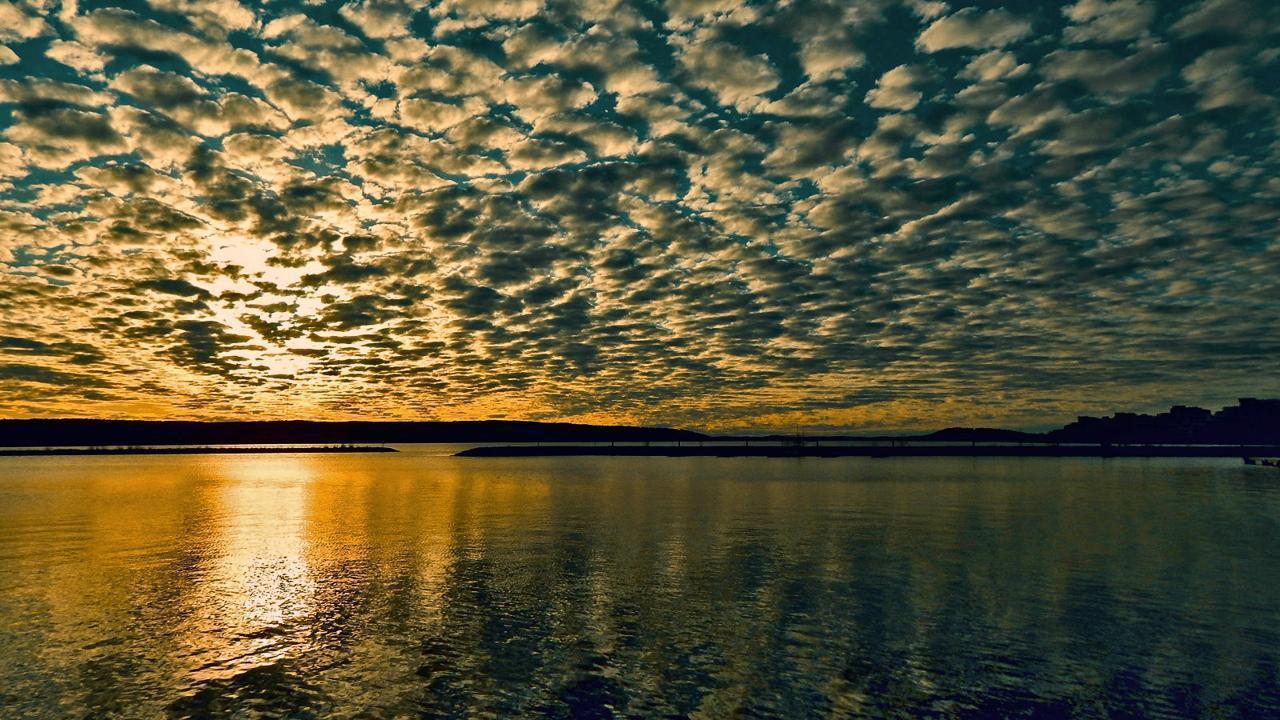} & \includegraphics[valign=c,width=.1\textwidth]{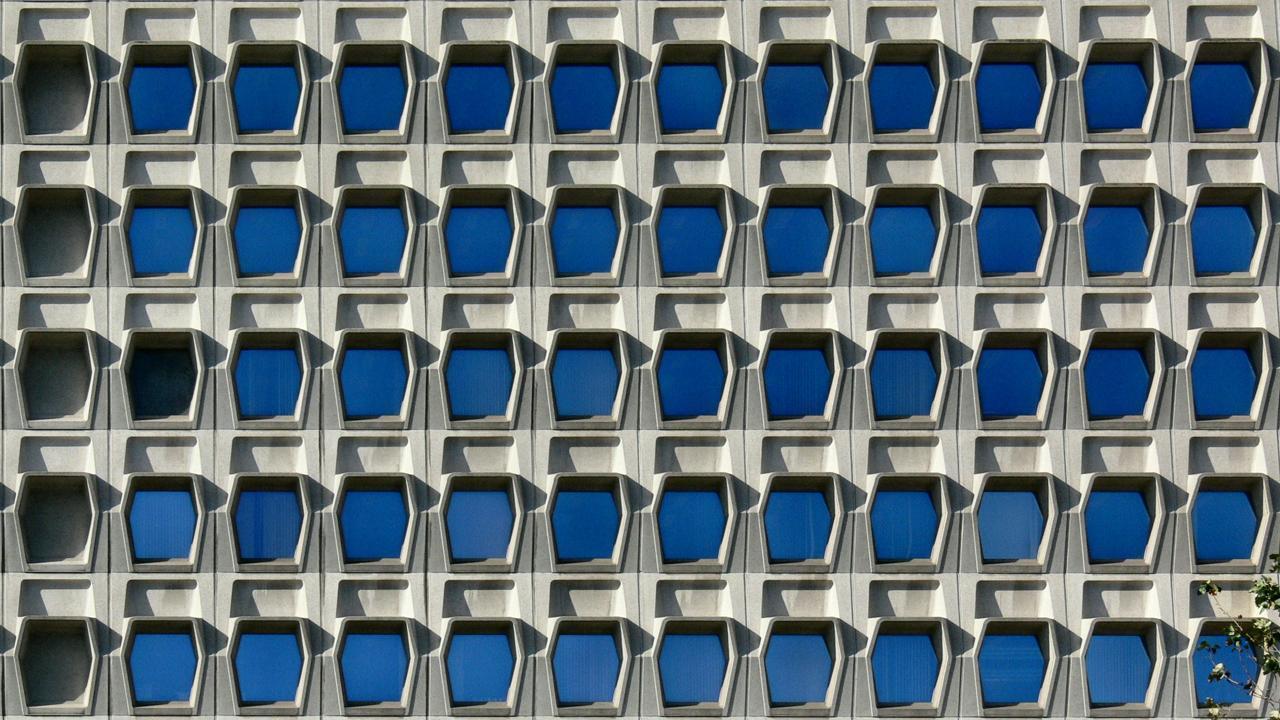} & \includegraphics[valign=c,width=.1\textwidth]{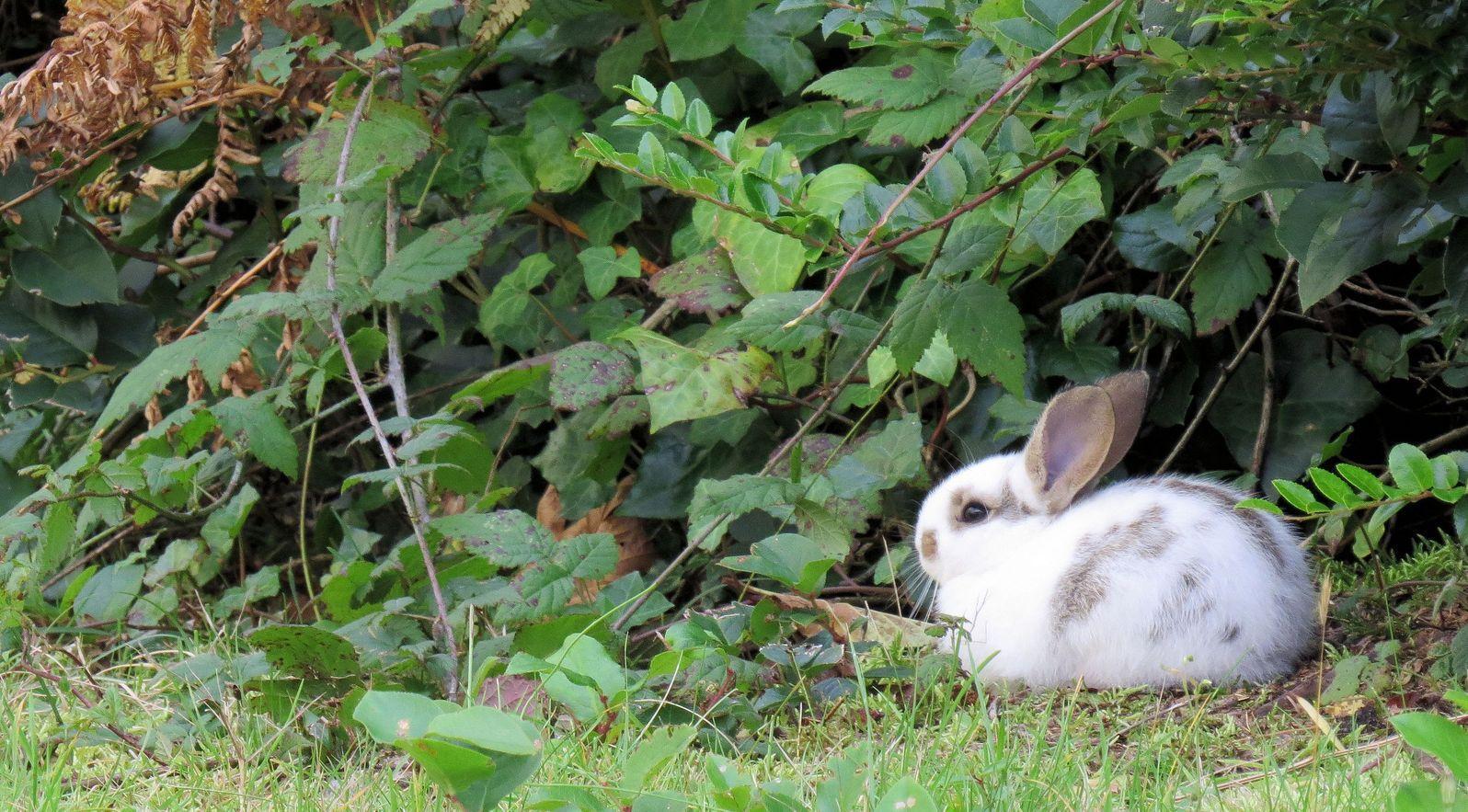} & \includegraphics[valign=c,width=.1\textwidth]{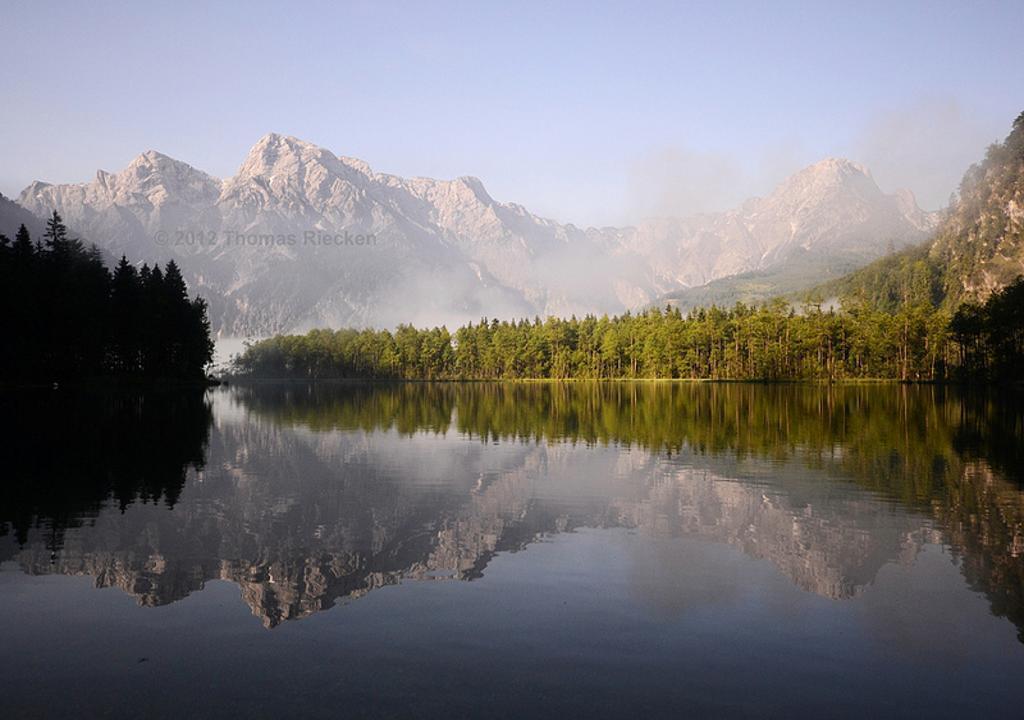} & \includegraphics[valign=c,width=.1\textwidth]{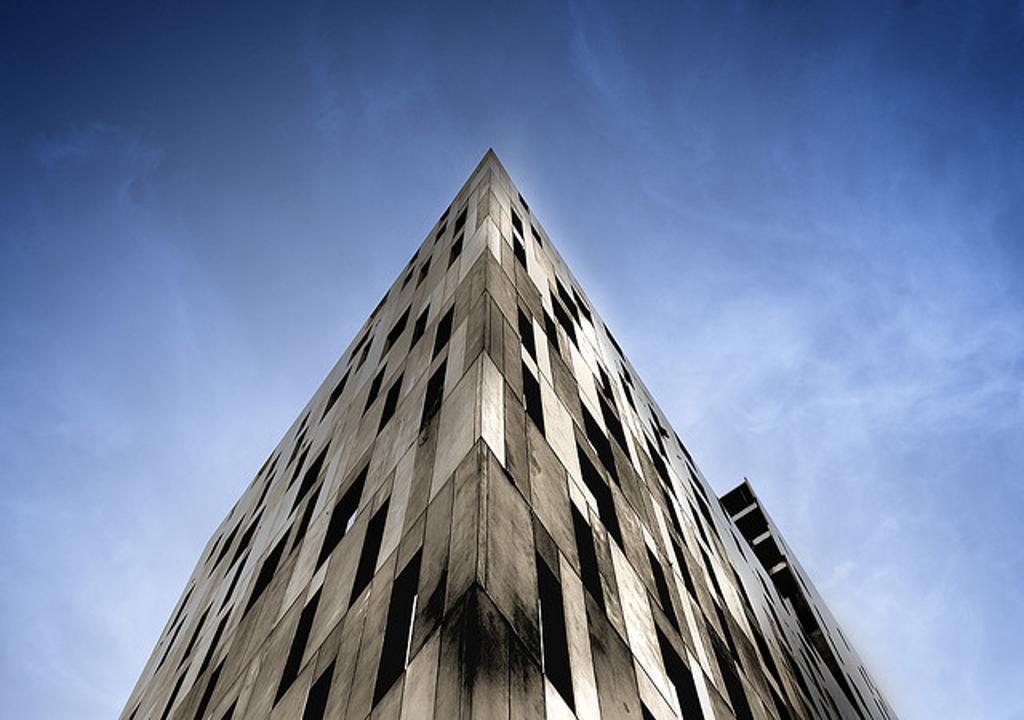} & \includegraphics[valign=c,width=.1\textwidth]{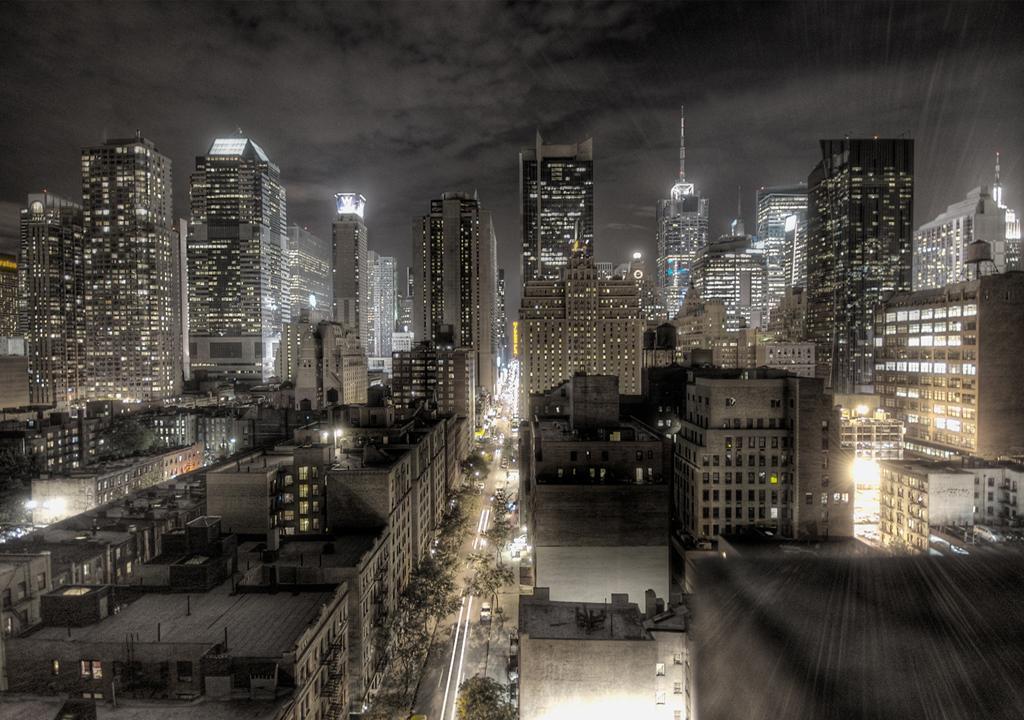}
  \end{tabular}}}
  \caption{Sample images from (a) the style categories of the FlickrStyle database \cite{karayev2014recognizing} and (b) the geometric composition categories of the KU-PCP database \cite{lee2018photographic}.}
  \label{fig:sample-images-flickr-kupcp}
\end{figure*}
The dataset is split into 64,000 training images and 16,000 testing images. At the time of writing, a total of 63,493 is still downloadable. Thus, the new splits result in 50,868 training images and 12,625 testing images. Images sampled from the dataset are shown in Figure \ref{fig:flickr-samples}. The FlickrStyle classes are mutually exclusive, that is an image can not have the label ``Bright'' and the label ``Romantic'' at the same time.

\textbf{KU-PCP:} The KU-PCP dataset \cite{lee2018photographic} consists of 4,244 outdoor photographs (3,169 for training and 1,075 for testing). It has been annotated by 18 human subject to categorize images into nine not mutually exclusive geometric classes: center, curved, diagonal, horizontal, pattern, Rule of Thirds (RoT), symmetric, triangle, and vertical. Sample images for each category are reported in Figure \ref{fig:kupcp-samples}. The KU-PCP classes are not mutually exclusive, that is an image may have the labels ``Rule of Thirds'' and ``Curved'' at the same time.

\subsection{Datasets for image aesthetic assessment}
\begin{figure*}
  \centering
  \subfloat[]{\label{fig:aadb-distr}\includegraphics[width=.3\textwidth]{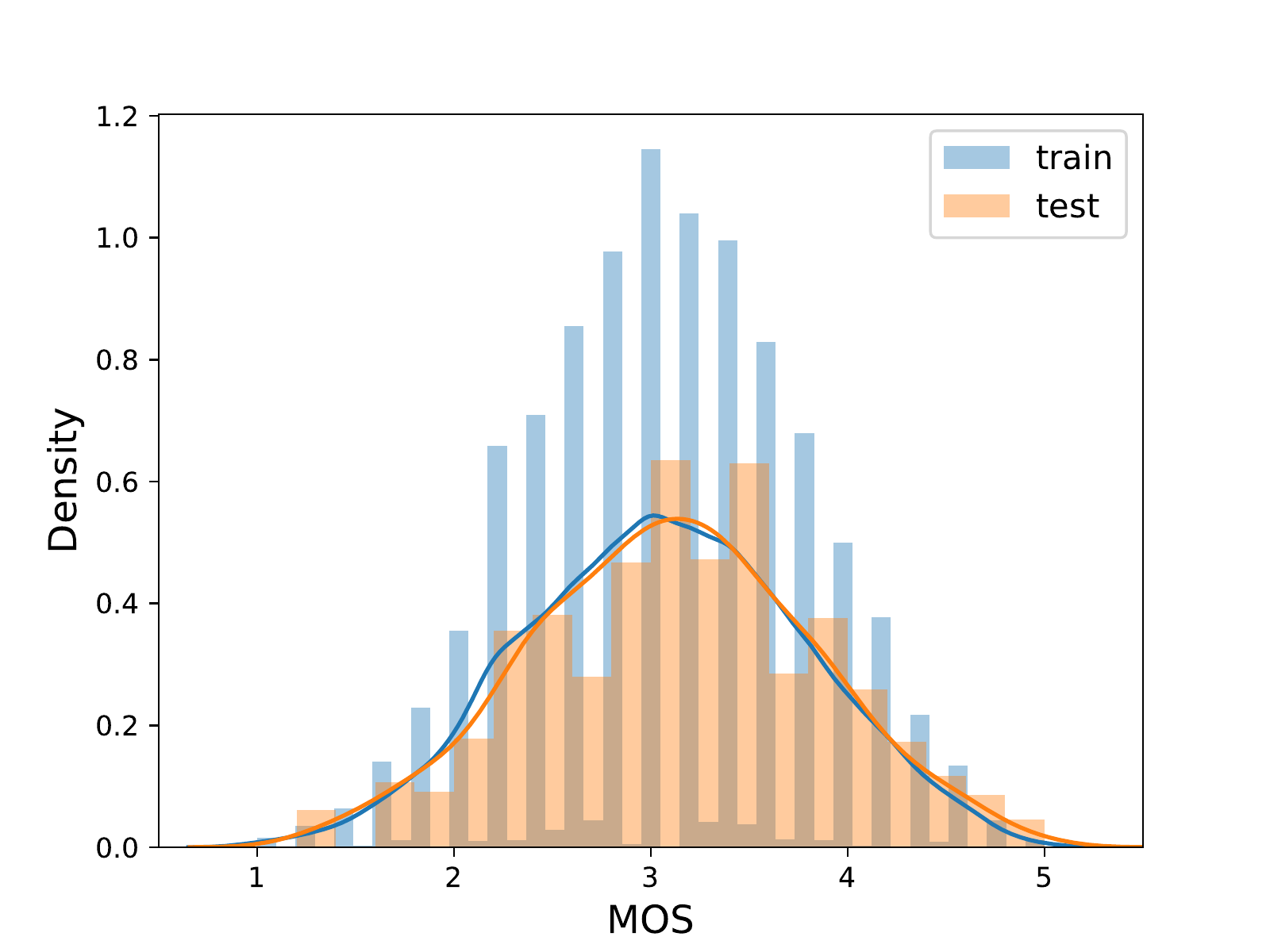}} \quad
  \subfloat[]{\label{fig:ava-distr}\includegraphics[width=.3\textwidth]{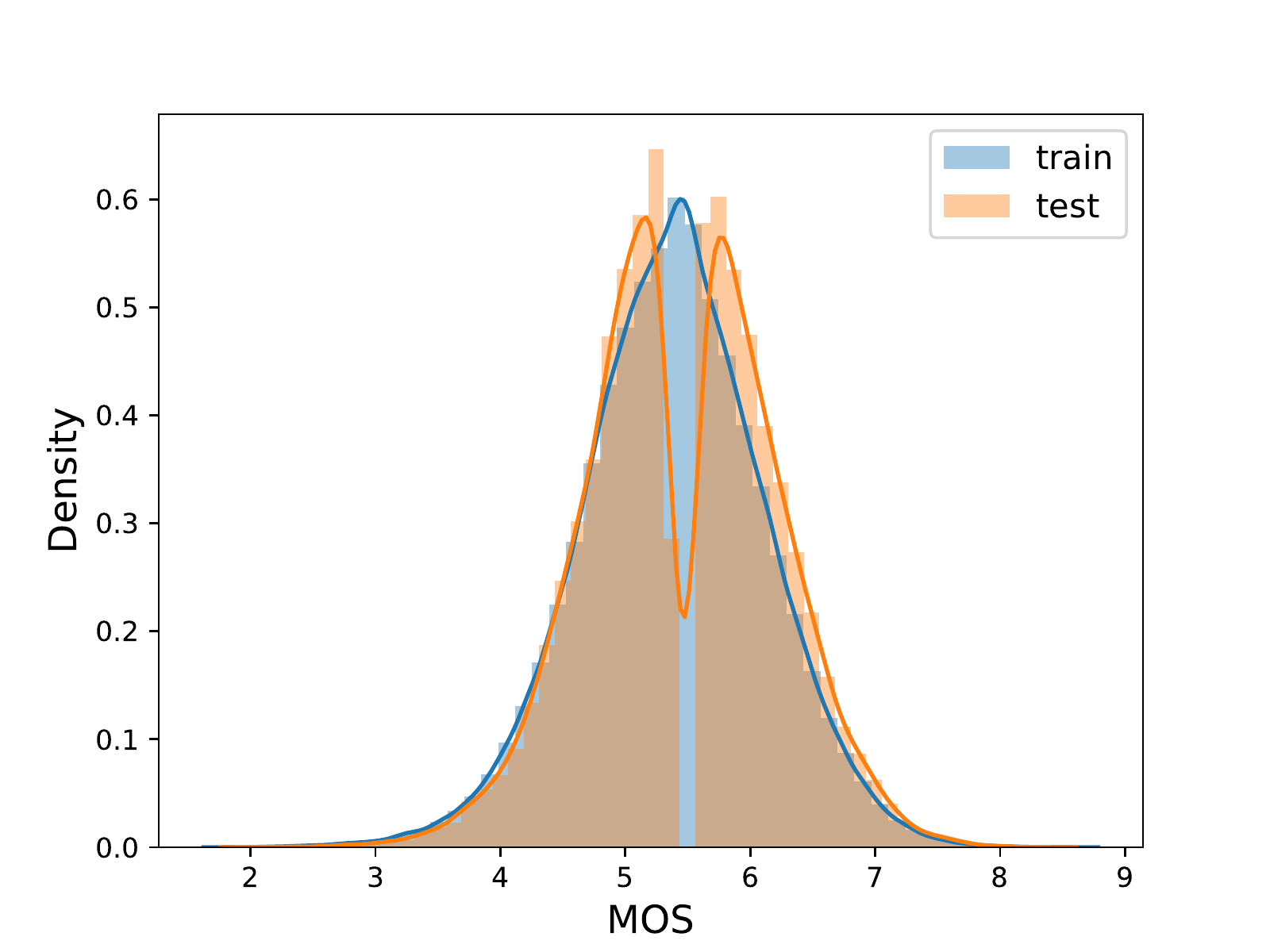}} \quad 
  \subfloat[]{\label{fig:photo-distr}\includegraphics[width=.3\textwidth]{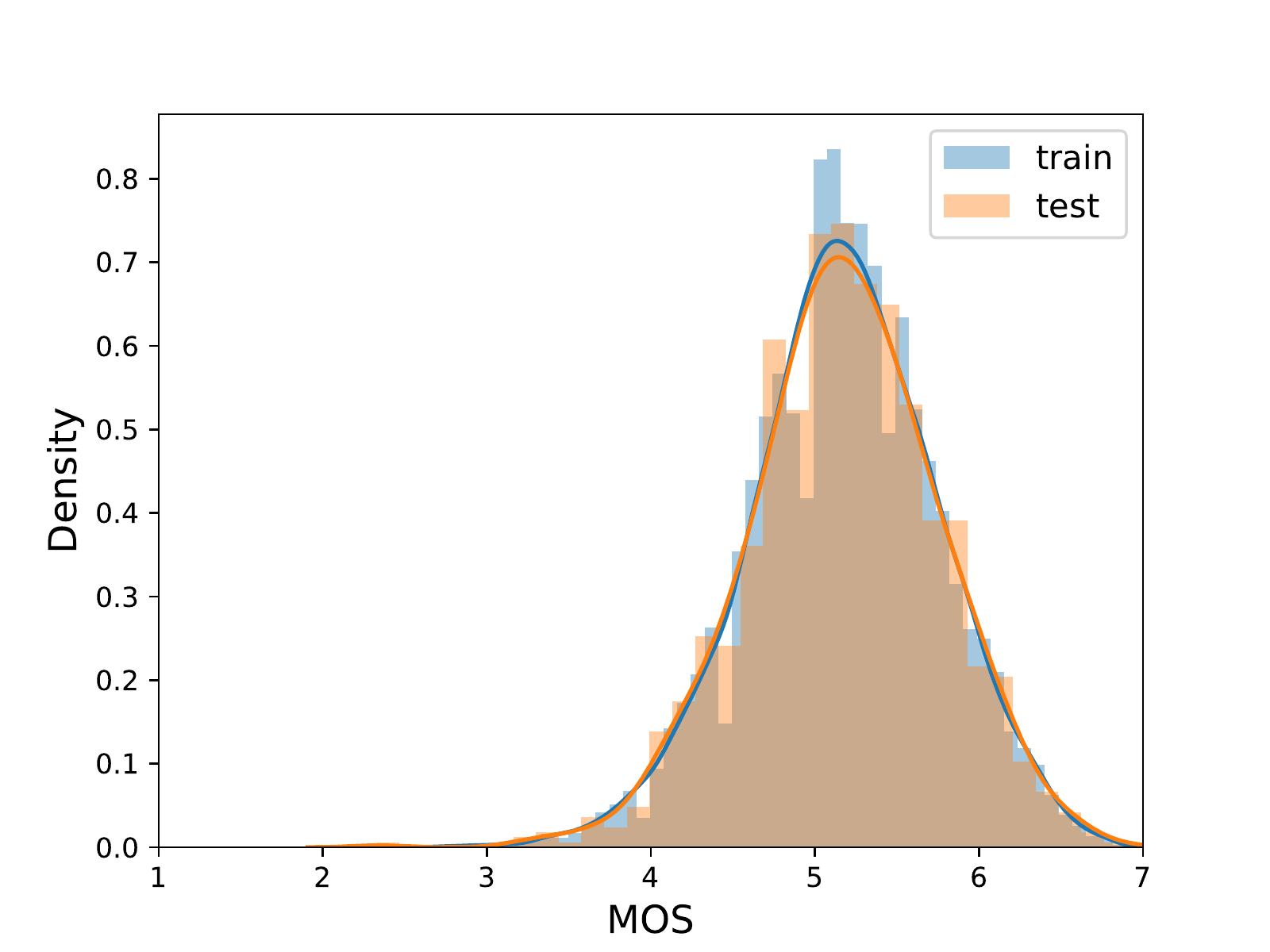}}
  \caption{Distributions of the aesthetic scores on the AADB \cite{kong2016aesthetics} (a), AVA \cite{murray2012ava} (b), and Photo.net \cite{datta2006studying} (c) datasets.}
  \label{fig:mos_distr}
\end{figure*}

\textbf{AADB:} The Aesthetics and Attributes DataBase (AADB) dataset \cite{kong2016aesthetics} contains a set of 10,000 images downloaded from the Flickr website.\footnote{\url{http://www.flickr.com}} Five Amazon Mechanical Turk (AMT) workers annotate each image with an overall aesthetic score and a set of eleven meaningful attributes. These attributes span traditional photographic principals of color, lighting, focus and composition, and are the following: interesting content, object emphasis, good lighting, color harmony, vivid color, shallow depth-of-field, motion blur, Rule of Thirds, balancing element, repetition, and symmetry. For the aesthetic score, AMT workers were allowed to express their judgement on a scale from 1 to 5. For each image, the aesthetic score is the average over all the users judgements. The AADB database was split by its authors into 8,500 images for training, 500 images for validation and 1,000 images for testing. Figure \ref{fig:aadb-distr} shows the distribution of mean ratings for training and test sets.

\textbf{AVA:} The Aesthetics Visual Analysis (AVA) dataset \cite{murray2012ava} is a large-scale and challenging dataset for image aesthetic assessment. It contains more than 250,000 photos gathered from \url{www.dpchallenge.com}. Each image provides three types of annotations: aesthetic ratings ranging from 1 to 10 given by about 200 voters; 0, 1 or 2 textual tags chosen from 66 that describe the semantic content of the image; photographic style annotations corresponding to 14 photographic techniques. From the overall set of images, the authors sampled 20,000 for testing (of which only 19,926 are currently available). Following \cite{hosu2019effective}, the remaining 235,574 images are further randomly split into training (95\%) and validation (5\%) sets. Figure \ref{fig:ava-distr} shows the distribution of mean ratings for training and test sets.

\textbf{Photo.net:} The Photo.net dataset \cite{datta2006studying} is collected from \url{www.photo.net}.\footnote{Available at \url{http://ritendra.weebly.com/aesthetics-datasets.html}} It contains 20,278 images annotated by at least 10 users to assess the aesthetic quality from one to seven. Of all the images in the dataset, only 16,662 have the distribution of aesthetics ratings and are available. Following \cite{zhang2019gated}, from the overall images, 1000 images are used for validation, 1200 images are used for test, and the remaining 14,462 images are used to train. Since the image indexes for each split are not available from \cite{zhang2019gated}, we randomly divide the images based on the previous partitioning. To mitigate any bias due to the data division, we repeat the partitioning 10 times and report the average performance across the 10 runs.
%

\section{Experiments}
\label{sec:experiments}
In this section, we detail the evaluation metrics used to estimate the performance and the training procedure of the proposed method.
\subsection{Evaluation Metrics}
We evaluate the proposed method with respect to three aesthetic quality tasks (i) aesthetic score regression, (ii) aesthetic quality classification, and (iii) aesthetic score distribution prediction. For the aesthetic score regression task, we estimate the mean score of the predicted score distribution via $\mu = \sum_{i=1}^{N}{s_i\times p_i}$, with $s_i$ representing the score bucket and $p_i$ that is the estimated probability for the $i$-th bucket. Following \cite{mai2016composition,ma2017lamp,kao2017deep}, for the aesthetic quality classification, we threshold the mean score using the threshold $T$ such that images with predicted score above $T$ are categorized as high quality and vice versa. The evaluation metrics related to the three task are the following:
\begin{enumerate}
  \item For the image aesthetic score regression, we report results in terms of:
  \begin{itemize}
      \item SROCC - Spearman's Rank-Order Correlation Coefficient: this measures the monotonic relationship between the ground-truth and the predicted scores. It ranges from -1 to 1;
      \item PLCC - Pearson's Linear Correlation Coefficient: PLCC measures the linear correlation between the actual and the predicted scores. It ranges from -1 to 1;
      \item RMSE - Root Mean Squared Error (RMSE) and MAE - Mean Absolute Error (MAE): these metrics range from 0 to $+\infty$ and smaller values indicate better results.
  \end{itemize}
  \item For the image aesthetic quality classification, we measure classification performance in terms of the overall accuracy, defined as $Accuracy = \frac{TP+TN}{P+N}$.
  \item  For the  image aesthetic score distribution prediction, we use Earth Mover's Distance (EMD) to estimate the closeness of the predicted and ground-truth rating distributions. The EMD is defined in Equation \ref{eq:emd} with $r=1$ and lower values of EMD mean better results.
\end{enumerate}
\subsection{Training Procedure}
\label{sec:train-proc}
The learnable parameters of the proposed model are exclusively $\theta_s$ and $\theta_h$, i.e. those of the AttributeNet $f_s$ and those of the HyperNet $h$. In fact, as previously described in Section \ref{sec:method}, the $\theta_b^*$ parameters belong to a  fixed ImageNet-trained Backbone. On the other end, the AestheticNet receives the generated parameters $\hat{\theta}_t$ from the HyperNet.

Similar to \cite{benjamin2018measuring}, we adopt a two-step optimization procedure to introduce attribute-constraint into the HyperNet. We first train the AttributeNet for aesthetic-related attributes recognition. We then freeze the AttributeNet and train the HyperNet for the aesthetic assessment.

Both the AttributeNet and the AestheticNet receive the embedding $\mathbf{e}_b$ produced by the pre-trained Backbone as input. To reduce the training time, as in \cite{hosu2019effective} and \cite{leonardi2021model}, we store the embedding produced by the Backbone for the dataset images instead of calculating them at each training process.

\paragraph{Training for aesthetic-related attributes recognition} We exploit Multi-Task Learning (MTL) to train the Multi Layer Perceptron parameters $\theta_s$ of the AttributeNet for predicting both image style and image composition. Let $\theta_s = \{\mathbf{W}_s, \mathbf{W}_v, \mathbf{W}_c\}$ represent the weights for the AttributeNet. The bias terms are eliminated for simplicity. Given the dataset $\mathcal{D}_{v} = \{(\mathbf{X}_v^{(i)},\mathbf{y}_v^{(i)})\}_{i=1}^M$ for image style recognition and the dataset $\mathcal{D}_{c} = \{(\mathbf{X}_c^{(i)},\mathbf{y}_c^{(i)})\}_{i=1}^N$ for image composition recognition, our AttributeNet aims to minimize the combined loss of both tasks:
\begin{equation}
    \mathrm{argmin}_\mathbf{\theta_s} a_v \sum_{i=1}^M \mathcal{L}_v(\mathbf{e}^{(i)}_b, \mathbf{y}_v^{(i)}) + a_c \sum_{j=1}^N \mathcal{L}_c(\mathbf{e}^{(j)}_b,\mathbf{y}_c^{(j)}),
    \label{eq:train-attrnet}
\end{equation}
where $a_v$ and $a_c$ control the importance of each task and correspond to 1 and 10, respectively (see Sec. \ref{sec:attrnet-hyper} for the analysis of these two hyperparameters). The embedding $\mathbf{e}^{(i)}_b = b(\mathbf{X}_v^{(i)};\theta_b^{*(L)})$ $\mathbf{e}^{(j)}_b = b(\mathbf{X}_c^{(j)};\theta_b^{*(L)})$ are obtained from the Backbone for both the training sets.

The dataset $\mathcal{D}_{v}$ used for style recognition is FlickrStyle. As described in Section \ref{sec:datasets}, the twenty FlickrStyle categories have been annotated as mutually exclusive therefore we adopt cross-entropy as $\mathcal{L}_v$. The KU-PCP dataset instead is adopted as $\mathcal{D}_{c}$. KU-PCP is labeled with nine not mutually exclusive image composition classes. Hence, we use the binary cross-entropy as the $\mathcal{L}_c$ loss function. The cardinality of the image style recognition dataset is greater than that of the image composition: the FlickrStyle dataset has 50,868 training images, while KU-PCP consists of 3,169 training images. For this reason, during the training phase, we balance the number of images between the two datasets by performing data augmentation on KU-PCP. We select augmentation techniques that do not affect the image composition, i.e., color jittering (random adjustment of brightness, contrast, saturation, hue), random horizontal flipping, random grayscale, and random patch erasing.

The size of the embedding vector $\mathbf{e}_s$ is fixed to 512. The learning rate is initially set to $1e^{-4}$ and then dropped by 10 every 20 epochs. We use a batch size of 32, randomly sampling images from both the KU-PCP and the FlickrStyle. We train the model for a maximum of 60 epochs using Adam \cite{kingma2014adam} as optimizer monitoring the accuracy over the validation set to select the best model.
\paragraph{Training for aesthetic assessment}
The second training concerns the $\theta_h$ parameters of the HyperNet for generating the parameters $\hat{\theta}_t$ of the AestheticNet, which in turn manages the aesthetic assessment. In this paper, we formulate the aesthetic assessment as a label distribution prediction problem. More in detail, our network $f_t$ is not trained to predict the Mean Opinion Score (MOS), instead it infers the $l_1$-normalized score distribution $\mathbf{\hat{q}} = [\hat{q}_{s_1}, \hat{q}_{s_2},..., \hat{q}_{s_B}]$. Where ${s_i}$ is the $i$-th score bucket, $B$ is the total number of score buckets, and $\hat{q}_{s_i}$ denotes the number of voters that give the discrete score $s_i$ to the image.

Given the dataset $\mathcal{D}_{t} = \{(\mathbf{X}_t^{(k)},\mathbf{q}^{(k)})\}_{k=1}^N$, the ground-truth of each image $k$ is represented by a score distribution $\mathbf{q} = [q_{s_1},q_{s_2},...,q_{s_B}]$ defined as above. We optimize the HyperNet as follows:
\begin{equation}
    \mathrm{argmin}_{\theta_h} \sum_{k=1}^N\mathcal{L}_{task}(\mathbf{e}_s^{(k)}, \mathbf{q}^{(k)}),
\end{equation}

where $\mathbf{e}_s^{(k)} = f_s(\mathbf{e}_b;\theta_s^{*(M)})$ is the attribute-conditioned embedding obtained from the previously trained AttributeNet for the training image $\mathbf{X}_t^{(k)}$. The loss $\mathcal{L}_{task}$ is the Earth Mover's Distance (EMD). Given the predicted $\mathbf{\hat{q}}$ and the ground-truth $\mathbf{q}$ score distributions, the EMD loss function is defined as follows:
\begin{equation}
  EMD(\mathbf{\hat{q}},\mathbf{q}) = \left(\frac{1}{N} \sum_{k=1}^{N}{|CDF_{\mathbf{\hat{q}}}(k)-CDF_{\mathbf{q}}(k)|^r} \right)^{\frac{1}{r}},
\label{eq:emd}
\end{equation}
where $CDF_*(k)$ is the cumulative distribution function, $r$ equal to 2 is used to penalize the Euclidean distance between the CDFs.

The AestheticNet consists of $M = 5$ linear layers whose output sizes are 512, 256, 256, 64, respectively. The last linear layer have a number of neurons in output equal to the number of buckets of the score distribution which depends on the training dataset: AADB dataset have a total of 5 buckets ($B=5$) with $s_1=1$ and $s_B=5$; AVA dataset have a total of 10 buckets ($B=10$) with $s_1=1$ and $s_B=10$; the Photo.net dataset, $B=7$, $s_1=1$, $s_B=7$. We run this training for 40 epochs exploiting the Adam optimizer. The initial learning rate corresponds to $1e^{-5}$, then it is divided by 10 every 20 epochs. We track the SROCC over the validation set to select the best model.

\section{Results}
\label{sec:results}
In this section, we first report the results obtained by the method proposed for aesthetic-related attribute recognition. We then measure the effectiveness of the proposed method for image aesthetic assessment on the three considered datasets, namely AADB, AVA, and Photo.net. Next, we compare our results with those of many other methods in the state-of-the-art. Finally, we perform an ablation study on the AVA dataset.

\begin{figure*}
  \centering
    \includegraphics[height=87px]{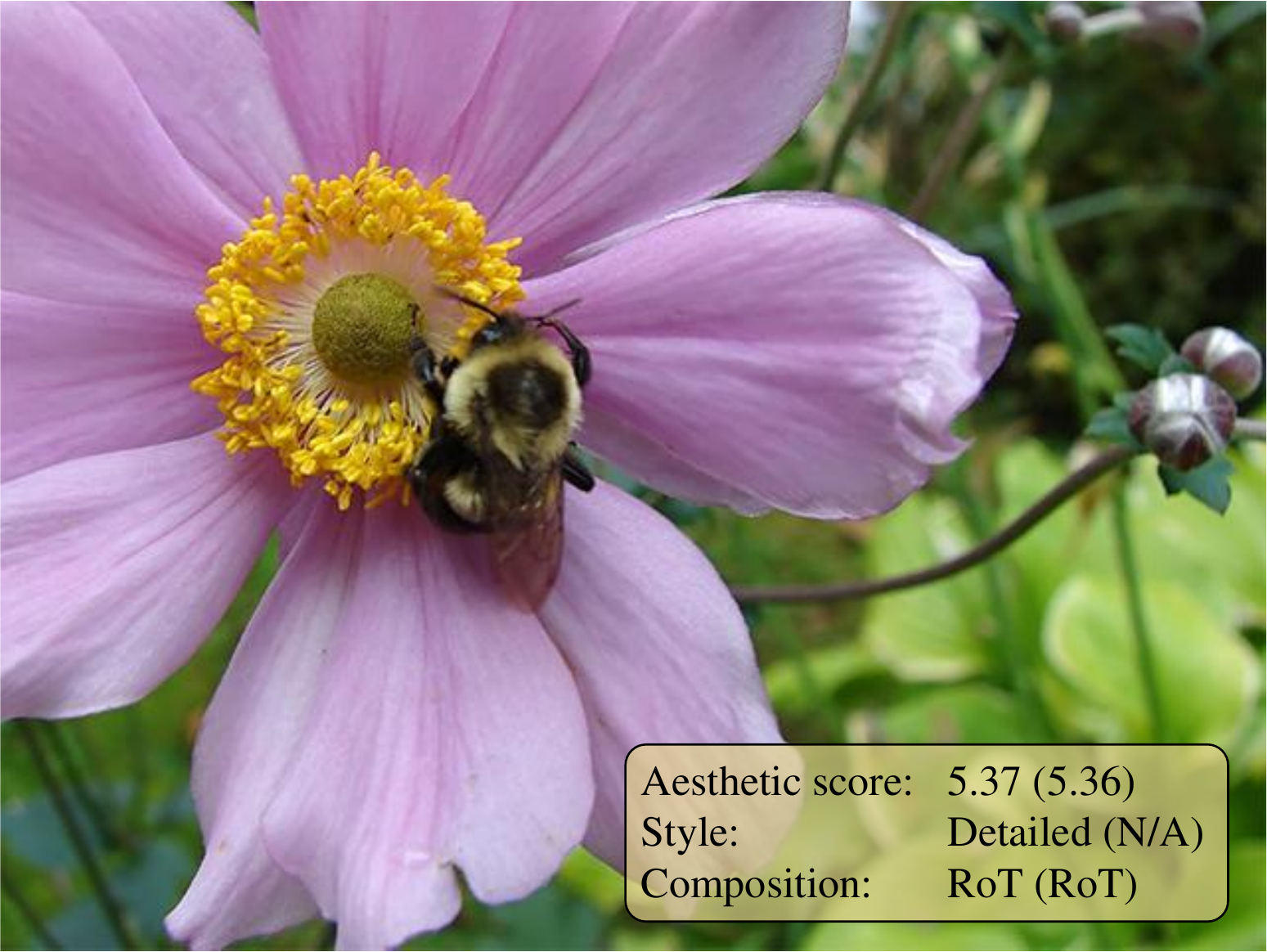} \includegraphics[height=87px]{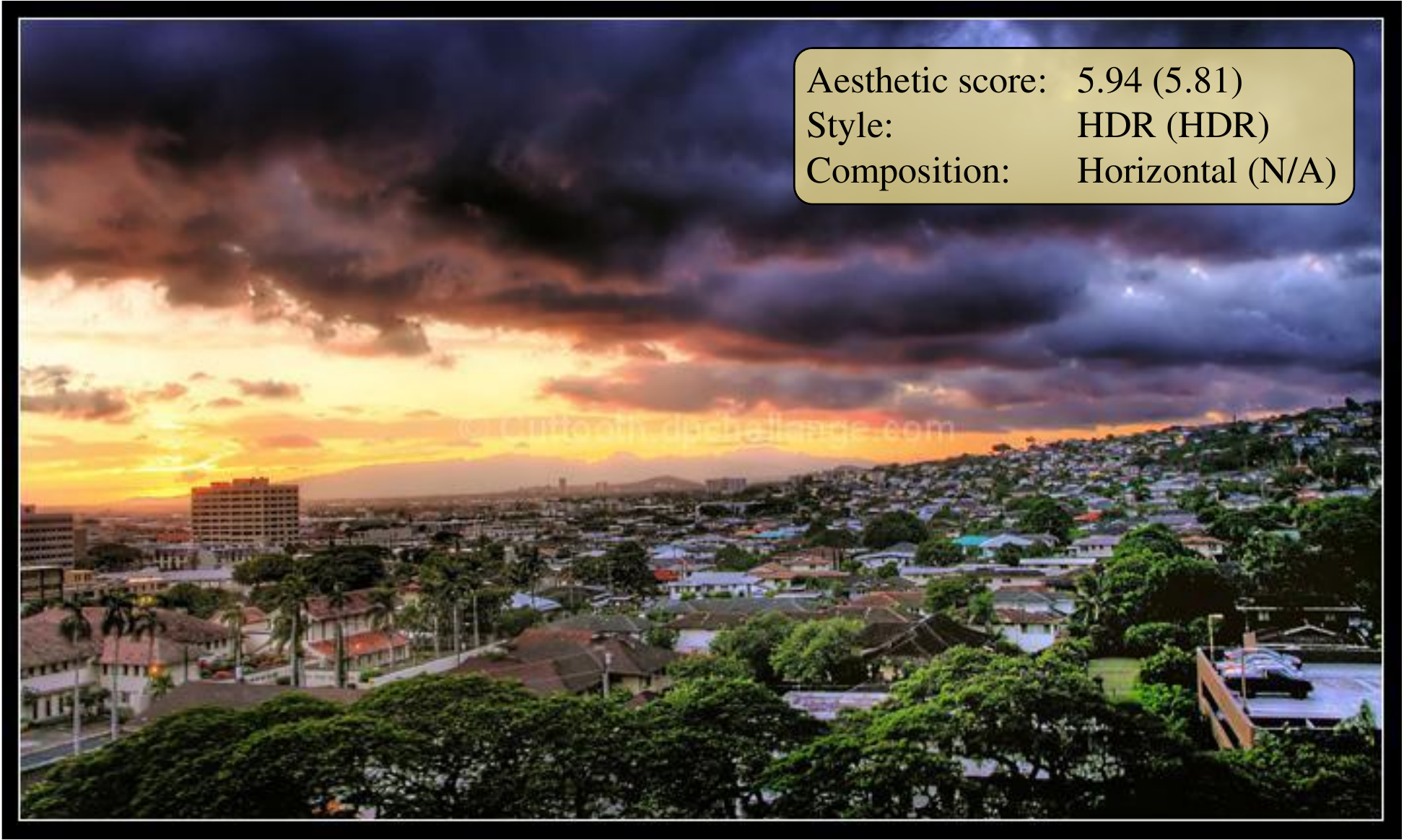} \includegraphics[height=87px]{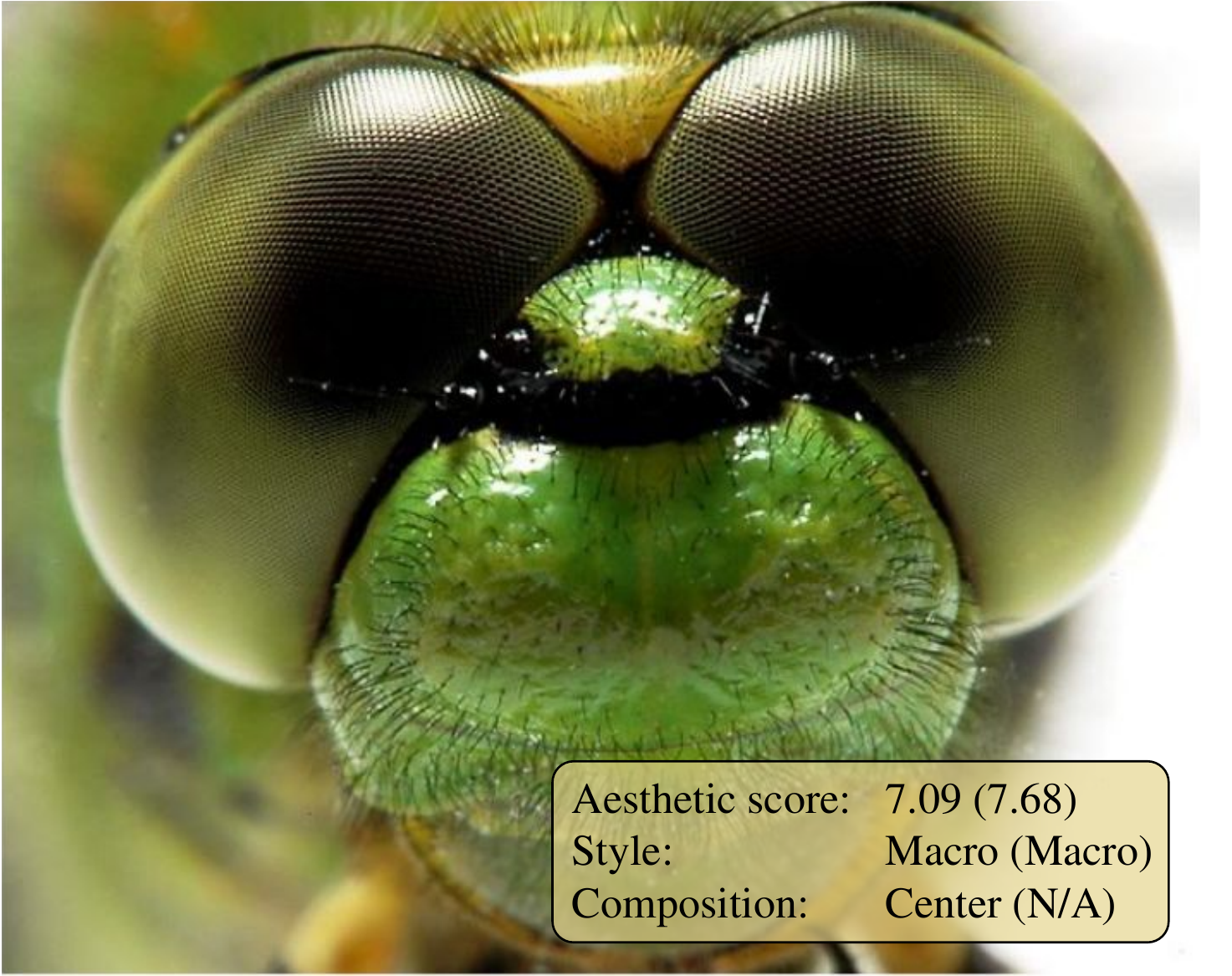} \includegraphics[height=87px]{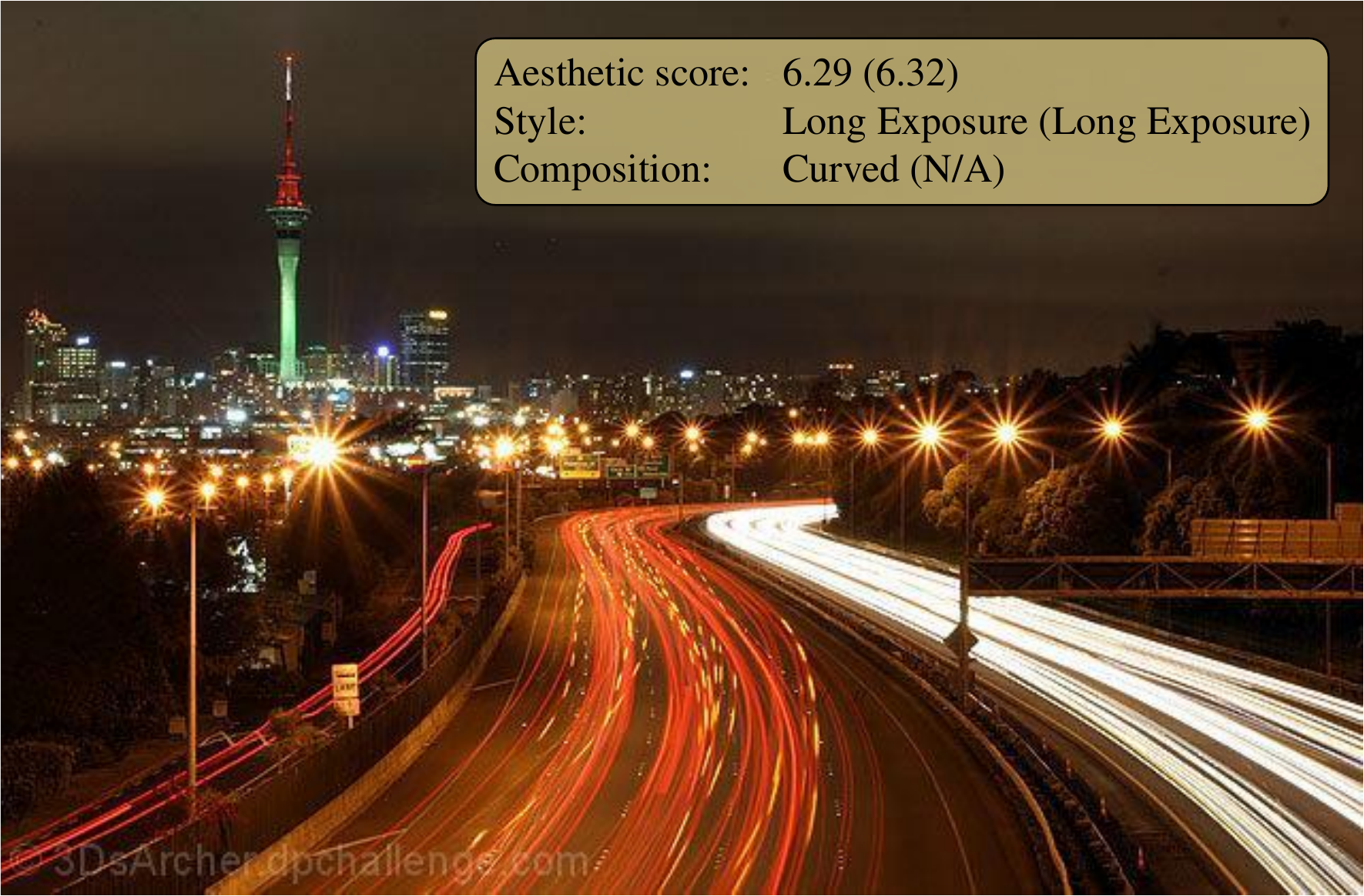}
    \caption{Output produced by the proposed method on sample images from the AVA dataset. For each image, the aesthetic score and the attributes predicted by the proposed method are reported (ground-truth is in brackets). ``N/A'' means that the dataset does not provide any style annotation for the image.}
  \label{fig:sample-output}
\end{figure*}

Figure \ref{fig:sample-output} shows some  predictions produced by the proposed method on the AVA test images. For each image, we report the mean score of the aesthetic distribution and the style and composition tags predicted by our method, as well as the corresponding ground-truths. As can be seen, the proposed method provides an aesthetic quality estimate in close agreement with human judgments. The predictions for composition and style attributes almost coincide with those provided by ground-truth, and even those for which a ground-truth is not available (N/A: absent style or composition information), the prediction is plausible.
\subsection{Image style and composition recognition}
We evaluate the performance of our method for the image style recognition task on the FlickrStyle dataset and for the image composition recognition on the KU-PCP dataset.

\begin{figure}
  \centering
  \includegraphics[width=\columnwidth,height=.85\columnwidth]{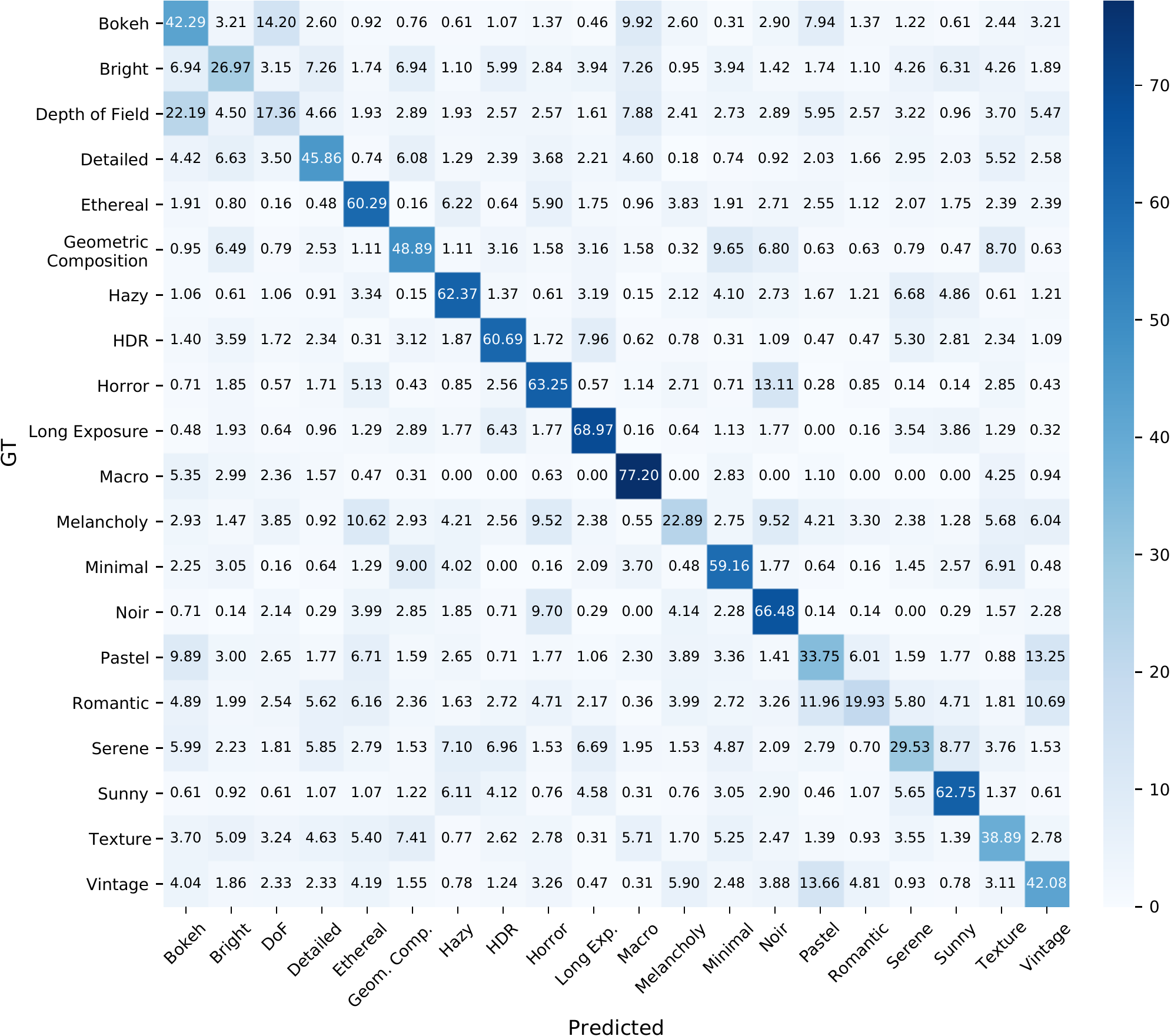}
  \caption{Confusion matrix on the Flickr Style categories.}
  \label{fig:conf-matrix-flick-style}
\end{figure}

We achieve 0.462 of average precision on the 12,625 test images of the FlickStyle dataset. On the other hand, the method proposed by Karayev \etal \cite{karayev2014recognizing} achieves 0.368 of average precision on the FlickStyle test set. 
Moreover, the average precision in \cite{karayev2014recognizing} is estimated on a random subset of the test data balanced so that each class has equal cardinality. Figure \ref{fig:conf-matrix-flick-style} reports the confusion matrix on the 20 style categories. We highlight that our model performs well on styles like Macro (77.20\%  of accuracy), Noir (66.48\%), and Sunny (62.75\%) while it is less effective on styles like Romantic (19.93\%) and Depth-of-Field (17.36\%). The worst misclassification cases concern the following couples: Depth-of-Field vs. Bokeh; Horror vs. Noir; Pastel vs. Vintage.
\begin{figure}
  \centering
  \includegraphics[width=.7\columnwidth]{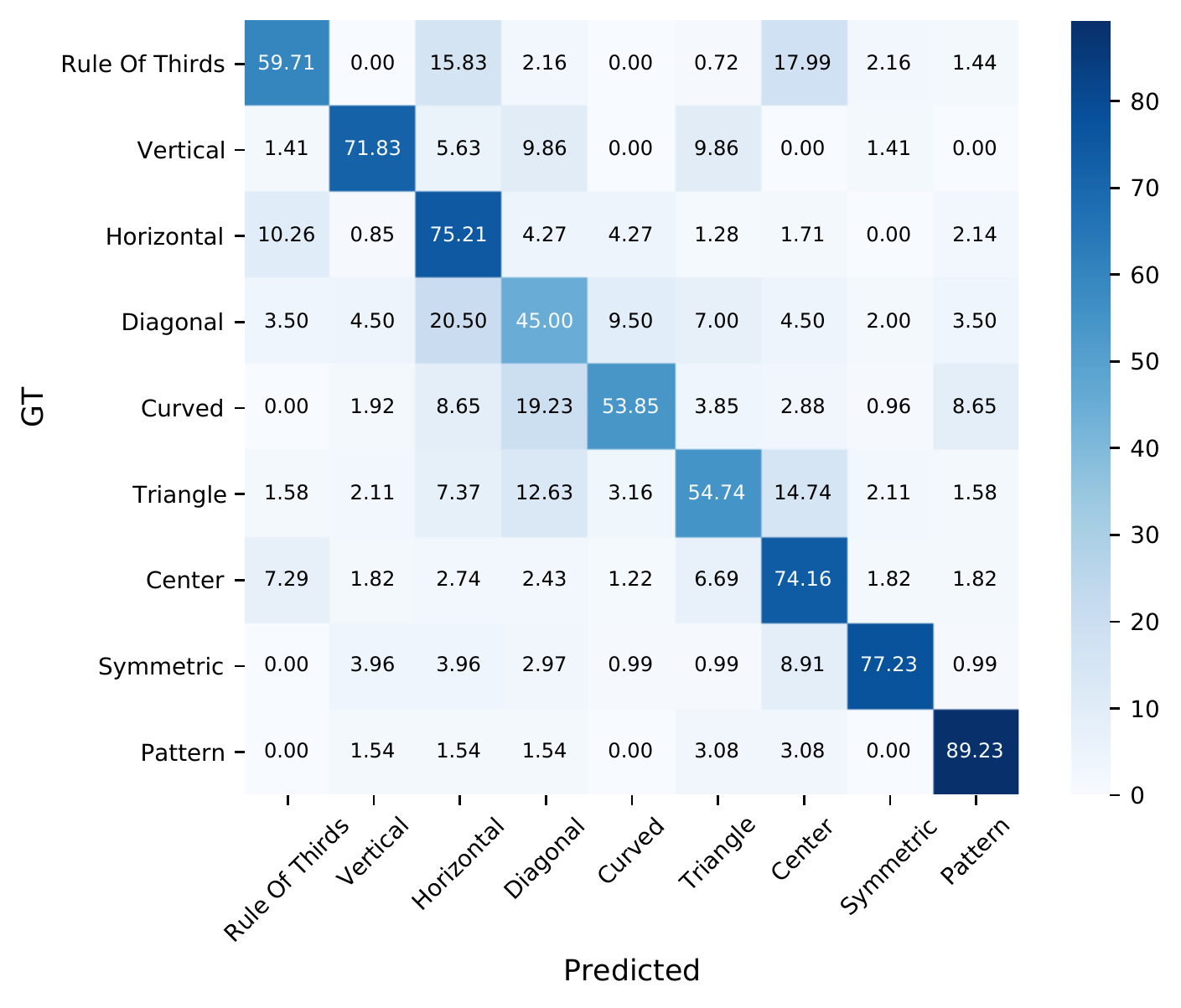}
  \caption{Confusion matrix for image composition recognition on the KU-PCP test set.}
  \label{fig:conf-matrix-kupcp}
\end{figure}

We evaluate the performance of our method for image composition recognition on the test set of the KU-PCP dataset. We measure the accuracy score in terms of $accuracy = \frac{N_{c}}{N}$ where $N$ and $N_{c}$ are the numbers of the total and correctly classified photographs, respectively. Following \cite{lee2018photographic}, we consider an image correctly classified if it is assigned to at least one of the ground-truth composition classes. Our model registers an accuracy of 70.87\% which is in line with the performance of the authors training an SVM classifier over the deep features extracted from a CNN pre-trained on ImageNet, which achieve an accuracy of 70.23\%. In Figure \ref{fig:conf-matrix-kupcp} we show the confusion matrix on the 9 categories of KU-PCP.
%
\subsection{Image aesthetic assessment}
In this section, we present the results obtained by the proposed method for the image aesthetic assessment. For each of the three benchmark datasets, namely AADB, AVA, and Photo.net, we measure performance in terms of Accuracy, Spearman's Rank-Order Correlation Coefficient (SROCC), Pearson's Linear Correlation Coefficient (PLCC), Mean Absolute Error (MAE), Root Mean Squared Error (RMSE), and Earth Mover's Distance (EMD).

We compare ours with several state-of-the-art methods. For these methods, we quote the performance declared in the original papers. In addition to the results of the different methods, we calculate a baseline (named as ``Baseline'') by considering a dummy solution that assigns the average of the training set scores to each test image. More specifically, for each test image, we generate a normal distribution with a mean equal to the mean of the training set scores and a randomly sampled standard deviation in the interval [-0.5,0.5].

Table \ref{tab:results-aadb} reports the results on the AADB dataset. From the results it is possible to draw several conclusions. First, the proposed method outperforms all state-of-the-art methods for the SROCC metric which is the only one to be reported by all methods. Second, our method outperforms Leonardi \etal \cite{leonardi2021model} for all metrics. In particular, our accuracy is 2\% higher than that of the Leonardi. Finally, the third method, i.e. \textit{RGNet} \cite{liu2020composition}, has a SROCC of 0.03 lower than the proposed method.
\begin{table*}
  \centering
  \caption{Comparison of the proposed method with state-of-the-art methods on the AADB dataset. The ``--'' means that the result is not available. The network architecture and whether it uses Multi-Task Learning (MTL) is indicated for each method.}
  \label{tab:results-aadb}
  \begin{tabular}{lcccccccc}
  \toprule
  Method & Network architecture & MTL & Accuracy (\%) $\uparrow$ & SROCC $\uparrow$ & PLCC $\uparrow$ & MAE $\downarrow$ & RMSE $\downarrow$ & EMD $\downarrow$ \\ \midrule
  Baseline & & & 61.58 & -0.0744 & -0.0543 & 0.1449 & 0.1799 & 0.1407 \\ \midrule
  Reg-Net~\cite{kong2016aesthetics} & AlexNet & & -- & 0.6782 & -- & -- & -- & -- \\
  Malu \etal~\cite{malu2017learning} & ResNet-50 & \checkmark & -- & 0.6890 & -- & -- & -- & -- \\ 
  PI-DCNN~\cite{shu2020learning} & ResNet-50 & \checkmark & -- & 0.7051 & -- & -- & -- & -- \\
  Chen \etal~\cite{chen2020data} & ResNet-50 & & -- & 0.7080 & -- & -- & -- & -- \\
  Pan \etal~\cite{pan2019image} & ResNet-50 & \checkmark & -- & 0.7041 & -- & -- & -- & -- \\
  Reddy \etal~\cite{reddy2020measuring} & EfficientNet-B4 & \checkmark & -- & 0.7059 & -- & -- & -- & -- \\
  RGNet~\cite{liu2020composition} & DenseNet-121 & & -- & 0.7104 & -- & -- & -- & -- \\
  Leonardi \etal~\cite{leonardi2021model} & EfficientNet-B4 & & 79.51 & 0.7454 & 0.7479 & 0.1062 & 0.1351 & -- \\
  Proposed & EfficientNet-B4 & \checkmark & \textbf{81.64} & \textbf{0.7567} & \textbf{0.7616} & \textbf{0.0832} & \textbf{0.1059} & \textbf{0.0951} \\
  \bottomrule
  \end{tabular}
\end{table*}
%

In Table \ref{tab:results-ava} the results on the AVA dataset are reported. The \textit{Baseline} obtained a precision of 71.28\% which is 12\% lower than the best precision corresponding to 83.59\% for \textit{RGNet} and only 3\% lower than the method with the worst accuracy, namely \textit{RAPID}. This indicates that methods having a good fit with the training score distribution tend to perform well on test data since the two distributions are very similar (see Fig. \ref{fig:ava-distr} for details). Interestingly, no method achieves the best performance for all metrics. \textit{RGNet} obtains the best accuracy, while MLSP \cite{hosu2019effective} shows the highest regression metrics against mid-ranking accuracy (81.68\%). The proposed method ranks second for the regression metrics, first for the EMD metric, while it achieves an accuracy 3\% lower than the 83.59\% by \textit{RGNet}.
\begin{table*}
  \centering
  \caption{Comparison of the proposed method with state-of-the-art methods on the AVA dataset. In each column, the best and second-best results are marked in \textbf{boldface} and \underline{underlined}, respectively. The ``--'' means that the result is not available. The network architecture and whether it uses Multi-Task Learning (MTL) is indicated for each method.}
  \label{tab:results-ava}
  \begin{tabular}{lcccccccc}
  \toprule
    Method & Network architecture & MTL & Accuracy (\%) $\uparrow$ & SROCC $\uparrow$ & PLCC $\uparrow$ & MAE $\downarrow$ & RMSE $\downarrow$ & EMD $\downarrow$ \\ \midrule
    Baseline & & & 71.28 & -0.0003 & -0.0021 & 0.6230 & 0.7550 & 0.0743 \\ \midrule
    RAPID~\cite{lu2014rapid} & AlexNet &  & 74.20 & – & – & – & – & – \\ 
    DMA-Net~\cite{lu2015deep} & AlexNet &  & 75.42 & – & – & – & – & – \\ 
    MNA-CNN~\cite{mai2016composition} & VGG16 &  & 76.10 & – & – & – & – & – \\ 
    Reg-Net~\cite{kong2016aesthetics} & AlexNet & & 77.33 & 0.5581 & -- & -- & -- & -- \\ %
    MTCNN~\cite{kao2017deep} & VGG16 & \checkmark & 78.56 & – & – & – & – & – \\ 
    Multimodal DBM Model~\cite{zhou2016joint} & VGG16 &  & 78.88 & – & – & – & – & – \\ 
    NIMA~\cite{talebi2018nima} & VGG16 &  & 80.60 & 0.5920 & 0.6100 & – & – & 0.0520 \\ 
    GPF-CNN~\cite{zhang2019gated} & VGG16 &  & 80.70 & 0.6762 & 0.6868 & 0.4144 & 0.5347 & 0.0460 \\ 
    NIMA~\cite{talebi2018nima} & InceptionNet &  & 81.51 & 0.6120 & 0.6360 & – & – & 0.0500 \\ 
    MLSP~\cite{hosu2019effective} & InceptionNet &  & 81.68 & \textbf{0.7524} & \textbf{0.7545} & \textbf{0.3831} & \textbf{0.4943} & – \\ 
    GPF-CNN~\cite{zhang2019gated} & InceptionNet &  & 81.81 & 0.6900 & 0.7042 & 0.4072 & 0.5246 & 0.0450 \\ 
    MULTIGAP~\cite{hii2017multigap} & InceptionNet &  & 82.27 & – & – & – & – & – \\ 
    A-Lamp~\cite{ma2017lamp} & VGG16 &  & 82.50 & -- & -- & -- & – & -- \\ 
    AFDC+SPP \cite{chen2020adaptive} & ResNet-50 &  & \underline{83.24} & 0.6489 & 0.6711 & -- & -- & \underline{0.0447} \\ 
    PI-DCNN~\cite{shu2020learning} & ResNet-50 & \checkmark & -- & 0.6578 & -- & -- & -- & -- \\ 
    Pan \etal~\cite{pan2019image} & ResNet-50 & \checkmark & -- & 0.7041 & -- & -- & -- & -- \\ 
    RGNet~\cite{liu2020composition} & DenseNet-121 &  & \textbf{83.59} & -- & -- & -- & -- & -- \\ 
    Proposed & EfficentNet-B4 & \checkmark & 80.75 & \underline{0.7318} & \underline{0.7329} & \underline{0.4011} & \underline{0.5128} & \textbf{0.0439} \\ 
    \bottomrule
    \end{tabular}
\end{table*}

\begin{figure*}
  \centering
  \setlength{\tabcolsep}{2pt}
  \begin{tabular}{ccccc}
    \includegraphics[width=.18\textwidth]{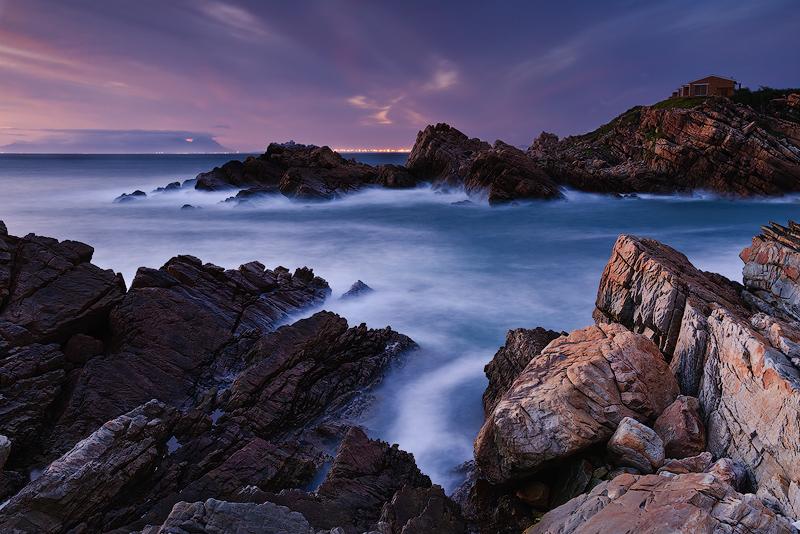} & \includegraphics[width=.18\textwidth]{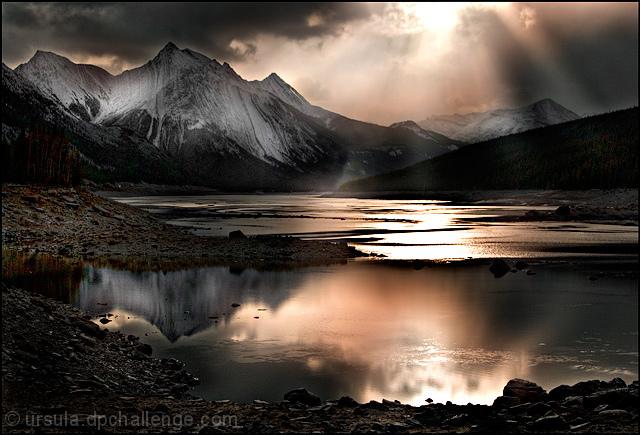} & \includegraphics[width=.18\textwidth]{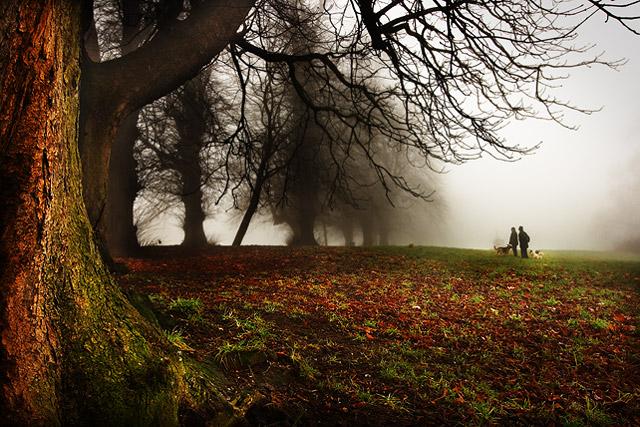} & \includegraphics[width=.18\textwidth]{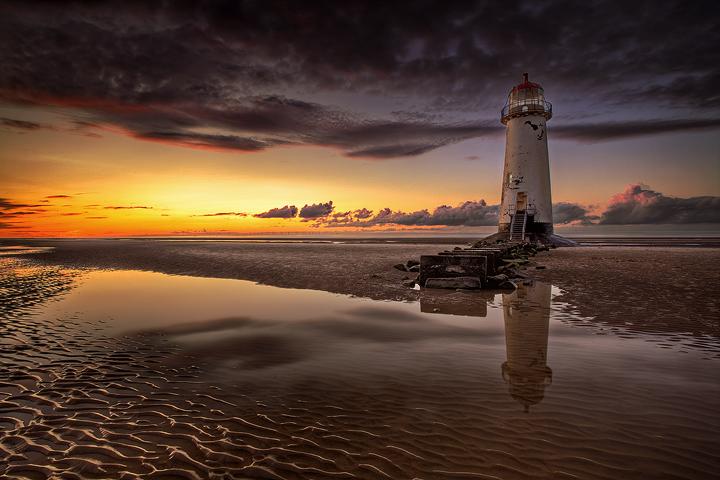} & \includegraphics[width=.18\textwidth]{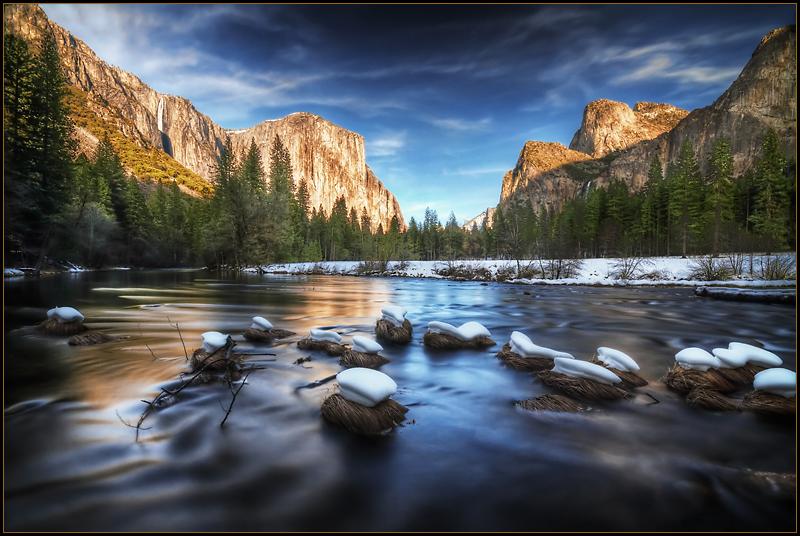} \\
    \includegraphics[width=.18\textwidth]{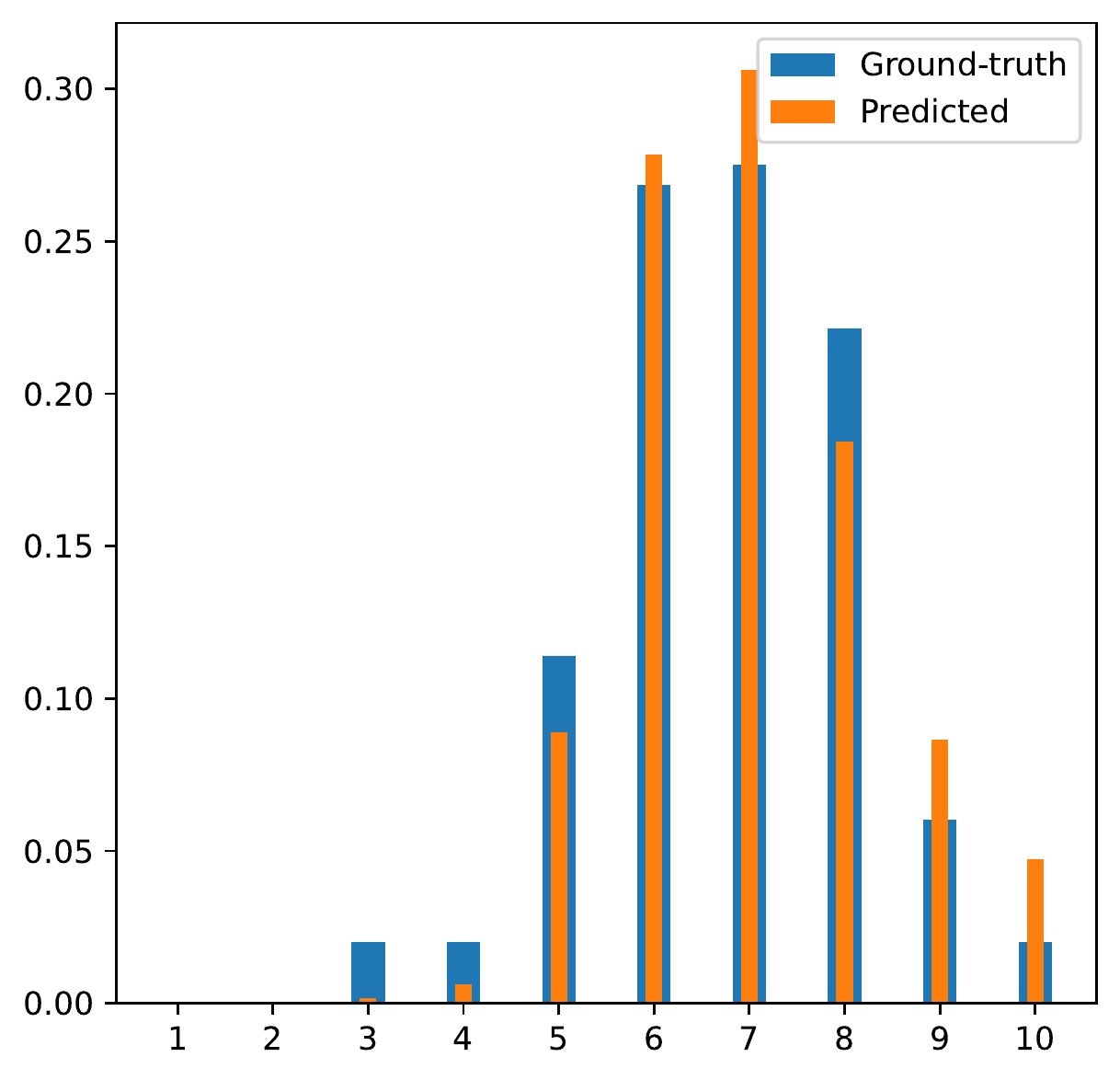} & \includegraphics[width=.18\textwidth]{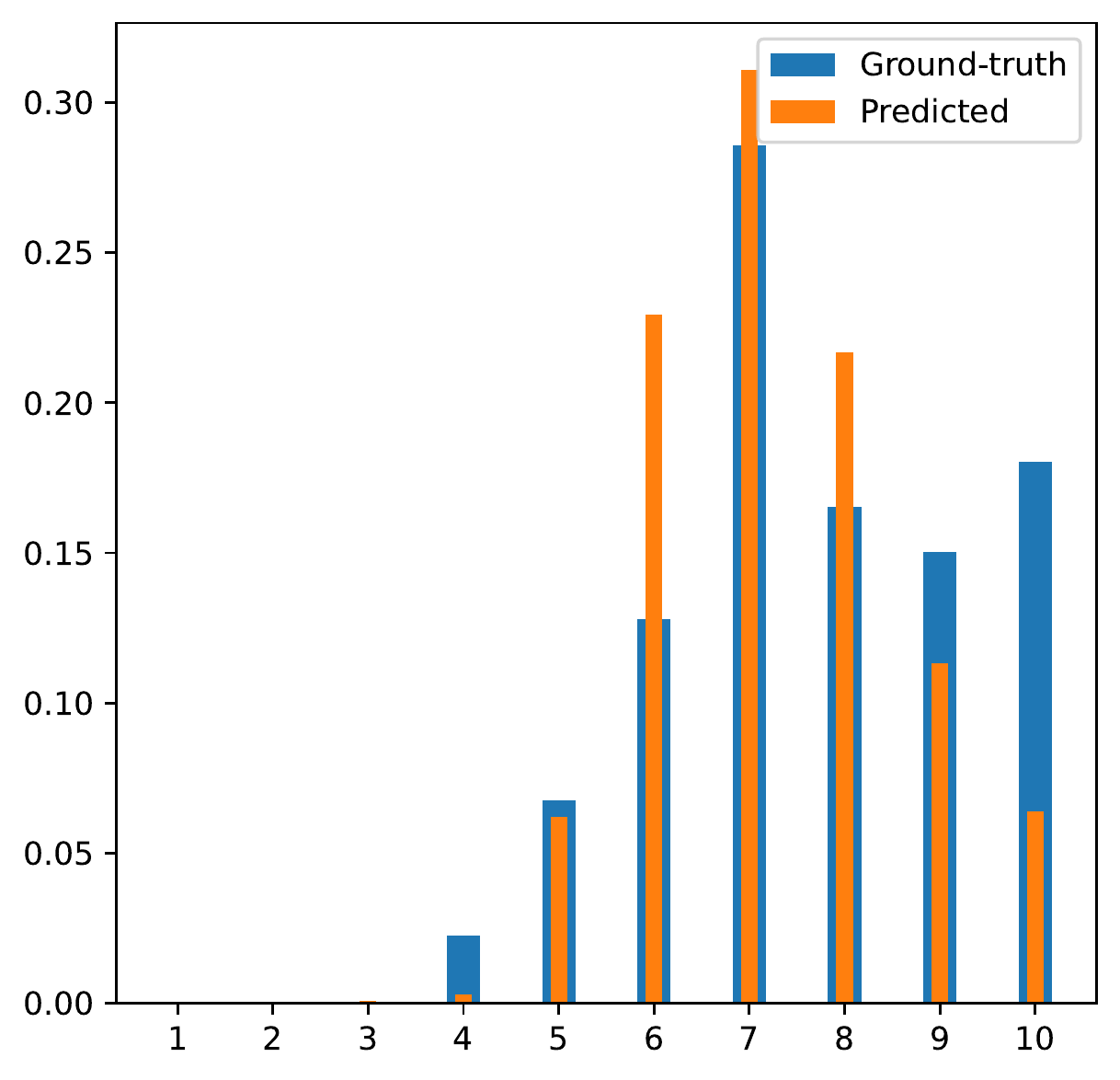} & \includegraphics[width=.18\textwidth]{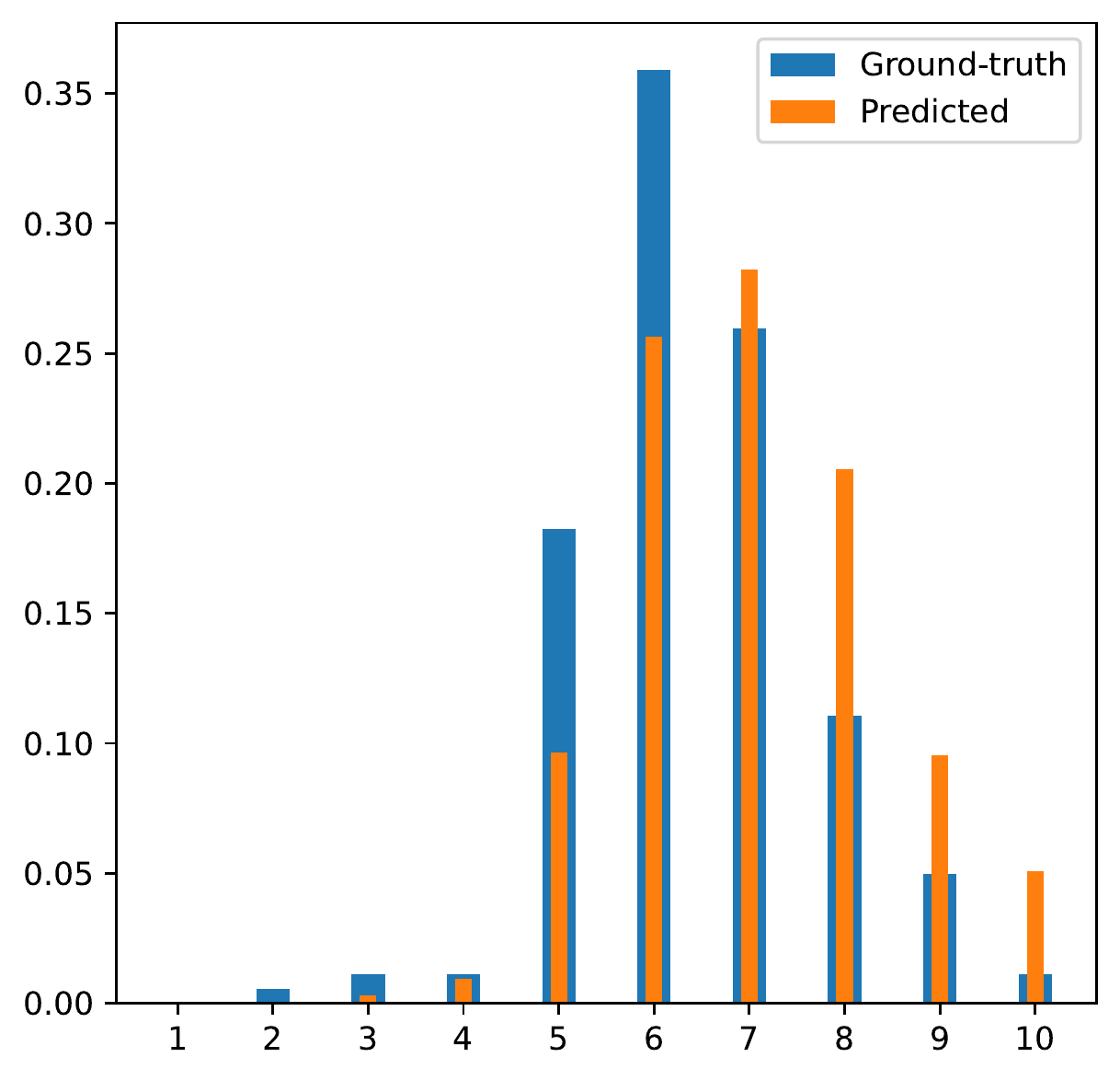} & \includegraphics[width=.18\textwidth]{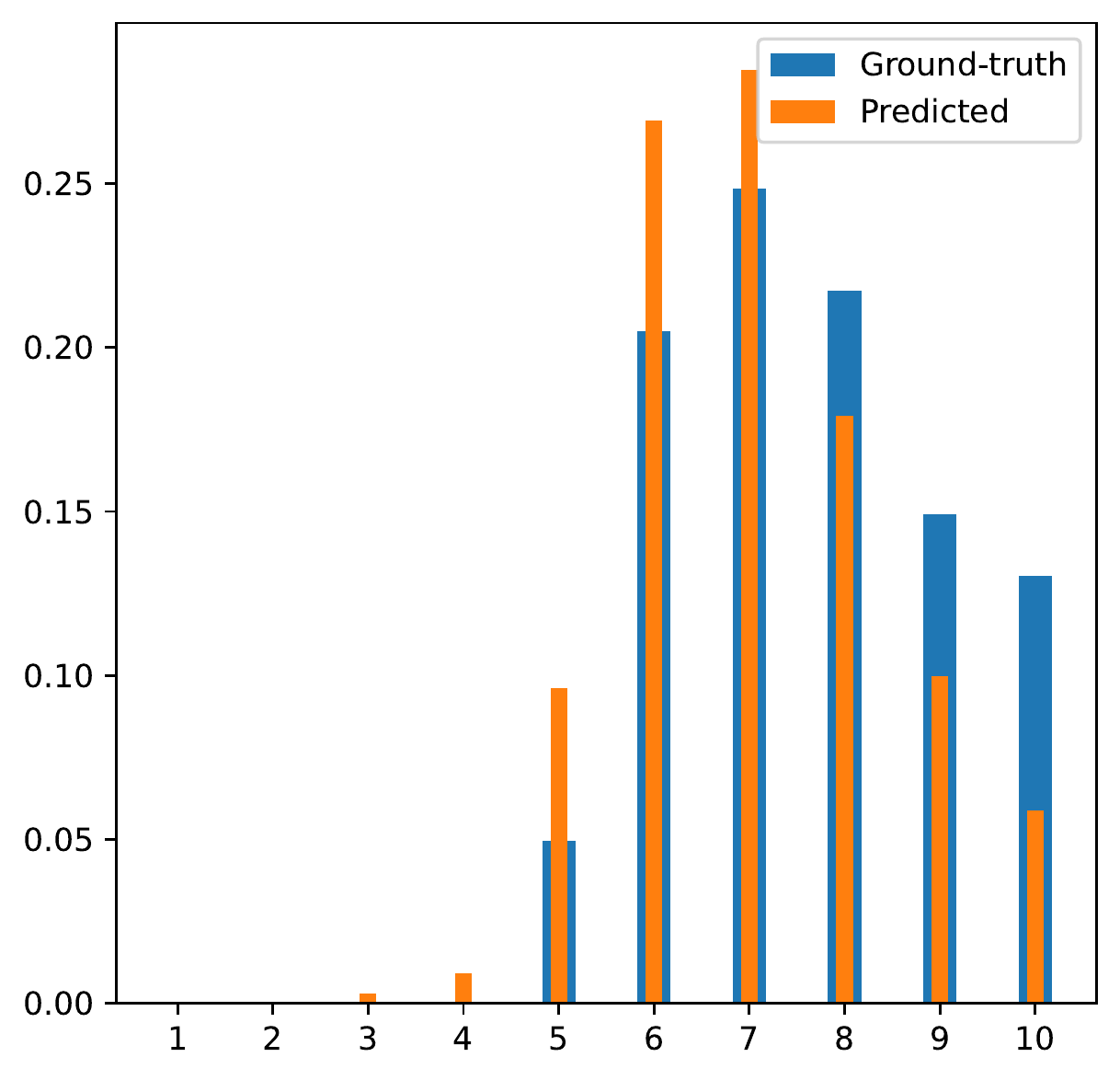} & \includegraphics[width=.18\textwidth]{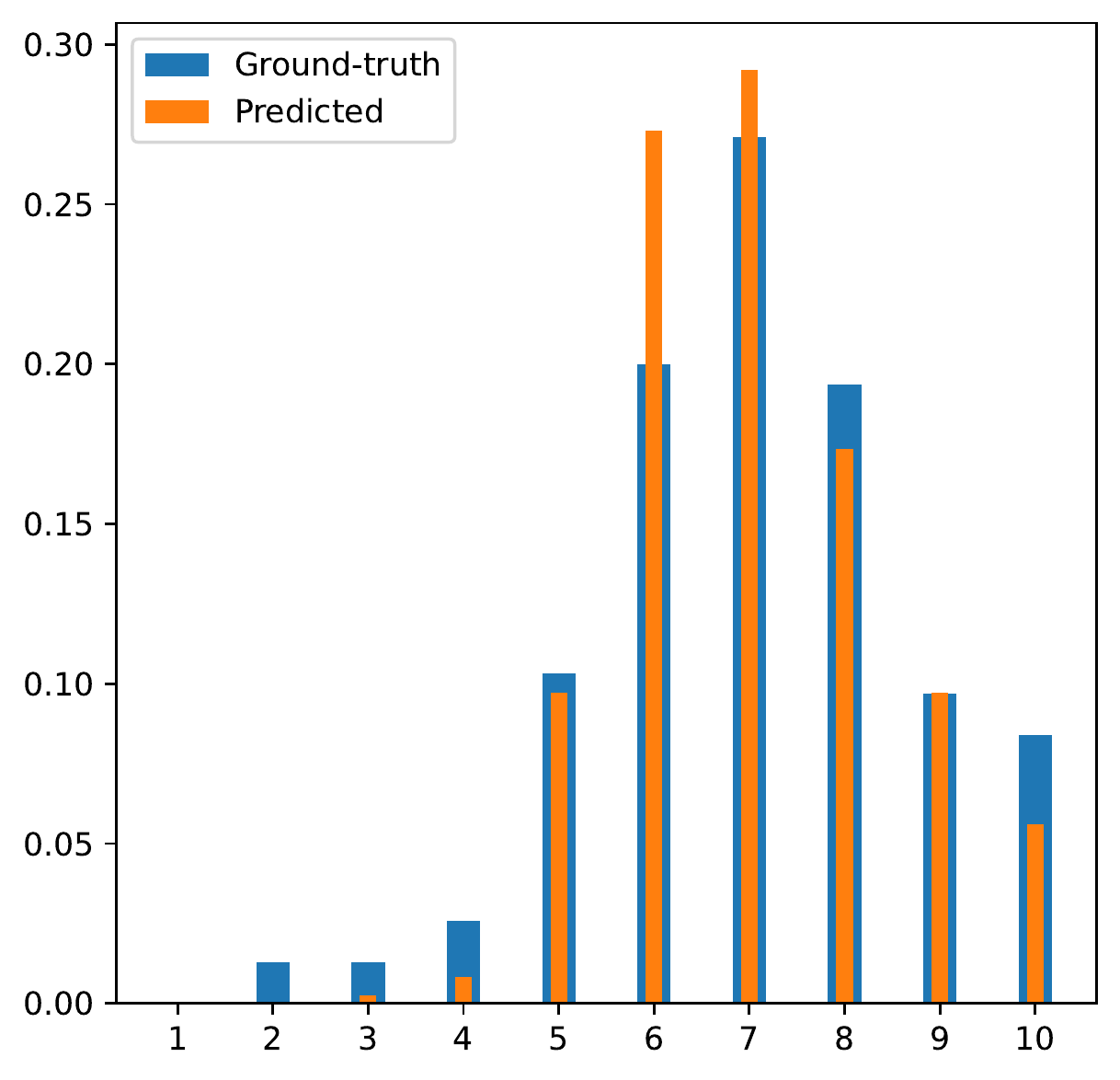} \vspace{1em} \\
    \includegraphics[width=.18\textwidth]{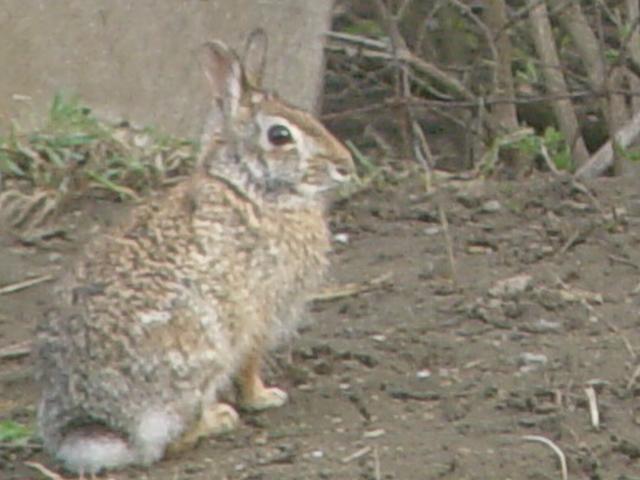} & \includegraphics[width=.14\textwidth]{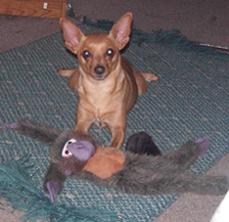} & \includegraphics[width=.18\textwidth]{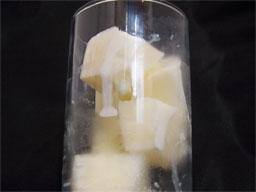} & \includegraphics[width=.18\textwidth]{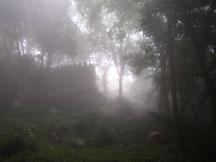} & \includegraphics[width=.18\textwidth]{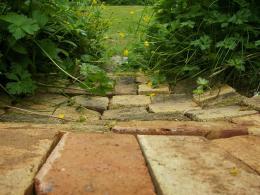} \\
    \includegraphics[width=.18\textwidth]{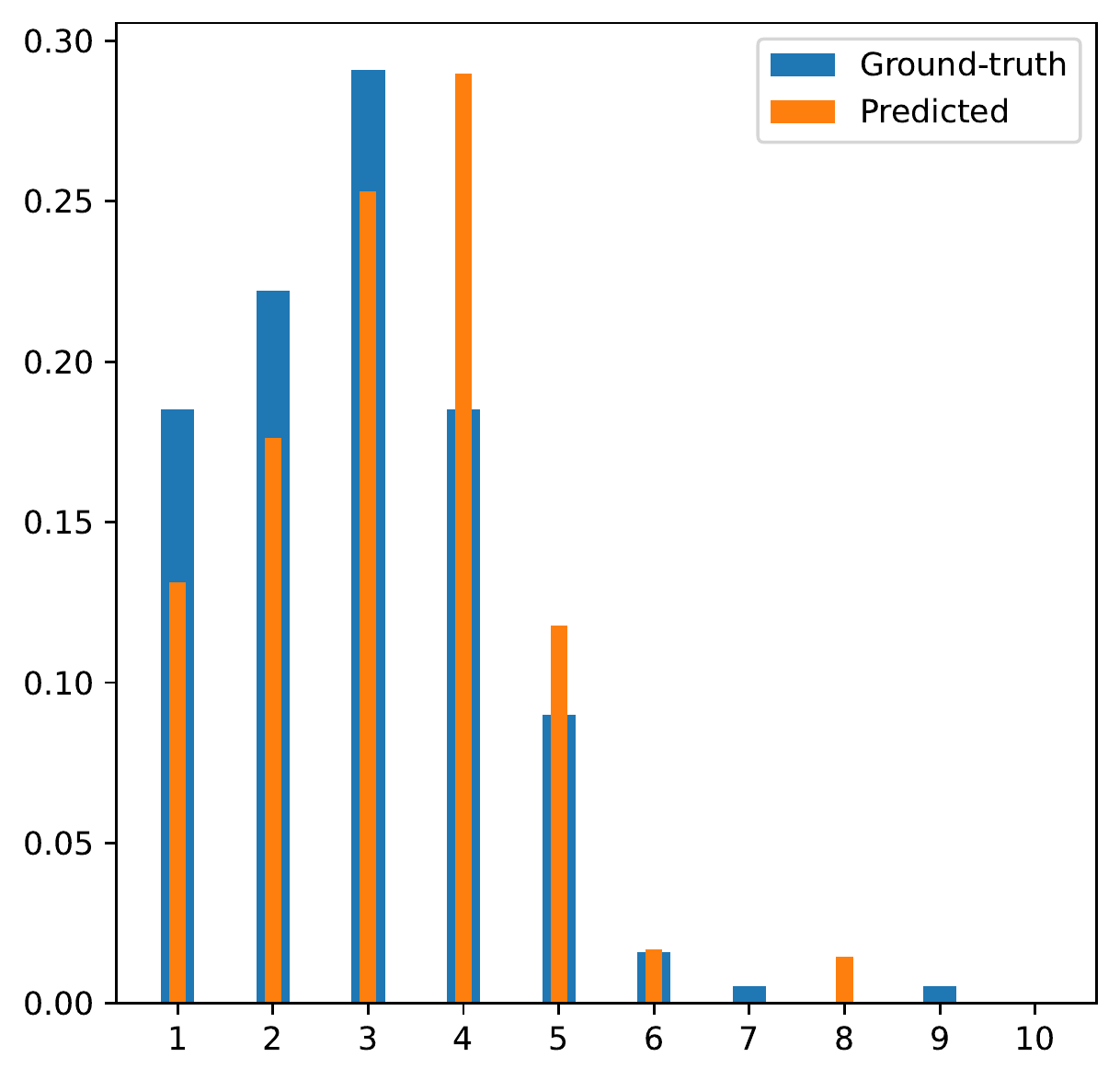} & \includegraphics[width=.18\textwidth]{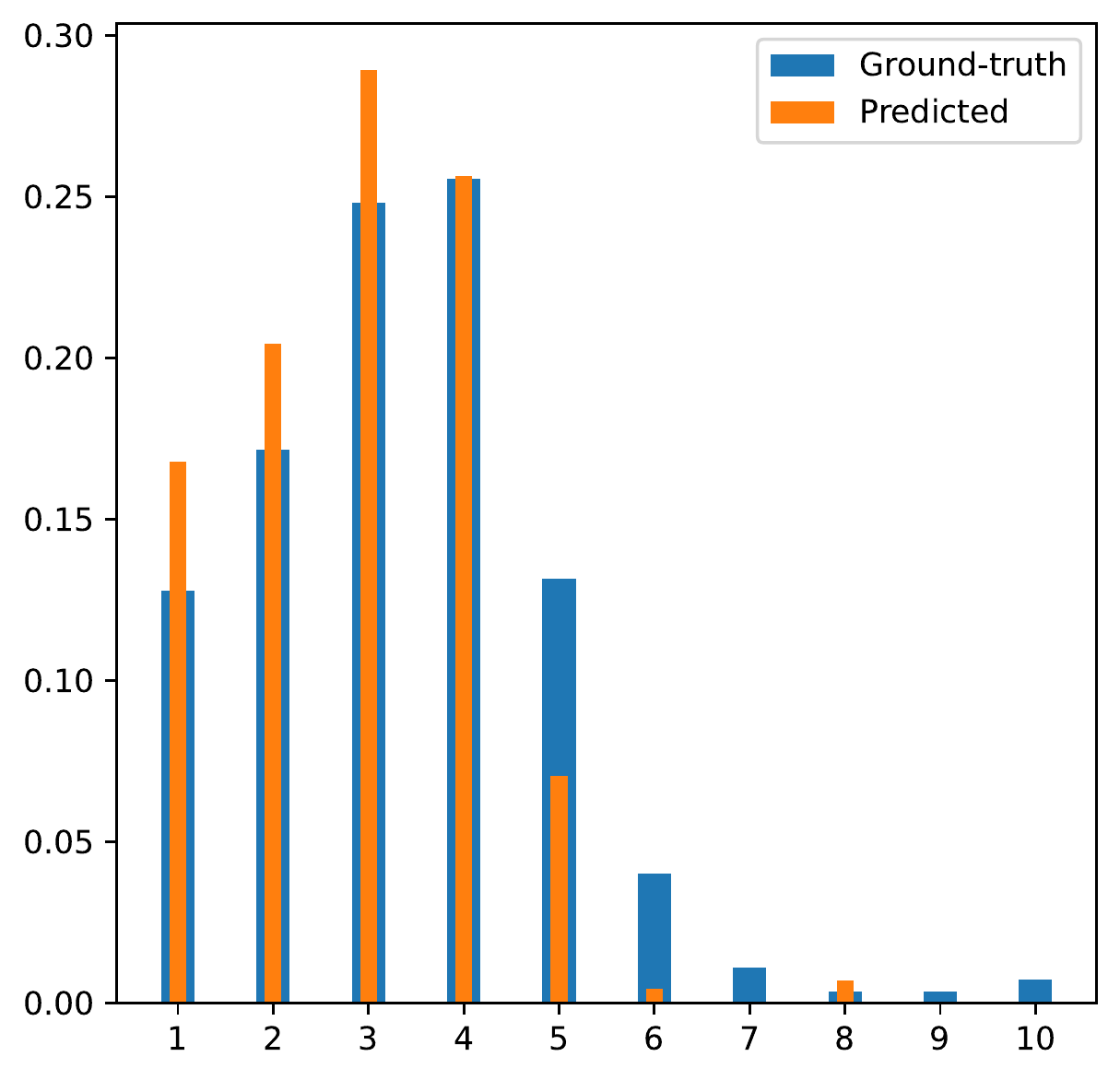} & \includegraphics[width=.18\textwidth]{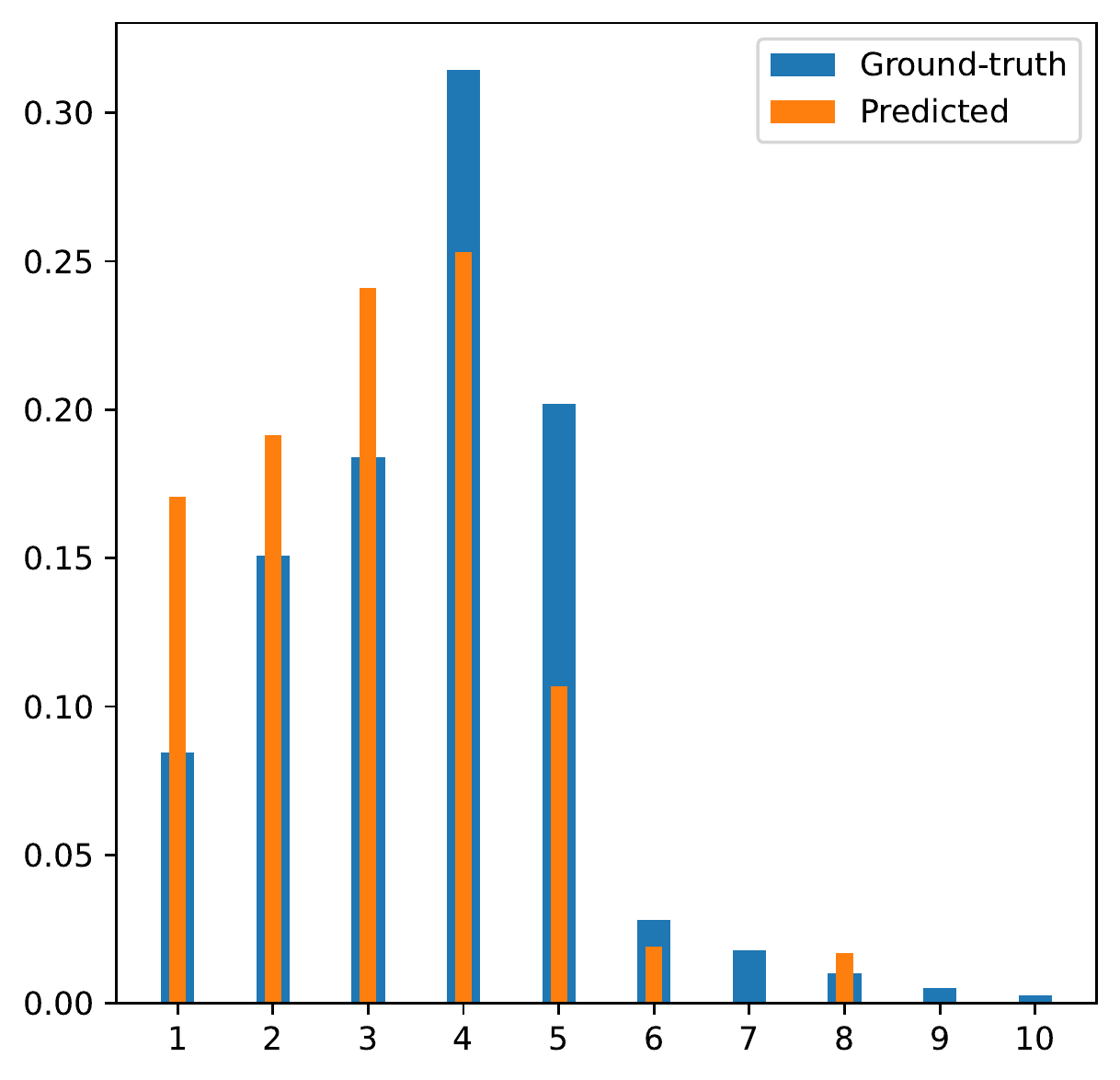} & \includegraphics[width=.18\textwidth]{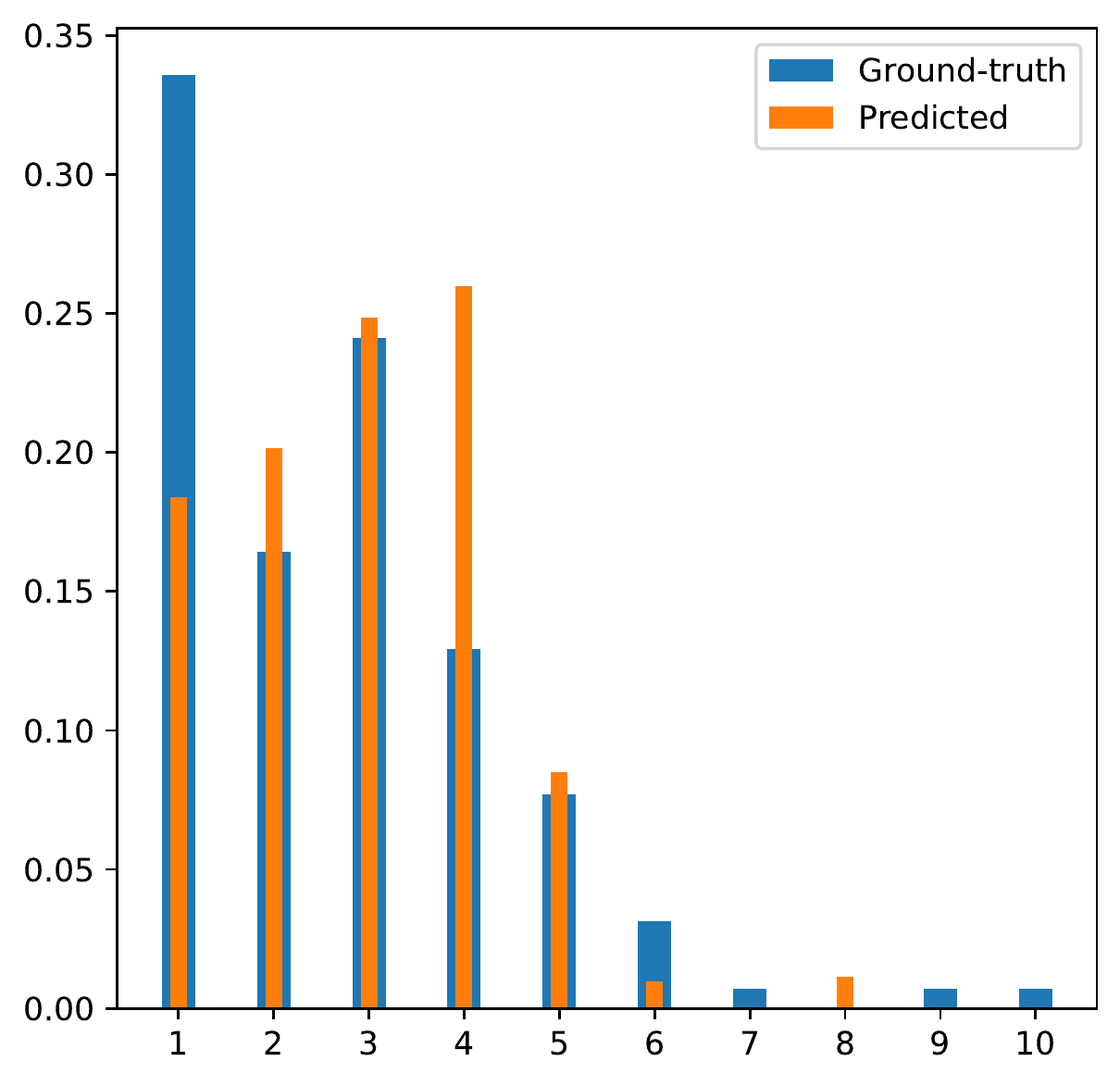} & \includegraphics[width=.18\textwidth]{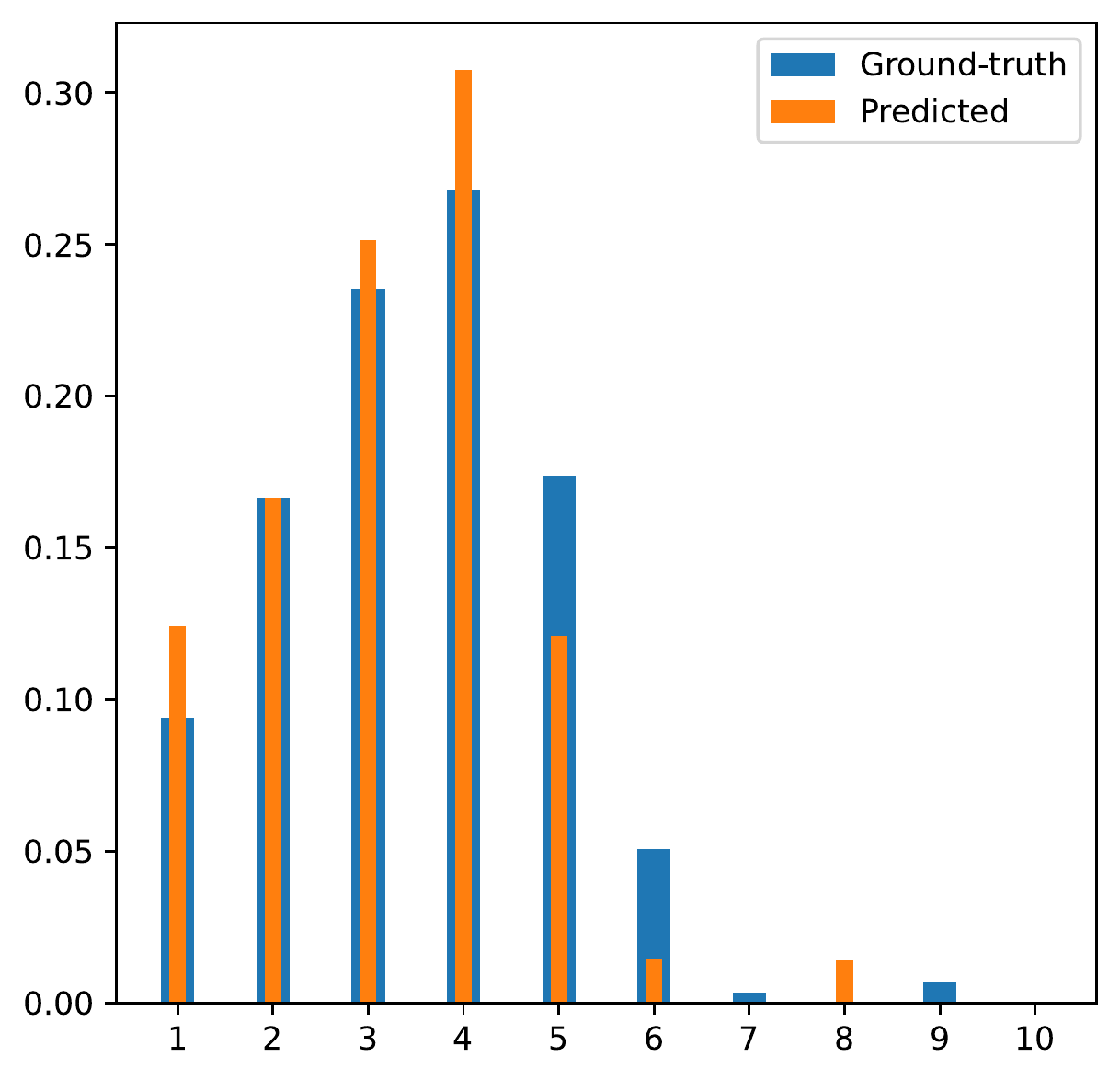}
  \end{tabular}
  \caption{Sample predictions by our method on AVA test images. Top 2 rows: predicted images with high aesthetic quality, coupled with plots of their ground-truth and predicted score distributions. Bottom 2 rows: predicted images with low aesthetic quality, coupled with plots of their ground-truth and predicted score distributions.}
  \label{fig:pred-samples}
\end{figure*}

\begin{figure}
  \centering
  \begin{tabular}{cc}
    \includegraphics[width=.18\textwidth]{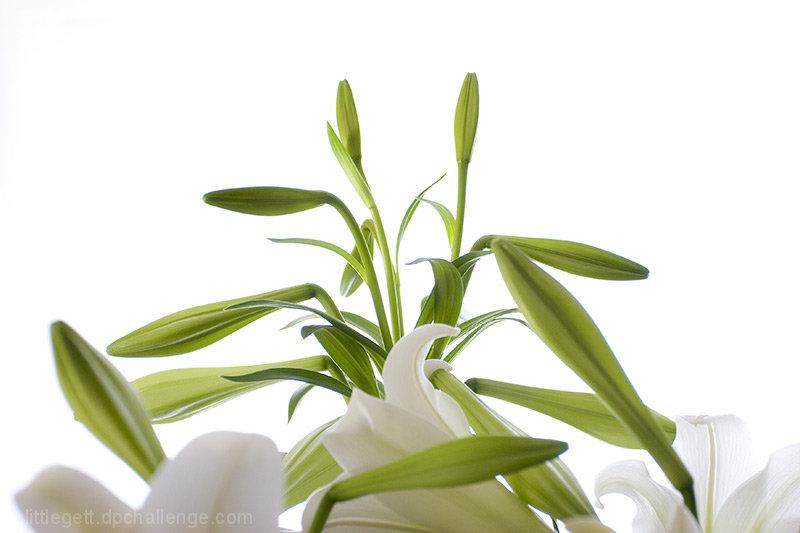} & \includegraphics[width=.18\textwidth]{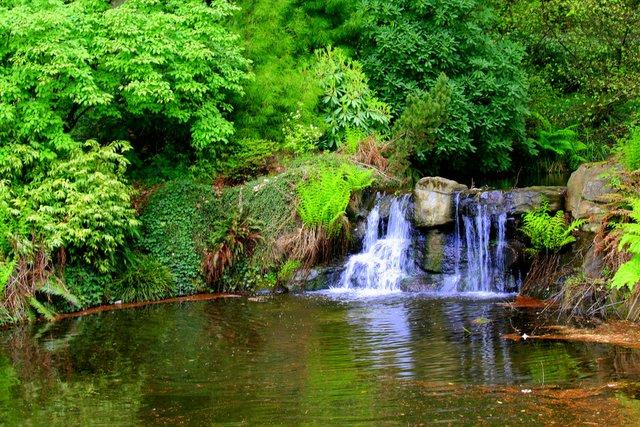} \\
    \includegraphics[width=.18\textwidth]{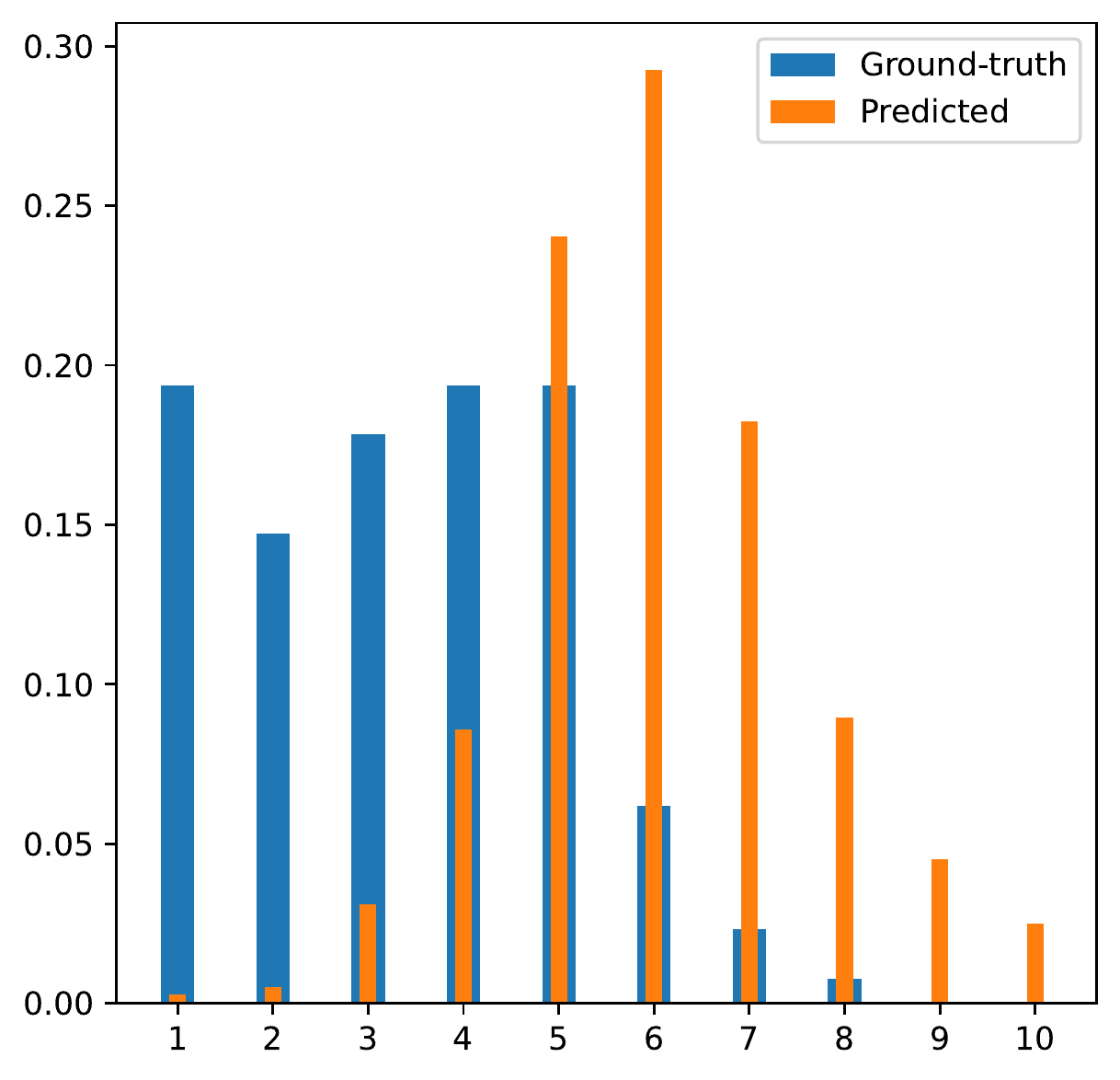} & \includegraphics[width=.18\textwidth]{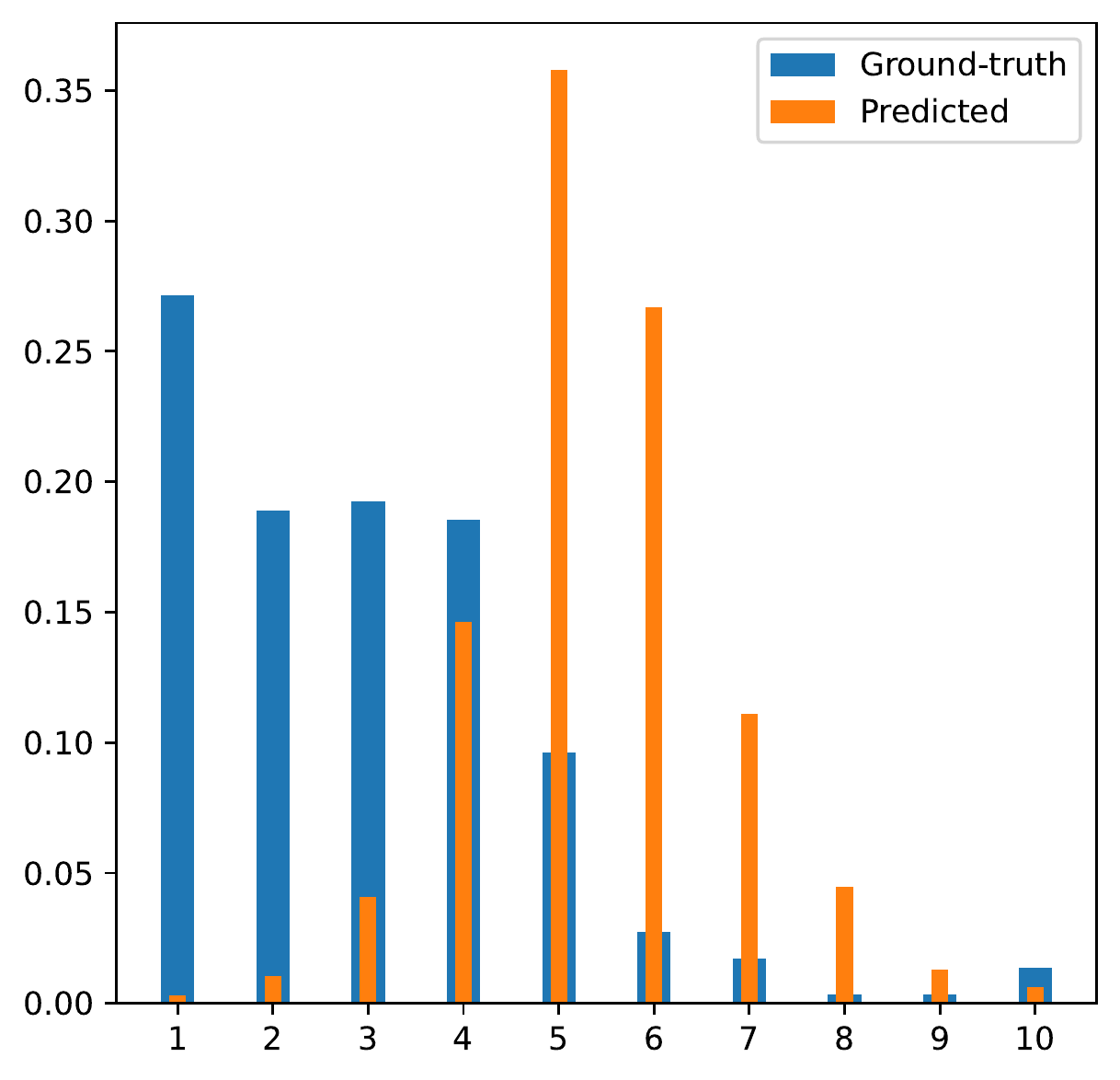}
  \end{tabular}

  \caption{Failure cases of our method on the AVA test set images.}
  \label{fig:failure-samples}
\end{figure}

Figure \ref{fig:pred-samples} shows ten samples from the AVA test set predicted by the proposed method as having high aesthetic quality (the top five images) and low aesthetic quality (the bottom five images), respectively. Plots of the ground-truth and predicted distributions are also shown. As it is possible to see, the model can achieve a high degree of accuracy, with an estimate of the score distribution almost perfect in some cases.
In Figure \ref{fig:failure-samples} there are two examples of failure of our method on the AVA test set images. The method behaves badly on images with not very Gaussian score distributions. This might depend on the fact that 99.77\% of the images in the dataset instead follows a Gaussian distribution \cite{murray2012ava}.
%


Finally, Table \ref{tab:results-photonet} shows the comparison on the Photo.net dataset. The results for state-of-the-art methods are reported by \cite{zhang2019gated}. We underline that the performance is not directly comparable. The evaluation protocols adopted for the methods are different (see the column ``Evaluation protocol'' for details about the adopted protocols). Our performance is comparable to that reported for \textit{GPF-CNN} that is the method achieving results similar to ours.

Several things can be deduced from the results. First, the \textit{Baseline} accuracy is very high (66.58\%) compared to the average obtained by the methods on this dataset. It exceeds that of three methods, i.e. \textit{GIST\_SVM} \cite{marchesotti2011assessing}, \textit{FV\_SIFT\_SVM} \cite{marchesotti2011assessing}, and \textit{MTCNN} \cite{kao2017deep}. Second, the proposed method records a significant improvement on all metrics apart from accuracy compared to \textit{GPF-CNN}. In particular, the SROCC of 0.5650 is 0.04 higher than that of \textit{GPF-CNN}, the MAE is 0.05 lower. Third, the small standard deviation indicates that the proposed method is able to generalize well.
\begin{table*}
  \centering
  \caption{Comparison of the proposed method with state-of-the-art methods on the Photo.net dataset. In each column, the best and second-best results are marked in \textbf{boldface} and \underline{underlined}, respectively. The ``--'' means that the result is not available. The network architecture and whether it uses Multi-Task Learning (MTL) is indicated for each method.}
  \label{tab:results-photonet}
  \resizebox{\textwidth}{!}{\begin{tabular}{lccccccccc}
  \toprule
  Method & Net. arch. & MTL & Accuracy (\%) $\uparrow$ & SROCC $\uparrow$ & PLCC $\uparrow$ & MAE $\downarrow$ & RMSE $\downarrow$ & EMD $\downarrow$ & Evaluation protocol \\ \midrule
  Baseline & & & 66.58 $\pm$ 0.30 & 0.0060 $\pm$ 0.02629 & 0.0053 $\pm$ 0.0285 & 0.4481 $\pm$ 0.0023 & 0.5652 $\pm$ 0.0023 & 0.0789 $\pm$ 0.0004 & 15K train, 1000 val, 1200 test* \\ \midrule
  GIST\_SVM \cite{marchesotti2011assessing} & & & 59.90 & -- & -- & -- & -- & -- & 5-fold cross-validation \\
  FV\_SIFT\_SVM \cite{marchesotti2011assessing} & & & 60.80 & -- & -- & -- & -- & -- & 5-fold cross-validation \\
  MTCNN~\cite{kao2017deep} & VGG16 & \checkmark & 65.20 & -- & -- & -- & -- & -- &  about 15K train, 3000 test \\
  GPF-CNN~\cite{zhang2019gated} & VGG16 & & \textbf{75.60} & \underline{0.5217} & \underline{0.5464} & \underline{0.4242} & \underline{0.5211} & \underline{0.0700} & 15K train, 1000 val, 1200 test \\
  Proposed & EfficinetNet-B4 & \checkmark & \underline{70.05$\pm$0.89} & \textbf{0.5650$\pm$0.0153} & \textbf{0.5698$\pm$0.0141} & \textbf{0.3714$\pm$0.0065} & \textbf{0.4700$\pm$0.0071} & \textbf{0.0689$\pm$0.0009} & 15K train, 1000 val, 1200 test* \\
  \bottomrule
  \multicolumn{7}{l}{* Average and standard deviation on the 10 iterations of train-val-test splits.}
  \end{tabular}}
\end{table*}
%
%
\subsection{Image aesthetic assessment on specific attributes}
In this section, we show how the method works separately for various aesthetic attributes. We select images from the test set of the AVA dataset for the five attributes that are present in AVA and can be discriminated by our method: i.e. Depth-of-field, HDR, Long Exposure, Macro, and Rule of Thirds. The collection consists of about 150 images for each attribute. In each category of images, we systematically estimate the ability of our classifier, namely AttributeNet, to recognize the dominant attribute and of our AestheticNet to estimate aesthetic quality. The experimental results are illustrated in Table \ref{tab:attribute-performance}. These results indicate that the proposed method is able to recognize the attributes even on unseen data, in fact the accuracy performance obtained is very similar to that reported for the same classes in the confusion matrix in Figure \ref{fig:conf-matrix-flick-style}. Macro is easier to recognize with 76.39\% accuracy, while Depth-of-Field is more challenging with a percentage equal to 56.46\%.

By analyzing the aesthetic performance for the five attributes, we notice that the best performance is attained for all the considered metrics for the images belonging to the category HDR. We can also find that the performance of the proposed method significantly drops for images labeled as Depth-of-field. This behavior, which may be due to our method failing to categorize images well for that attribute, indicates that our aesthetic assessment process is indeed attribute-driven.
\begin{table*}
    \centering
    \caption{Performance on five aesthetic-related attributes of the AVA dataset.}
    \label{tab:attribute-performance}
    \begin{tabular}{lccccccc}
        \toprule
        \multirow{2}{*}{Attribute} & Attribute recognition & \multicolumn{6}{c}{Aesthetic assessment} \\
        & Acc. (\%) $\uparrow$ & Acc. (\%) $\uparrow$ & SROCC $\uparrow$ & PLCC $\uparrow$ & MAE $\downarrow$ & RMSE $\downarrow$ & EMD $\downarrow$\\ \midrule
        Depth-of-field & 56.46 & 79.31 & 0.7320 & 0.7575 & 0.4129 & 0.5006 & 0.0448 \\
        HDR & 59.52 & 92.86 & 0.7944 & 0.8007 & 0.3110 & 0.3902 & 0.0347 \\
        Long Exposure & 65.25 & 81.35 & 0.6506 & 0.7741 & 0.4025 & 0.4985 & 0.0435 \\
        Macro & 76.39 & 82.32 & 0.7941 & 0.7616 & 0.3528 & 0.4518 & 0.0408 \\
        Rule of Thirds & 64.52 & 80.29 & 0.6942 & 0.6732 & 0.3757 & 0.4804 & 0.0406 \\
        \bottomrule
    \end{tabular}
\end{table*}
\subsection{Ablation study}
\subsubsection{Effectiveness of the designed model architecture}
Unlike traditional methods and architectures, the use of an attribute-conditioned hypernetwork in the proposed method allows to have a dedicated aesthetic evaluator for each test image that depends on the composition and style attributes it contains. At the same time the hypernetwork adds an extra compute load to the network to generate the AestheticNet parameters. Here, we present two experiments aimed at justifying the design choices of the proposed architecture. In the first experiment we validate the effectiveness of the hypernetwork. We compare the performance obtained by the proposed method with that of a simplified version of our proposal without hypernetwork. Specifically, we remove the HyperNet, and the embedding of the AttributeNet $\mathbf{e}_s$ is mapped to the aesthetic score distribution through a linear layer. We follow the same training procedure presented in Section \ref{sec:train-proc} for a fair comparison. The second experiment is aimed at demonstrating the importance of guiding the aesthetic prediction by exploiting composition and style attributes. For this purpose, we perform a surgical operation to the proposed method in which we remove AttributeNet and HyperNet. The embedding vector $e_b$ estimated by the Backbone is mapped to an aesthetic score distribution using the AestheticNet's MLP, whose weights are learned on the aesthetic dataset.

Table \ref{tab:ablation} reports the results of the ablation study on the AVA test images. In the ``AttrNet + AesthNet (Linear)'' row we show the results for the solution with a linear layer as AestheticNet, while in the ``AesthNet (MLP)'' there are the results of the Backbone followed by an MLP as AestheticNet. As can be seen, both variants of the method achieve worse results than the proposed method. In detail, the drop obtained by the ``AttrNet + AesthNet (Linear)'' variant is 2\% in terms of accuracy and 0.07 for the SROCC. The fact that the drop obtained by ``AesthNet (MLP)'' compared to our best solution is even greater (i.e., 3\% less for binary classification and 0.08 for both PLCC and SROCC) experimentally supports the choice of having characterized the attributes and the adopted design choices.
\begin{table*}
    \centering
    \caption{Ablation study results on the AVA test set. Each row reports the results for a modified version of the proposed method. The Backbone is present in all variants but is omitted for simplicity.}
    \label{tab:ablation}
    \begin{tabular}{lcccccc}
    \toprule
    Method & Acc. (\%) $\uparrow$ & SROCC $\uparrow$ & PLCC $\uparrow$ & MAE $\downarrow$ & RMSE $\downarrow$ & EMD $\downarrow$ \\ \midrule
    AttrNet + AesthNet (Linear) & 78.59 & 0.6600 & 0.6633 & 0.4420 & 0.5652 & 0.0481 \\
    AesthNet (MLP) & 77.65 & 0.6546 & 0.6559 & 0.4529 & 0.5875 & 0.0496 \vspace{0.3em} \\
    AttrNet (only style) + HyperNet + AesthNet & 79.78 & 0.7269 & 0.7302 & 0.4115 & 0.5243 & 0.0475 \\
    AttrNet (only comp.) + HyperNet + AesthNet & 80.23 & 0.7295 & 0.7312 & 0.4092 & 0.5197 & 0.0461 \\
    AttrNet (on AVA) + HyperNet + AesthNet & 78.76 & 0.6676 & 0.6714 & 0.4360 & 0.5595 & 0.0485 \vspace{0.3em} \\
    AttrNet + HyperNet + AesthNet & 80.75 & 0.7318 & 0.7329 & 0.4011 & 0.5128 & 0.0439 \\ \bottomrule
    \end{tabular}
\end{table*}
\begin{figure*}
    \centering
    \resizebox{\linewidth}{!}{
    \begin{tabular}{ccccccc}
        & \makecell{\small AttrNet +\\ AesthNet (Linear)} & \makecell{\small AesthNet (MLP)} & \makecell{\small AttrNet (only style) +\\ HyperNet + AesthNet} & \makecell{\small AttrNet (only comp.) + \\ HyperNet + AesthNet} & \makecell{\small AttrNet (on AVA) +\\ HyperNet + AesthNet} & \makecell{\small AttrNet + \\ HyperNet + AesthNet} \\
        \raisebox{+0.25\height}{\includegraphics[width=.2\linewidth]{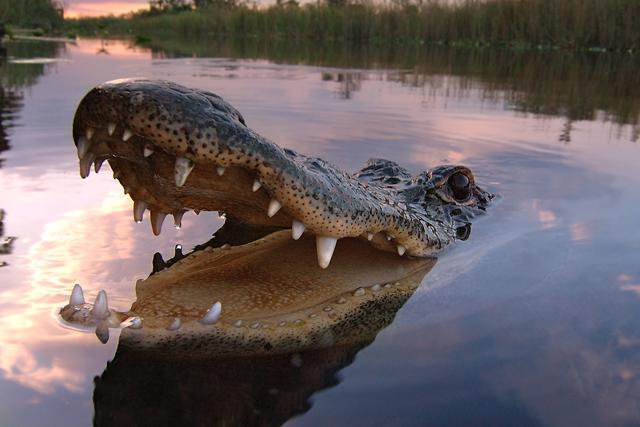}} & \includegraphics[width=.2\linewidth]{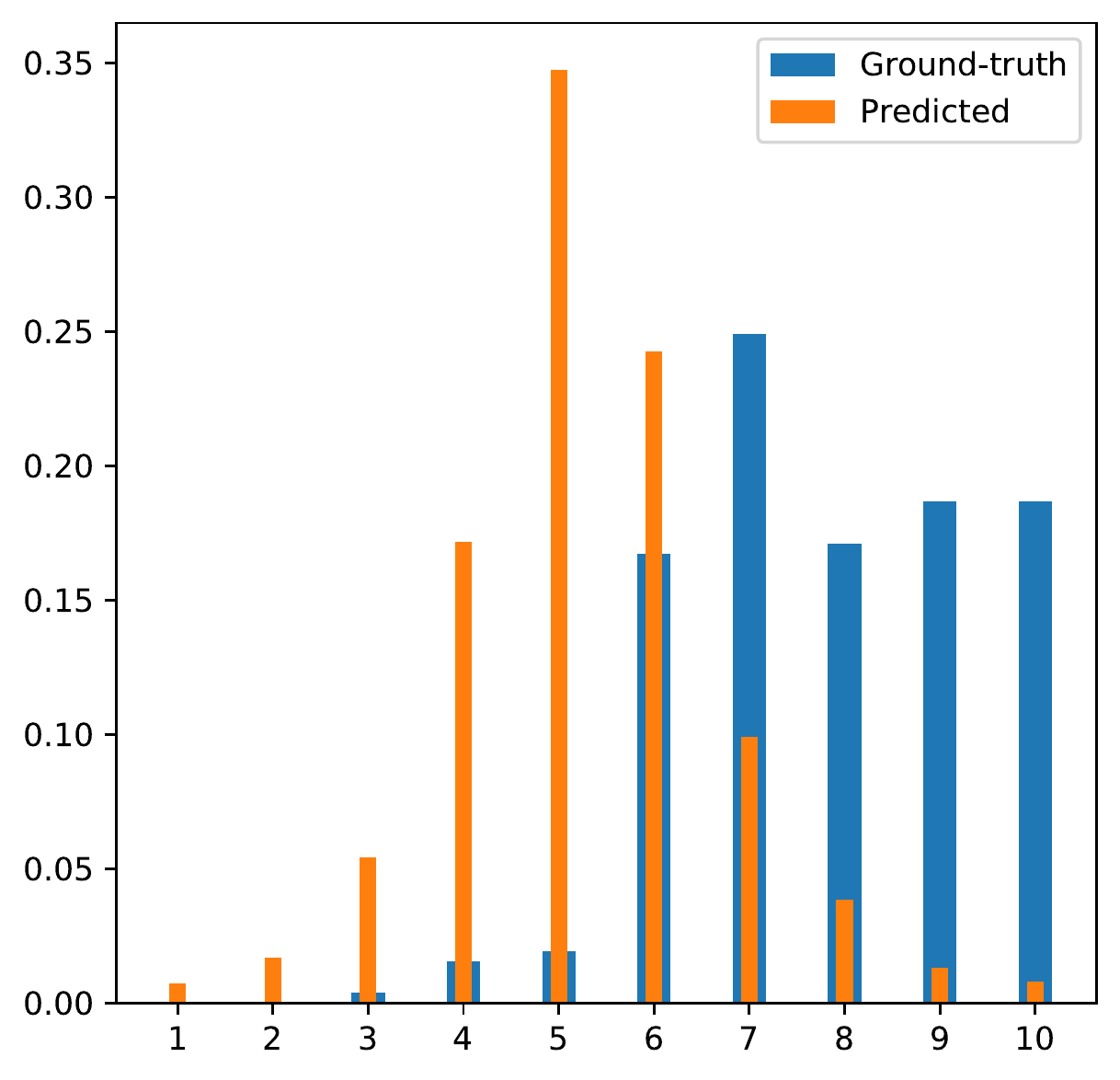} & \includegraphics[width=.2\linewidth]{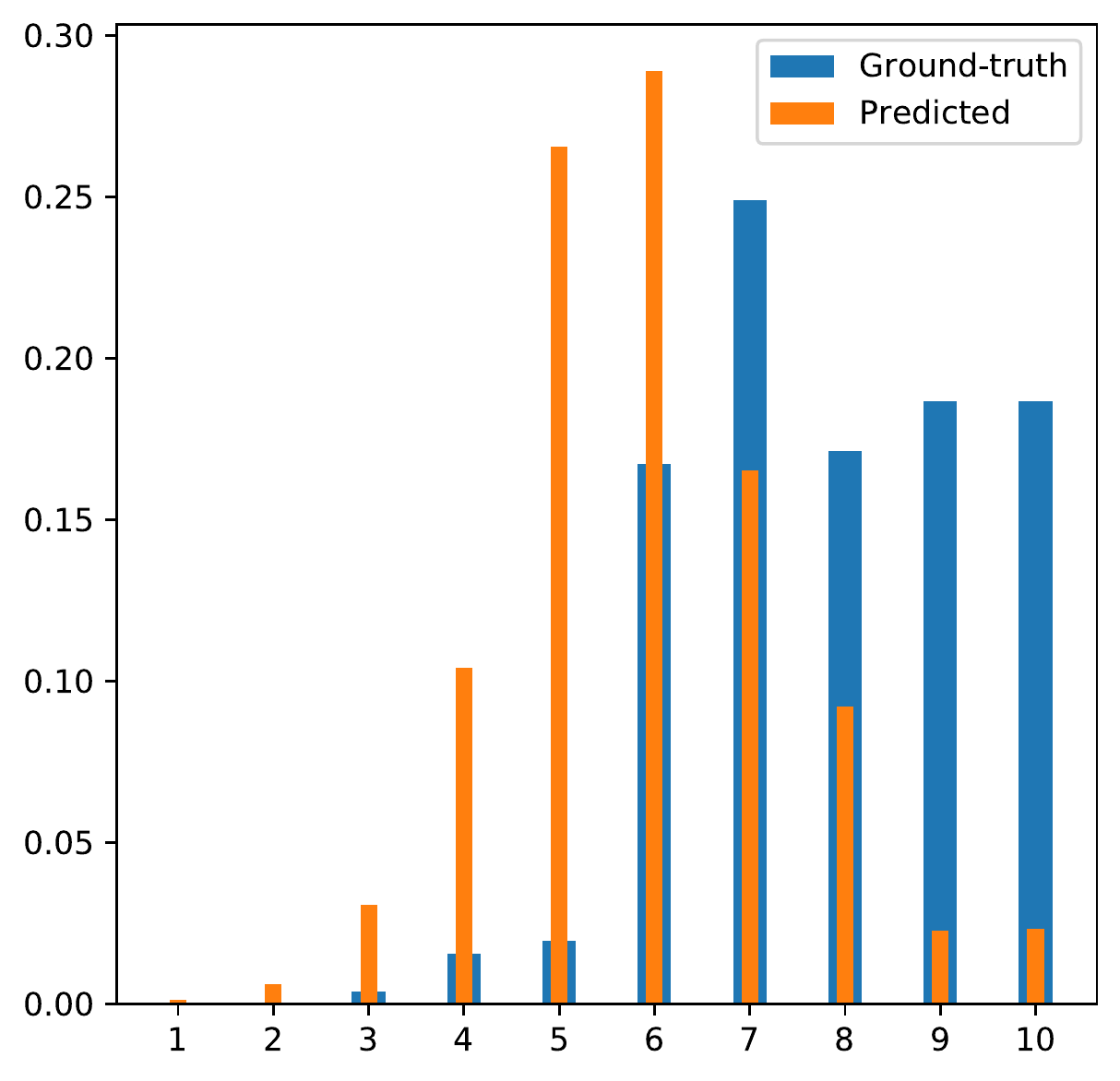} & \includegraphics[width=.2\linewidth]{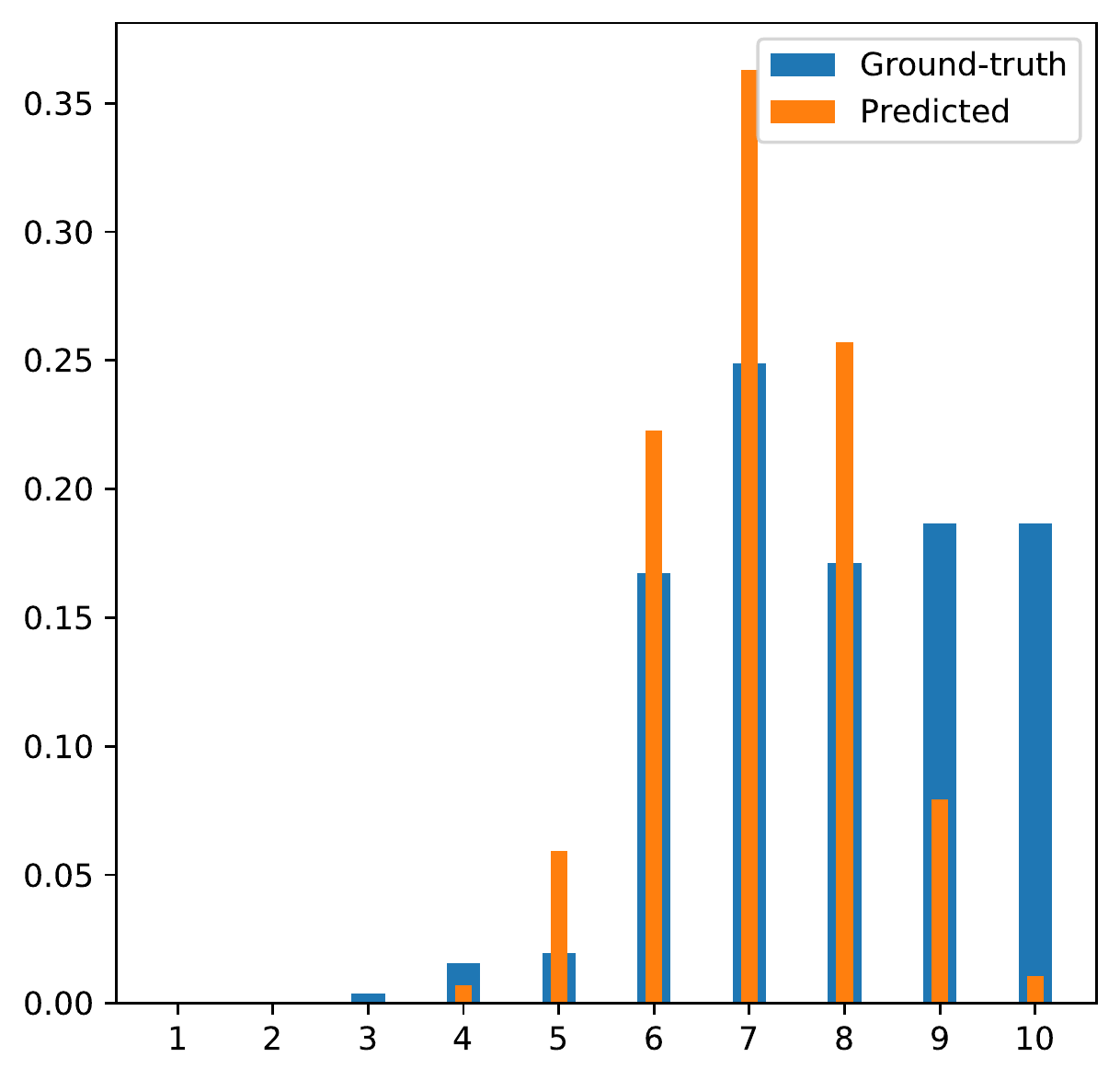} & \includegraphics[width=.2\linewidth]{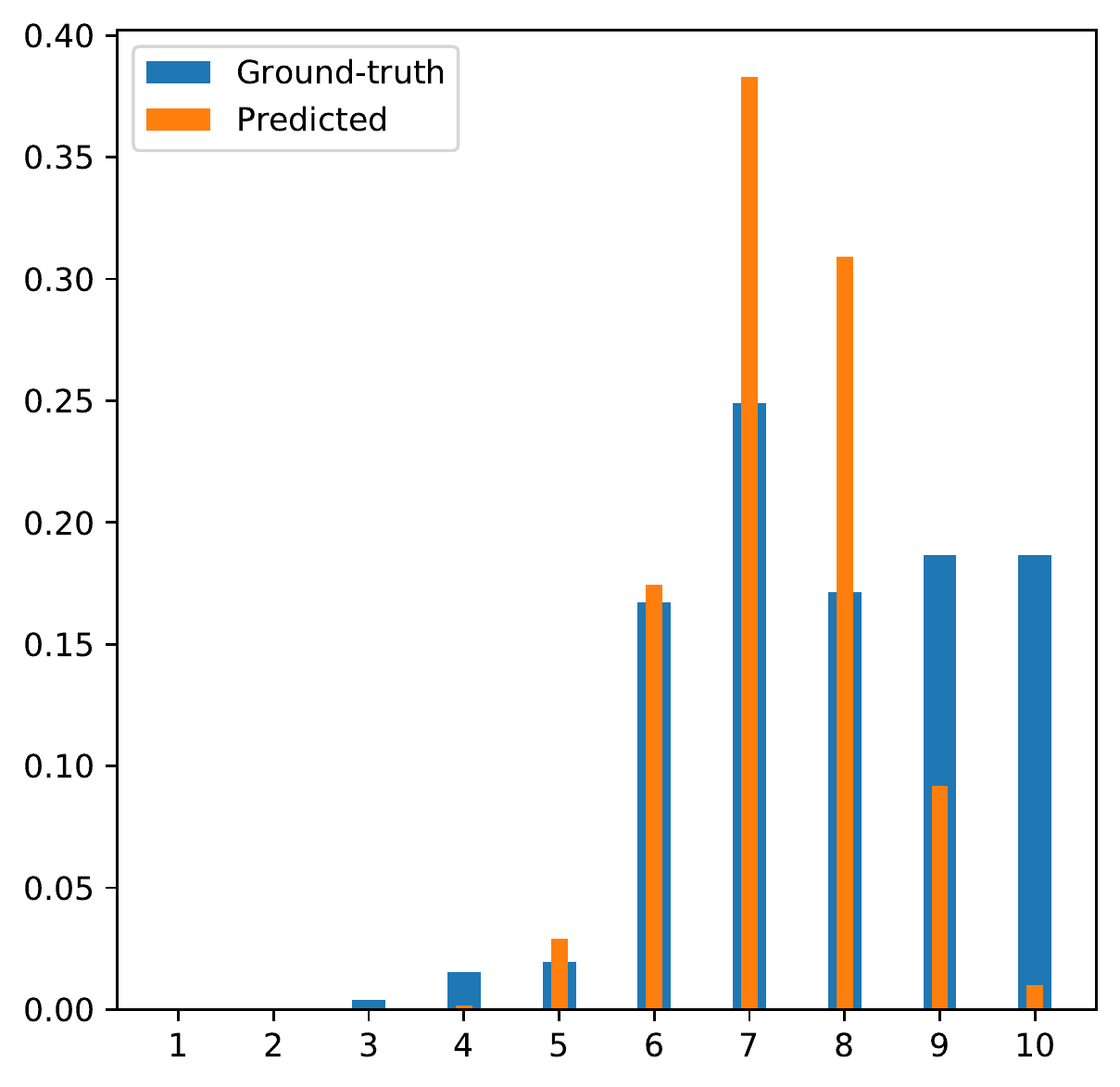} & \includegraphics[width=.2\linewidth]{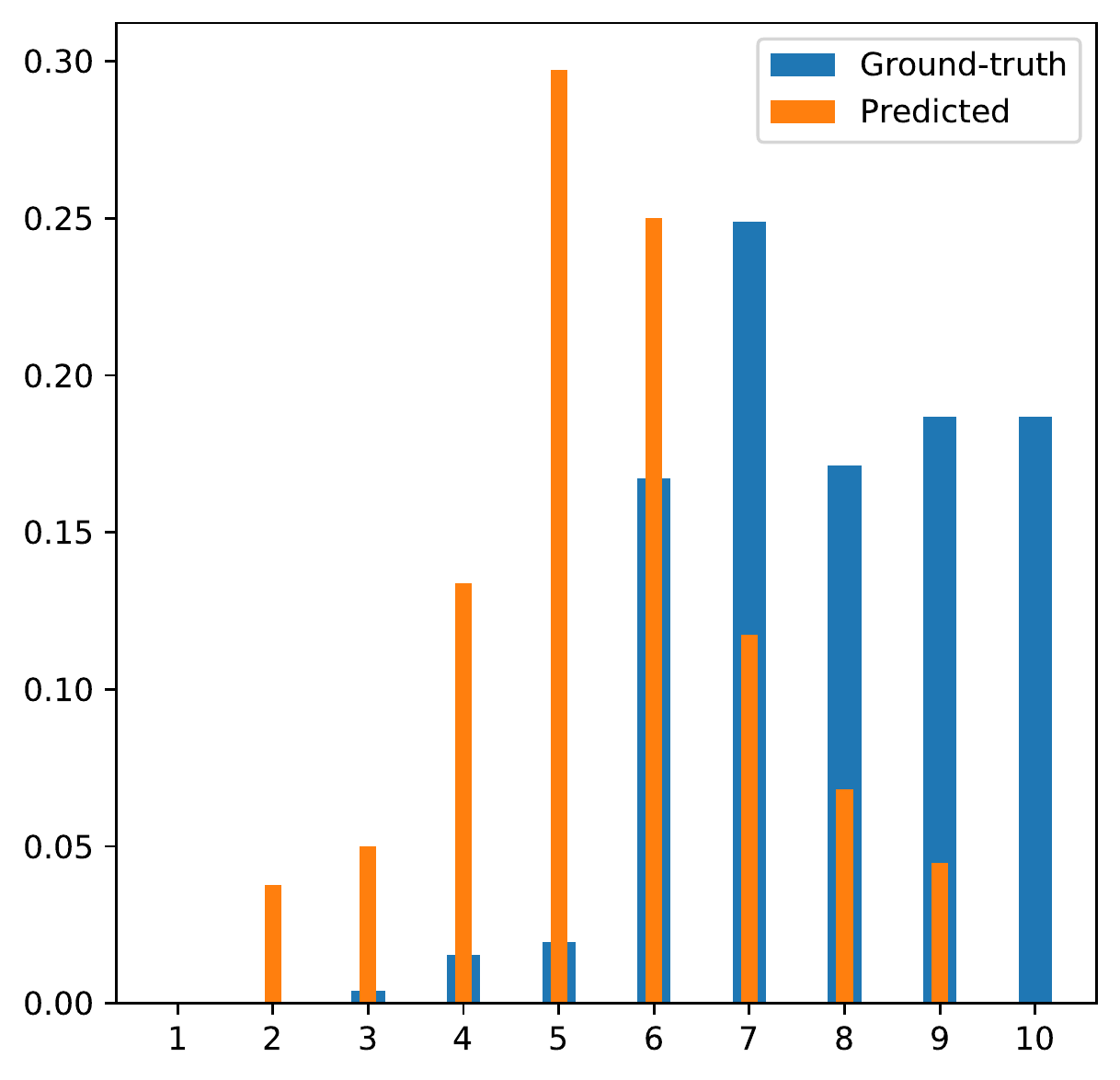} & \includegraphics[width=.2\linewidth]{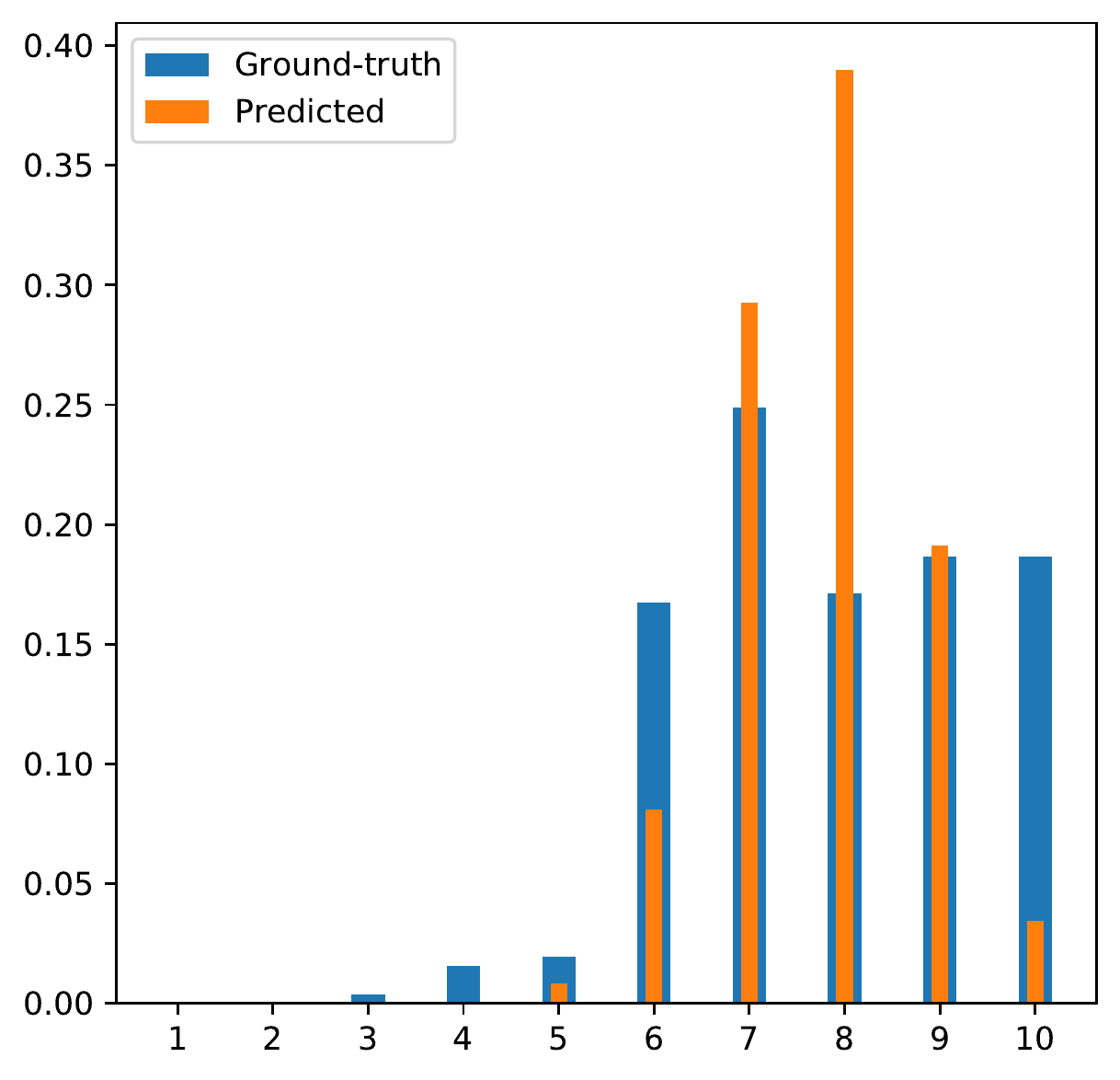} \\
        & (7.8366) 5.2902 & (7.8366) 5.9148 & (7.8366) 6.9225 & (7.8366) 7.0738 & (7.8366) 5.5218 & (7.8366) 7.1253 \vspace{1em}\\
        \raisebox{+0.07\height}{\includegraphics[width=.2\linewidth]{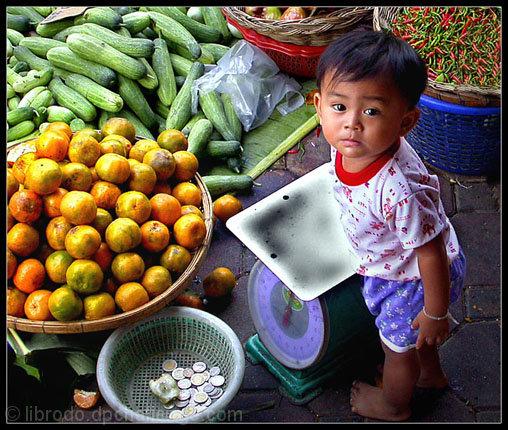}} & \includegraphics[width=.2\linewidth]{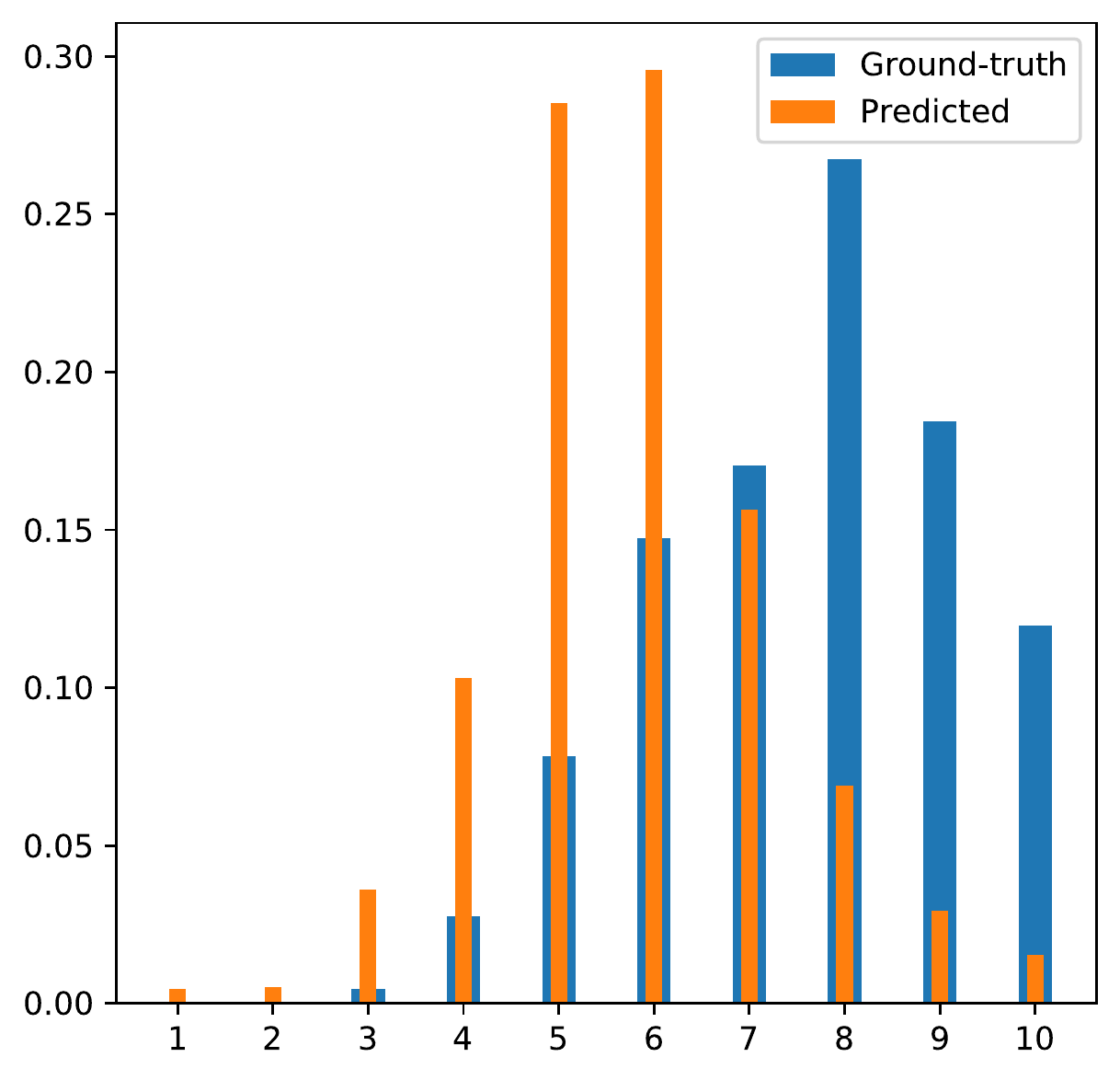} &
        \includegraphics[width=.2\linewidth]{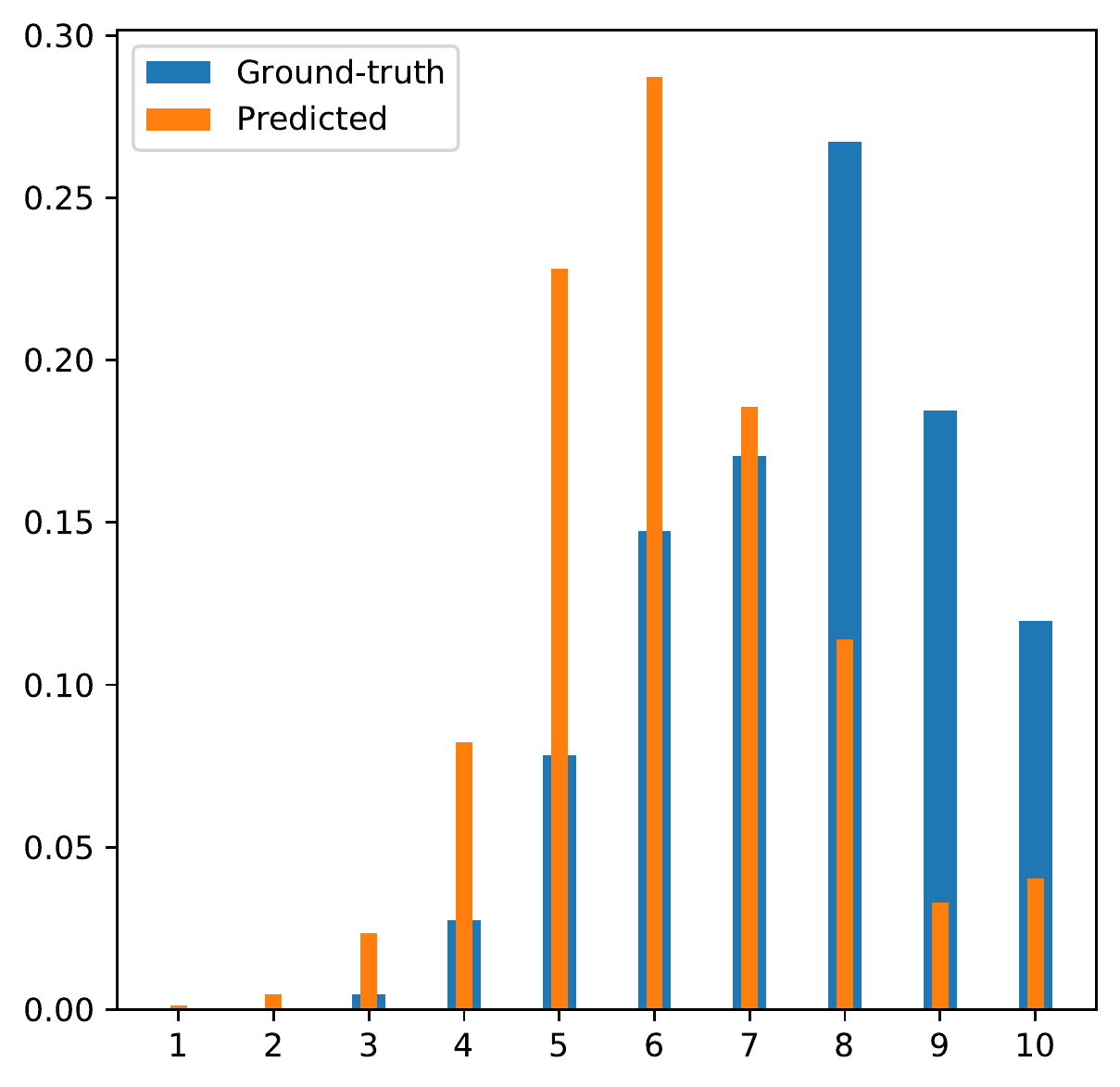} &
        \includegraphics[width=.2\linewidth]{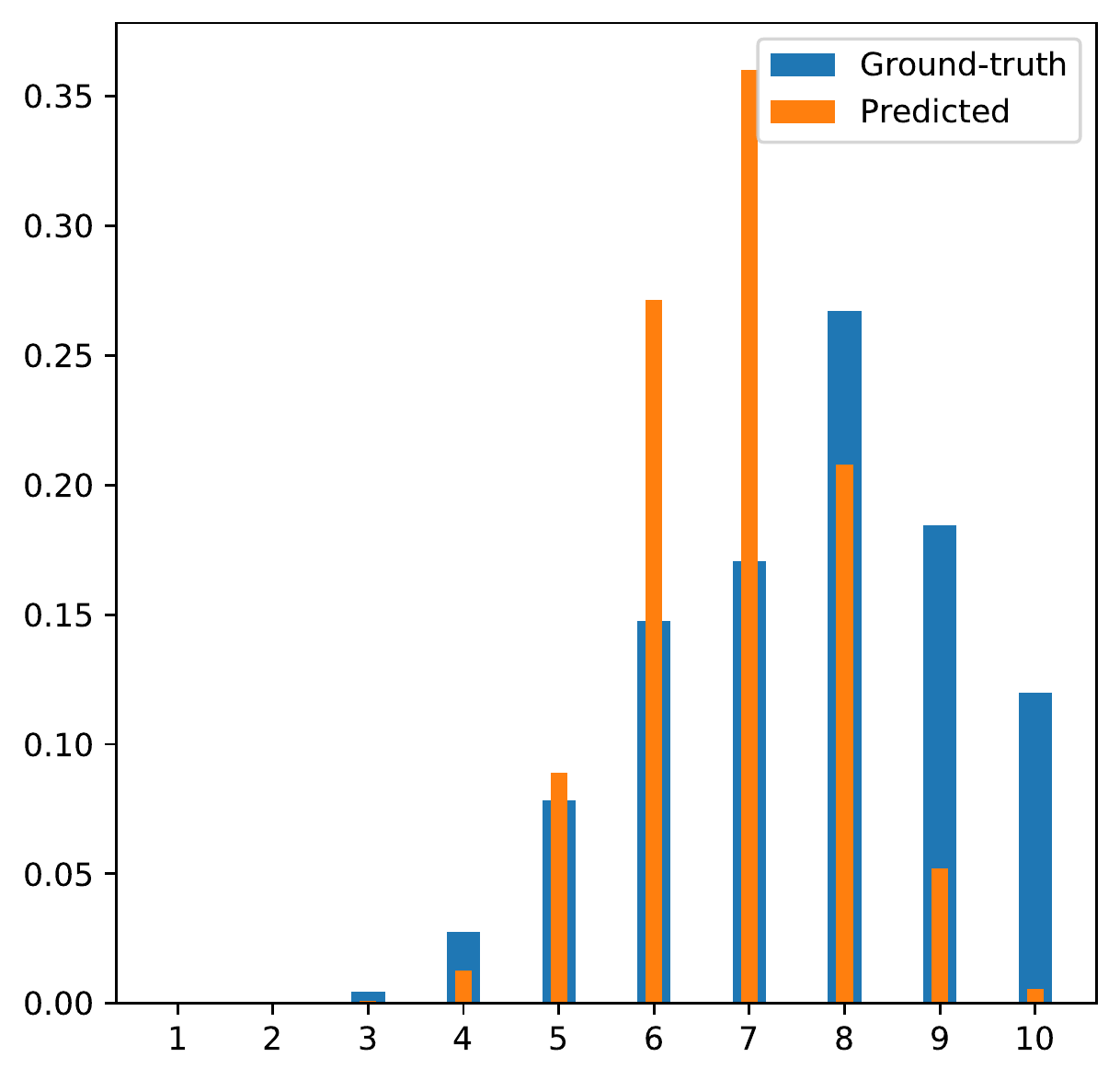} &
        \includegraphics[width=.2\linewidth]{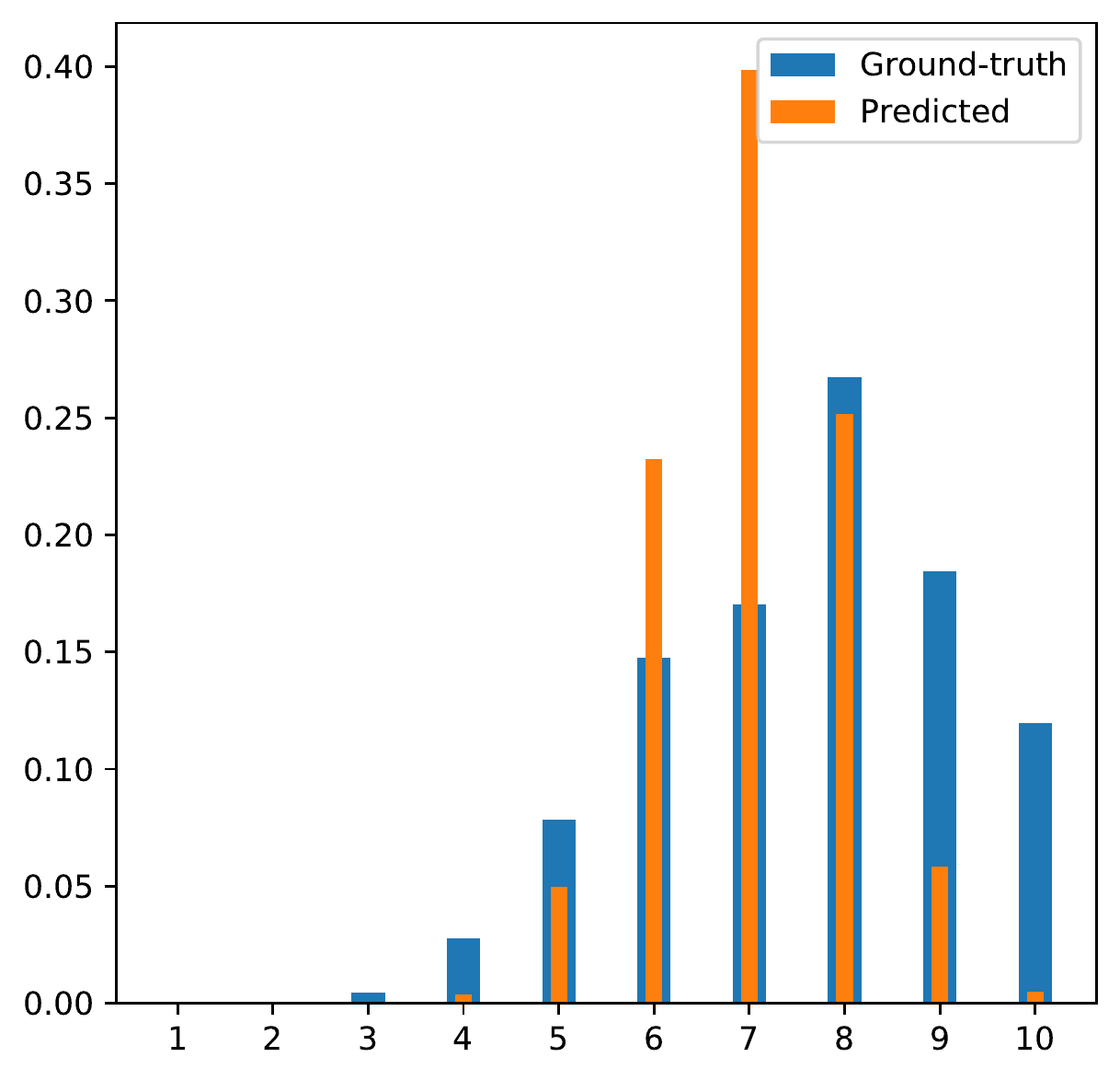} &
        \includegraphics[width=.2\linewidth]{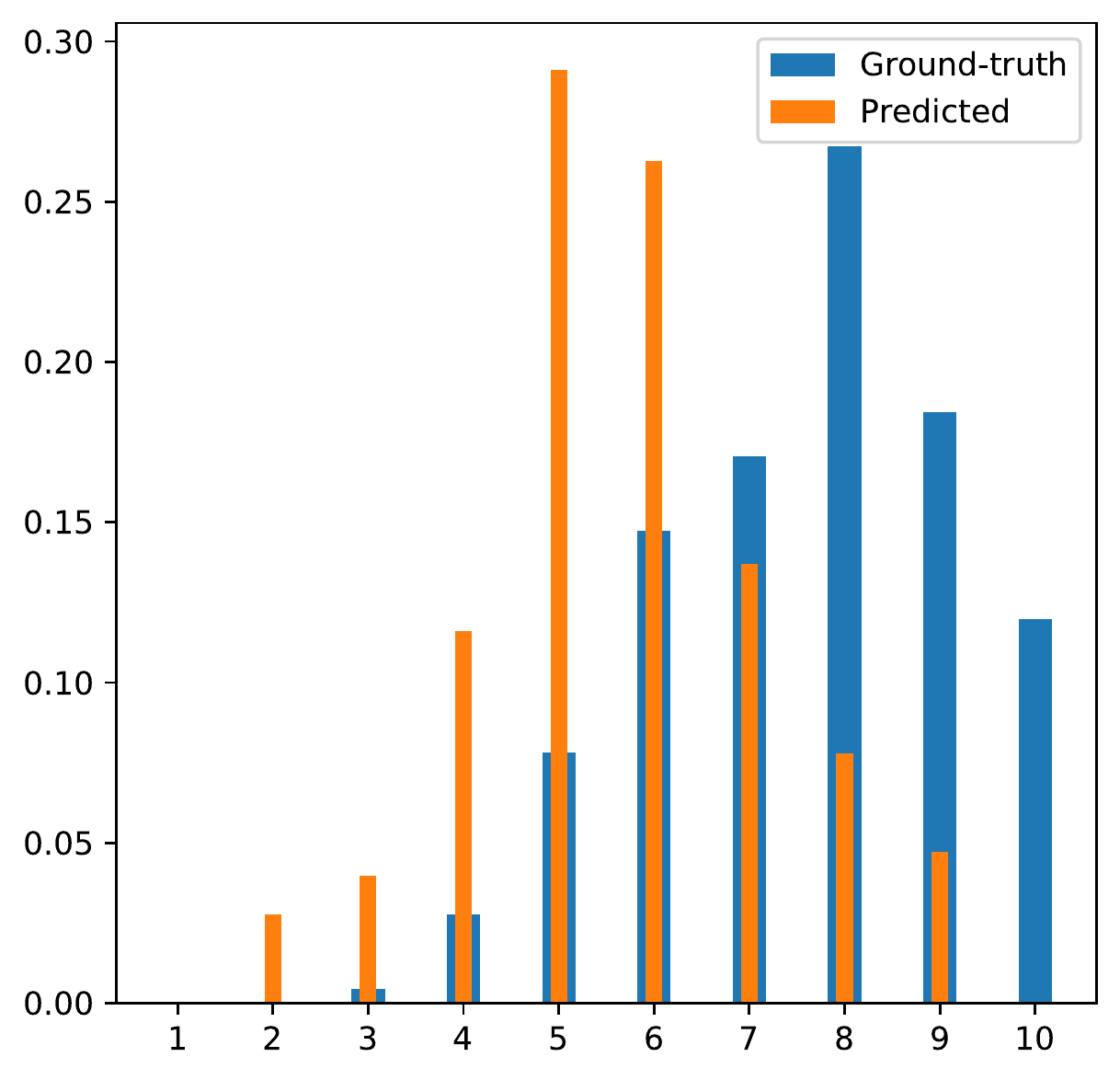} & \includegraphics[width=.2\linewidth]{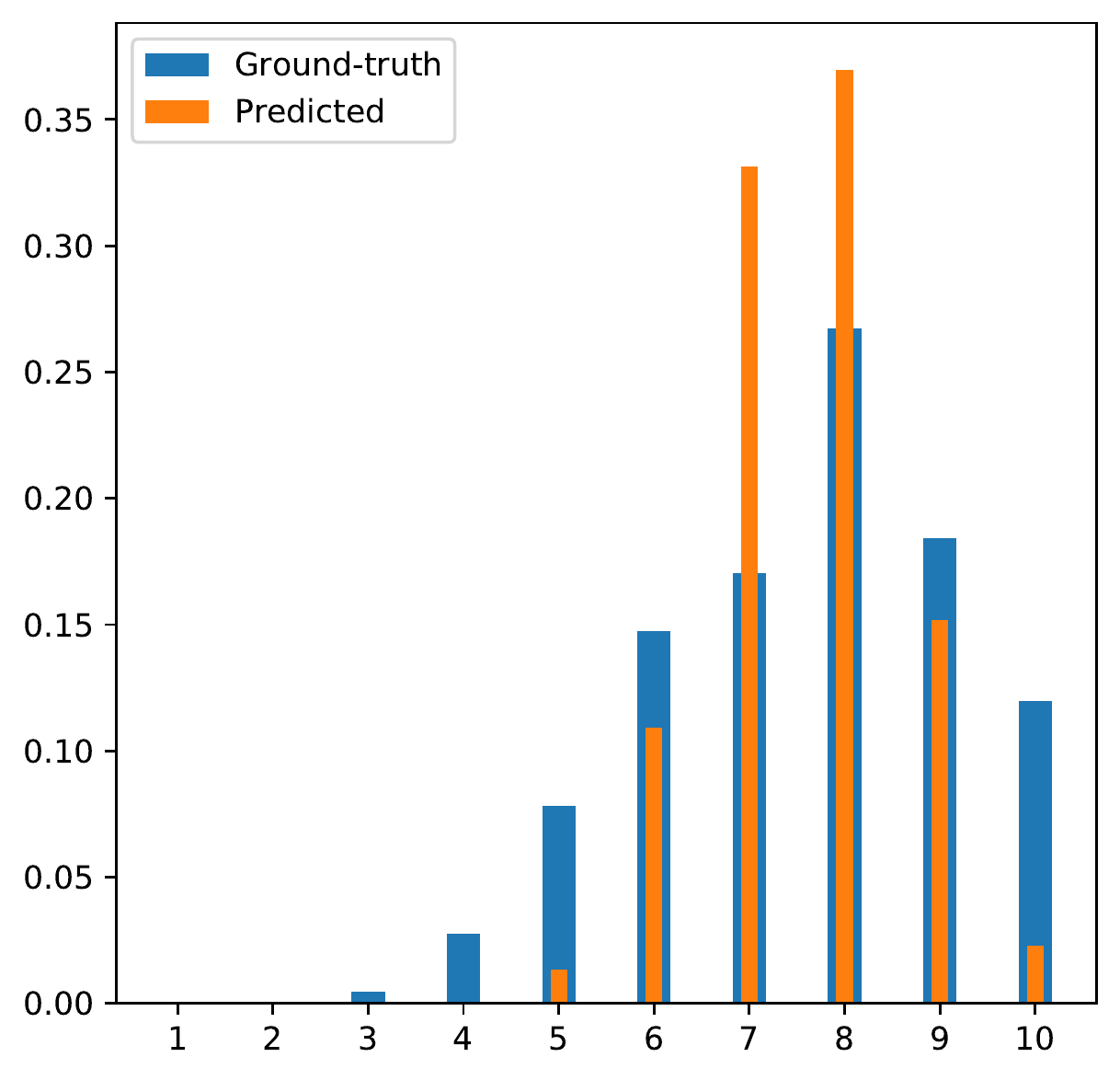} \\
        & (7.5899) 5.8005 & (7.5899) 6.1854 & (7.5899) 7.0554 & (7.5899) 7.1269 & (7.5899) 6.8810 & (7.5899) 7.2272 \vspace{1em}\\
        \raisebox{+0.25\height}{\includegraphics[width=.2\linewidth]{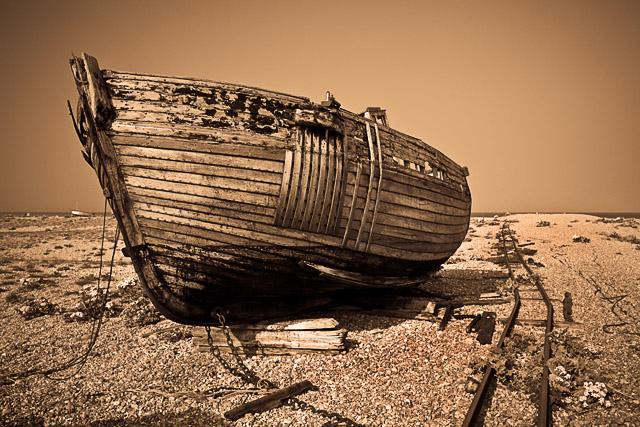}} & \includegraphics[width=.2\linewidth]{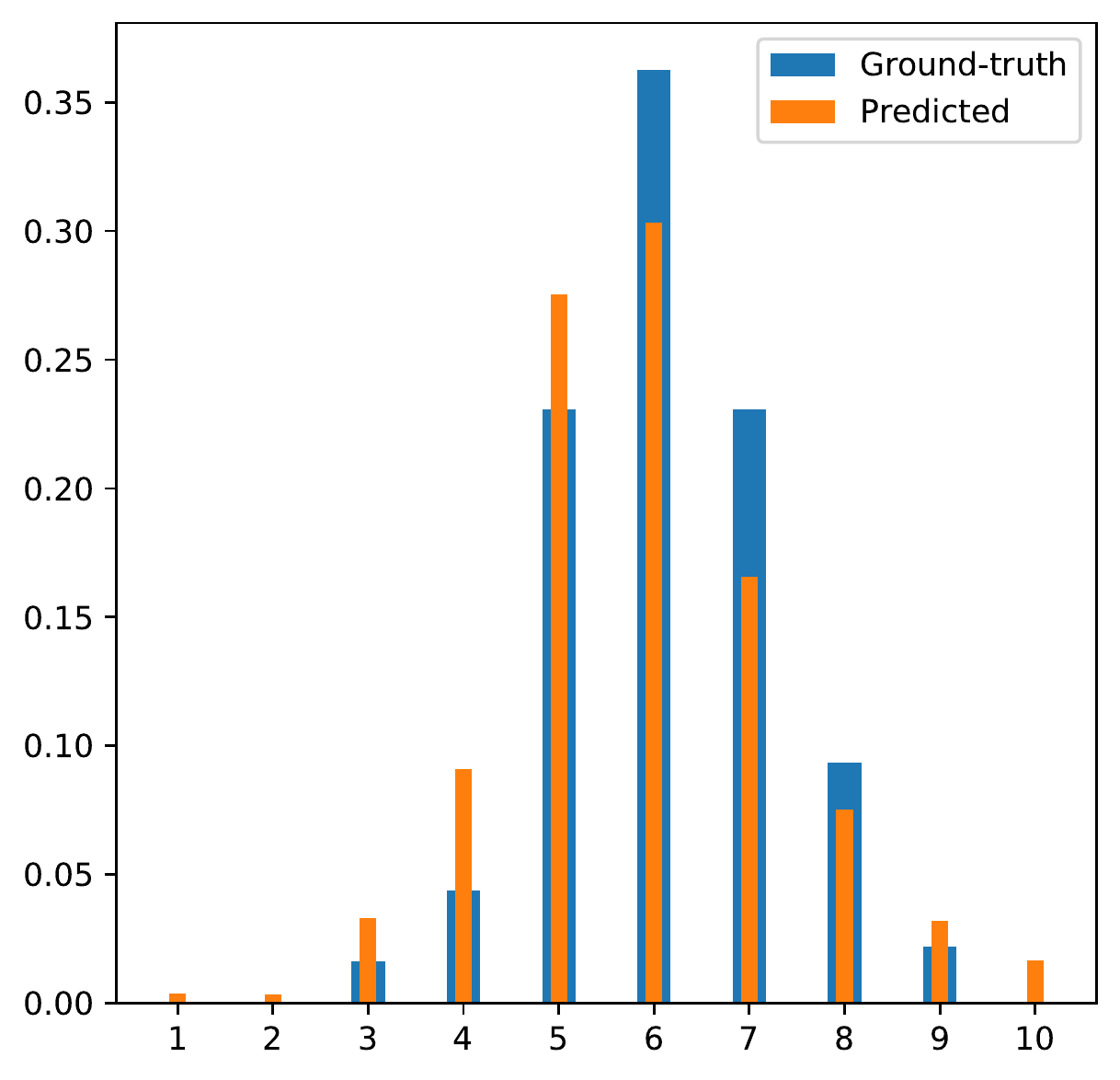} &
        \includegraphics[width=.2\linewidth]{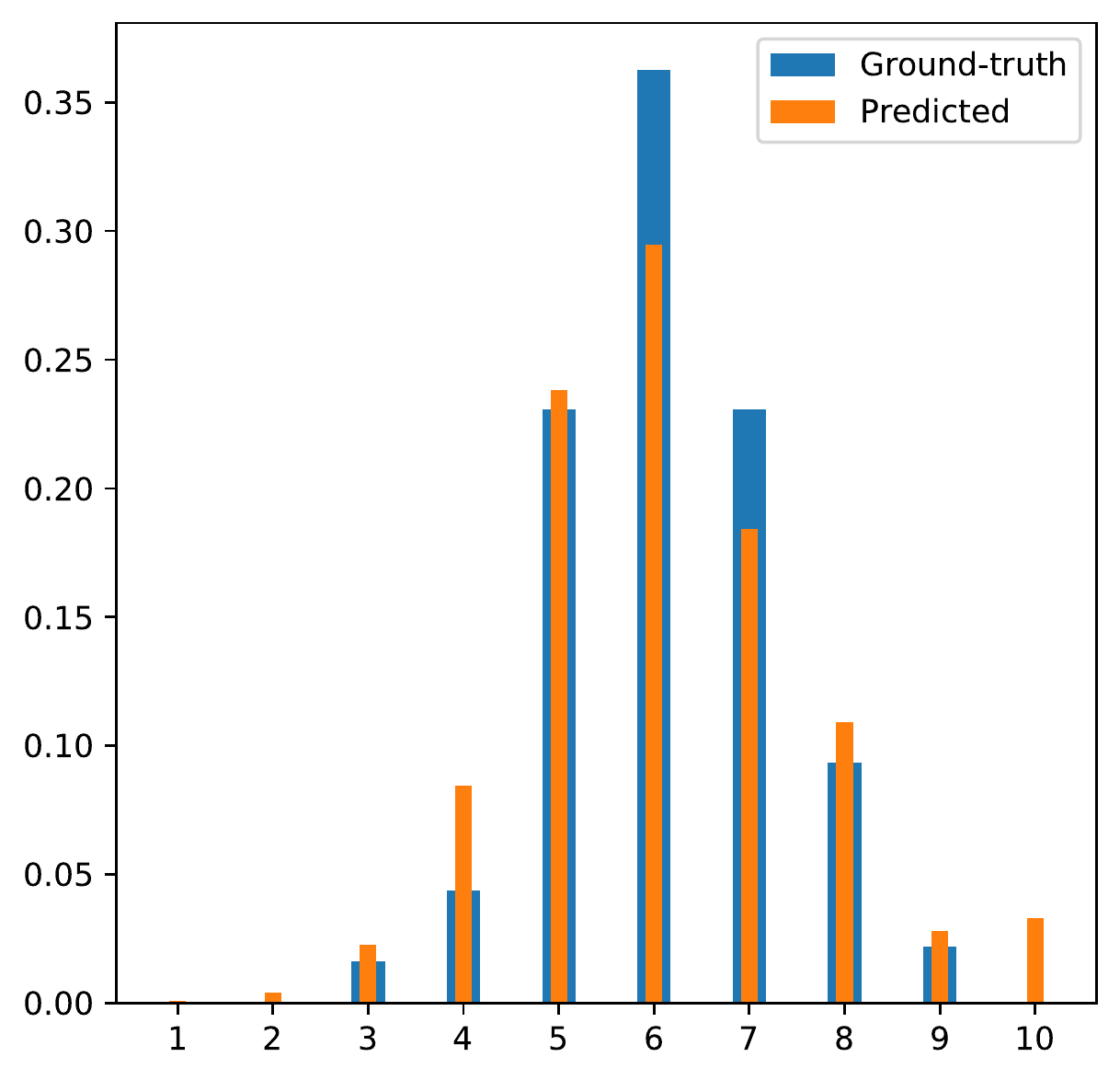} &
        \includegraphics[width=.2\linewidth]{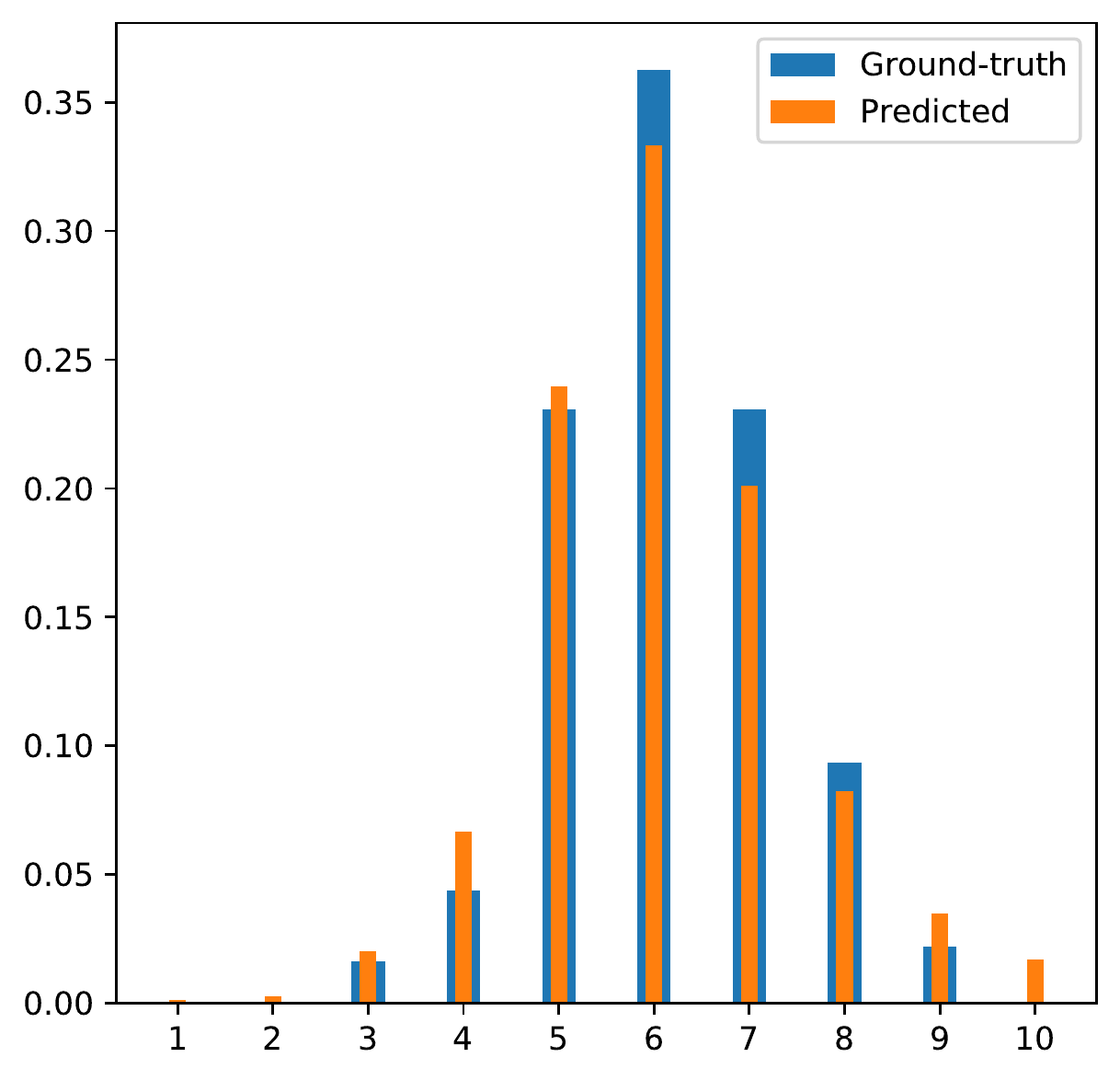} &
        \includegraphics[width=.2\linewidth]{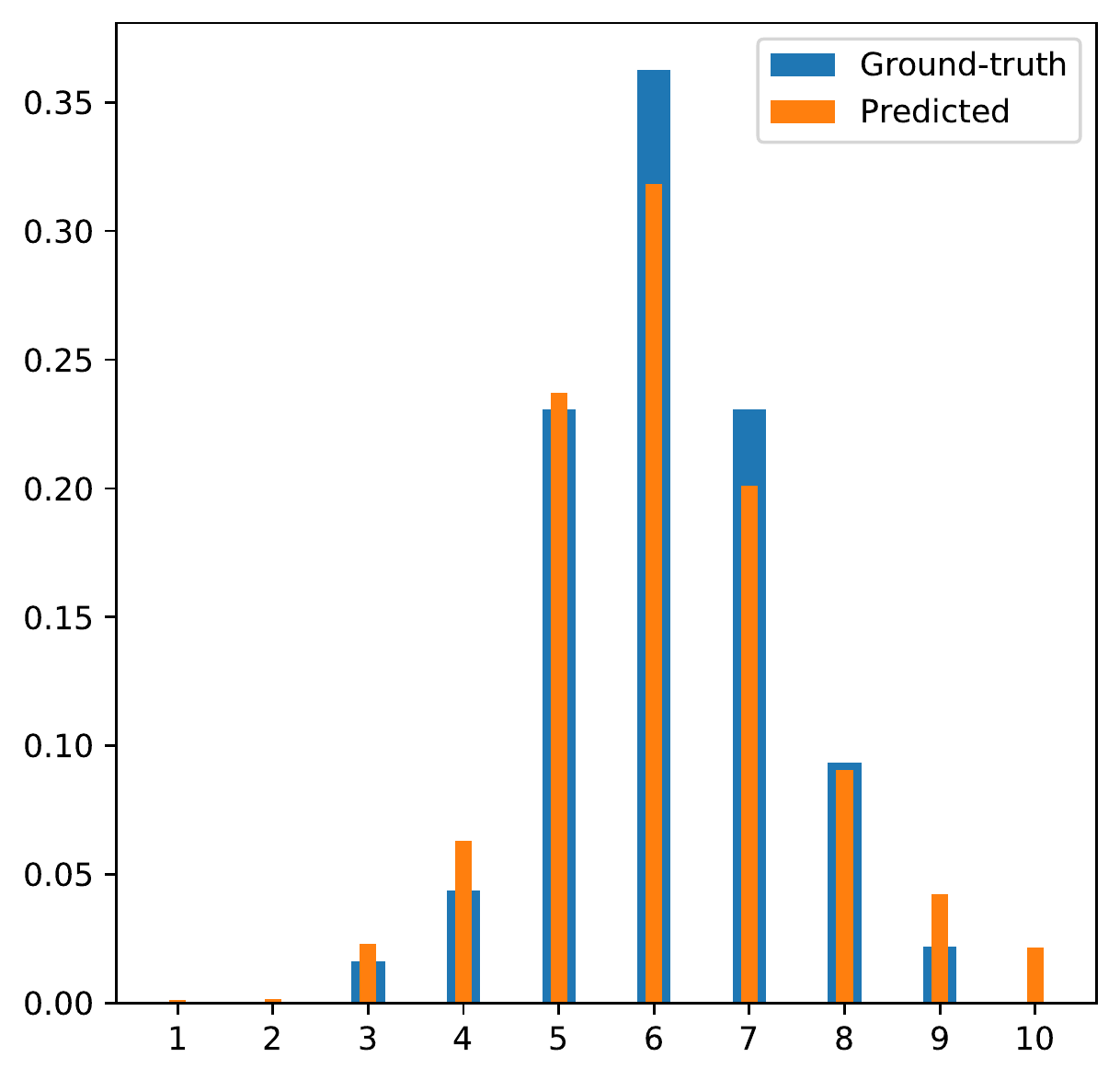} &
        \includegraphics[width=.2\linewidth]{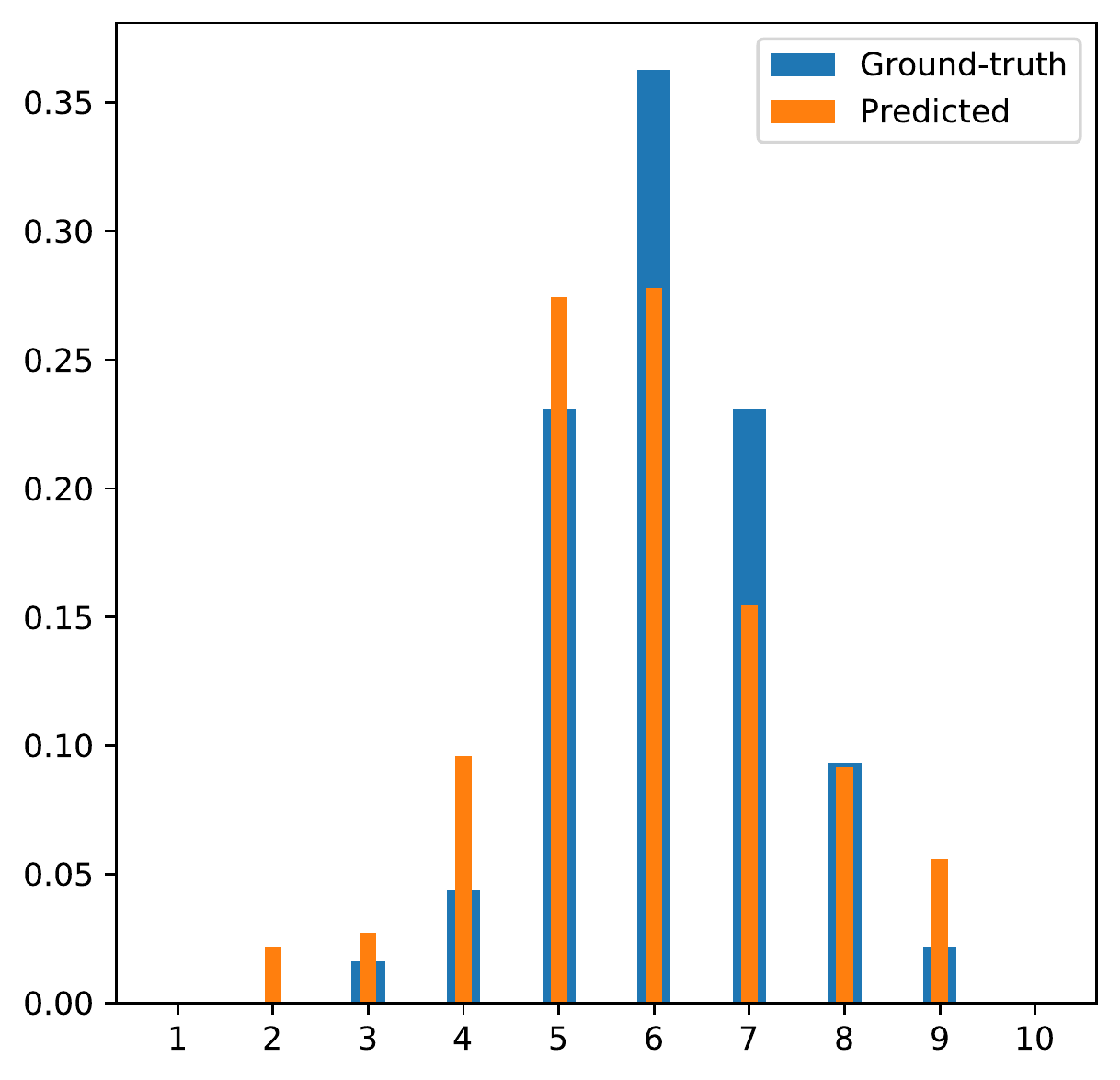} & \includegraphics[width=.2\linewidth]{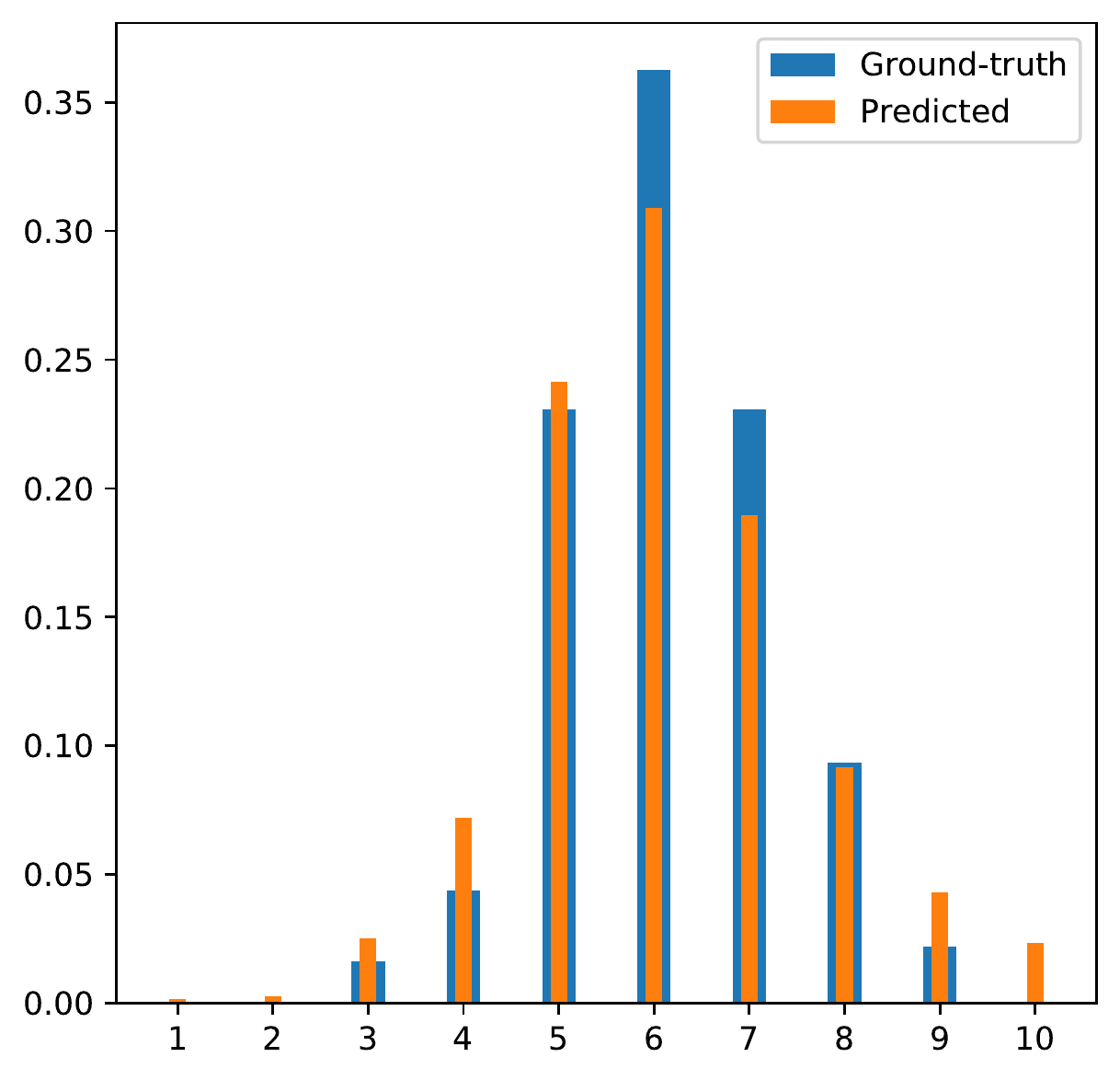} \\
         & (6.1154) 5.8887 & (6.1154) 6.1218 & (6.1154) 6.0864 & (6.1154) 6.1508 & (6.1154) 5.8684 & (6.1154) 6.1154
    \end{tabular}}
    \caption{Qualitative comparison of different versions of the proposed method on three AVA test images. The sampled images are paired with plots of their ground-truth and predicted score distributions. The aesthetic and ground-truth score (in brackets) are also reported under each plot.}
    \label{fig:ablation-qualitative-comp}
\end{figure*}
\subsubsection{Impact of the AttributeNet characterization on the aesthetic assessment}
\label{sec:ablation-data}
Since the aesthetic assessment process is guided by the characterized attributes, the effectiveness of the proposed method also depends on the goodness of the representation obtained from AttributeNet and consequently on the concepts on which it is trained. In this section we comprehensively measure the effect on aesthetic evaluation determined by the data used to train the AttributeNet.

The first set of experiments consists of training AttributeNet to discriminate only the style or composition of the image. In the first experiment, we run the first training of our method for aesthetic-related attributes recognition to discriminate the 20 photographic styles of the FlickrStyle dataset. The training for aesthetic assessment, which affects only the HyperNet, exploits the weights learned on FlickrStyle of the AttributeNet. The results for this solution are reported in the row ``AttrNet (only style) + HyperNet + AesthNet'' of Table \ref{tab:ablation}. Differently from the previous experiment, in the second experiment the training of the AttributeNet learns how to estimate the compositional rules of the KU-PCP. In the row ``AttrNet (only comp.) + HyperNet + AesthNet'' of Table \ref{tab:ablation}, we show the results for this experiment. We can see how the results obtained for both experiments are worse than those obtained by training AttributeNet for simultaneous estimation of style and composition. Furthermore, the version in which the AttributeNet is trained for discriminating the image composition works better than the one in which the photographic styles are estimated. This latter result indicates that somehow composition has a greater importance than styles in estimating image aesthetics.

In another experiment, we evaluate the effectiveness of using only AVA's photographic styles to train the AttributeNet. This experiment serves to quantitatively evaluate the outcome due to the use of additional data to characterize the aesthetic-related attributes. Of the 250,000 images in the AVA dataset, we use the 14,079 images labeled with one of the 14 photographic styles to training AttributeNet (see Table \ref{tab:attributes-list} for details on what those styles are). The training for aesthetic assessment is then carried out by exploiting the AttributeNet trained on AVA styles and without the use of additional data. As can be seen from the results in the row ``AttrNet (on AVA) + HyperNet + AesthNet'' of the Table \ref{tab:ablation}, the performance obtained in this experiment is significantly lower than that of our final solution. The reason for this drop is due to the fact that AVA provides labels for the photographic style and no attributes regarding the composition. We have seen in previous experiments that the composition helps the aesthetic estimation with respect to the photographic style. Certainly the main cause of this poor result is that a few data is available for AttributeNet training as few AVA images are annotated with the style.

A qualitative comparison of the variants of the proposed method evaluated in the ablation study is shown in Figure \ref{fig:ablation-qualitative-comp}.
\subsubsection{AttributeNet hyperparameter tuning}
\label{sec:attrnet-hyper}
\begin{figure}
    \centering
    \includegraphics[width=.8\columnwidth]{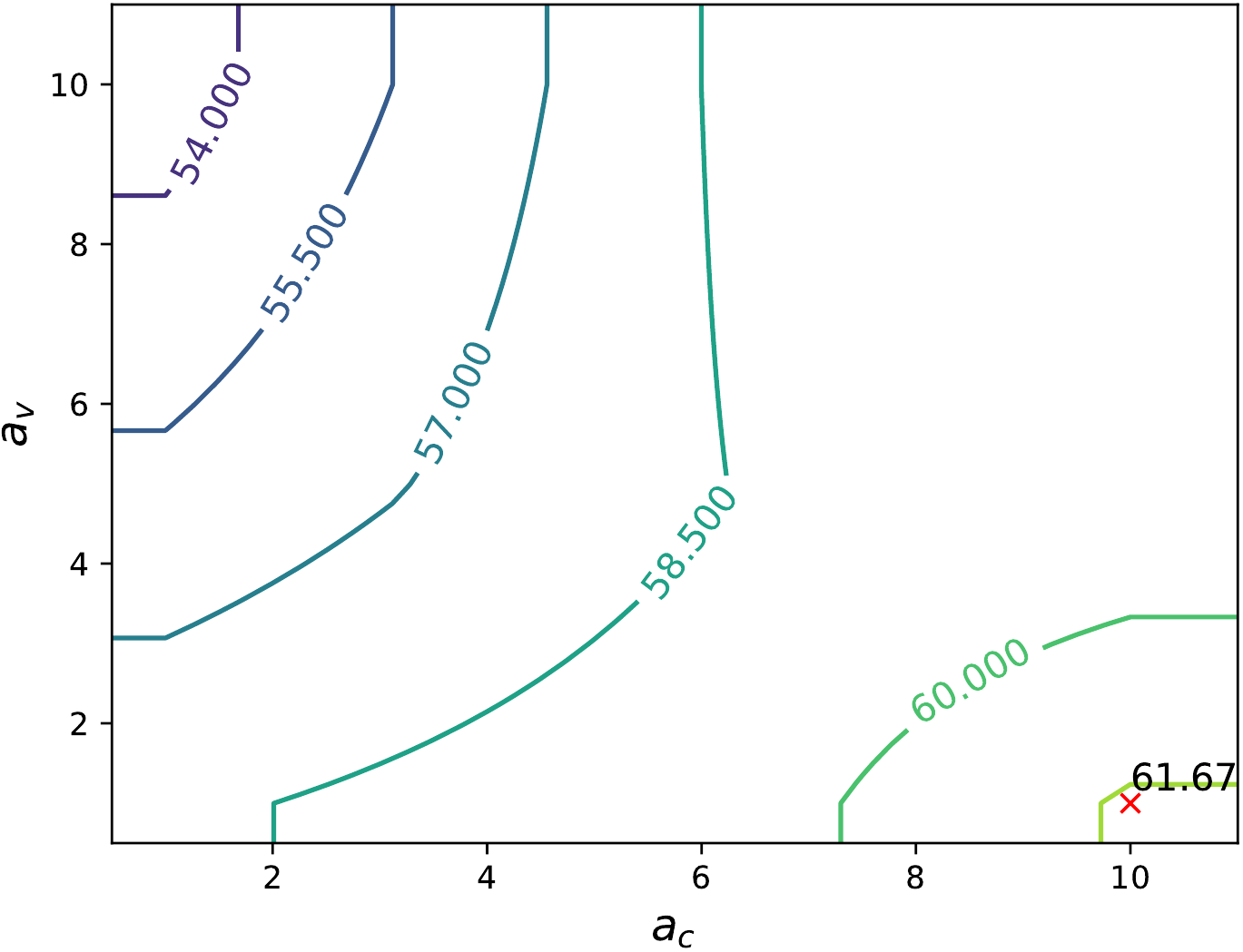}
    \caption{Average accuracy for the composition and style recognition tasks as the values assumed by the parameters $a_c$ and $a_v$ defined in the Equation \ref{eq:train-attrnet} vary. The red cross highlights the best configuration that achieves the best accuracy.}
    \label{fig:ablation-attributenet-training}
\end{figure}

In this subsection we study the effect of the two parameters $a_v$ and $a_c$ defined in Equation \ref{eq:train-attrnet} for AttributeNet training. Since two different datasets are used for training, the choice of these parameters can affect the stability of the training. To this end, the training procedure presented in Section \ref{sec:train-proc} for aesthetics-related attributes recognition is carried out by varying the value of the previous parameters. Figure \ref{fig:ablation-attributenet-training} shows the average accuracy for the two tasks by varying the parameter values. It is possible to notice how as the value of $a_v$ decreases and the value of $a_c$ increases, the accuracy on the two tasks increases until it reaches the highest value of 61.67\% for $a_v=1$ and $a_c=10$, respectively.
\subsection{Visualization of predicted weights}
To assess the influence of the style of the images over the aesthetics prediction, we extract the weights $\hat{\theta}_t$ generated by the HyperNet from several images of the AVA test set.

We annotate all 12,625 test images of the FlickrStyle dataset with the proposed method. We randomly select 200 images for each of the 20 style categories and store the weights $\hat{\mathbf{W}}_5$ of the last AestheticNet linear layer generated by the HyperNet. We then reduce the size of the predicted weights with t-distributed Stochastic Neighbor Embedding (t-SNE) \cite{van2008visualizing} and plot them in a 2D space for visualization.

\begin{figure*}
  \centering
  \includegraphics[width=\textwidth]{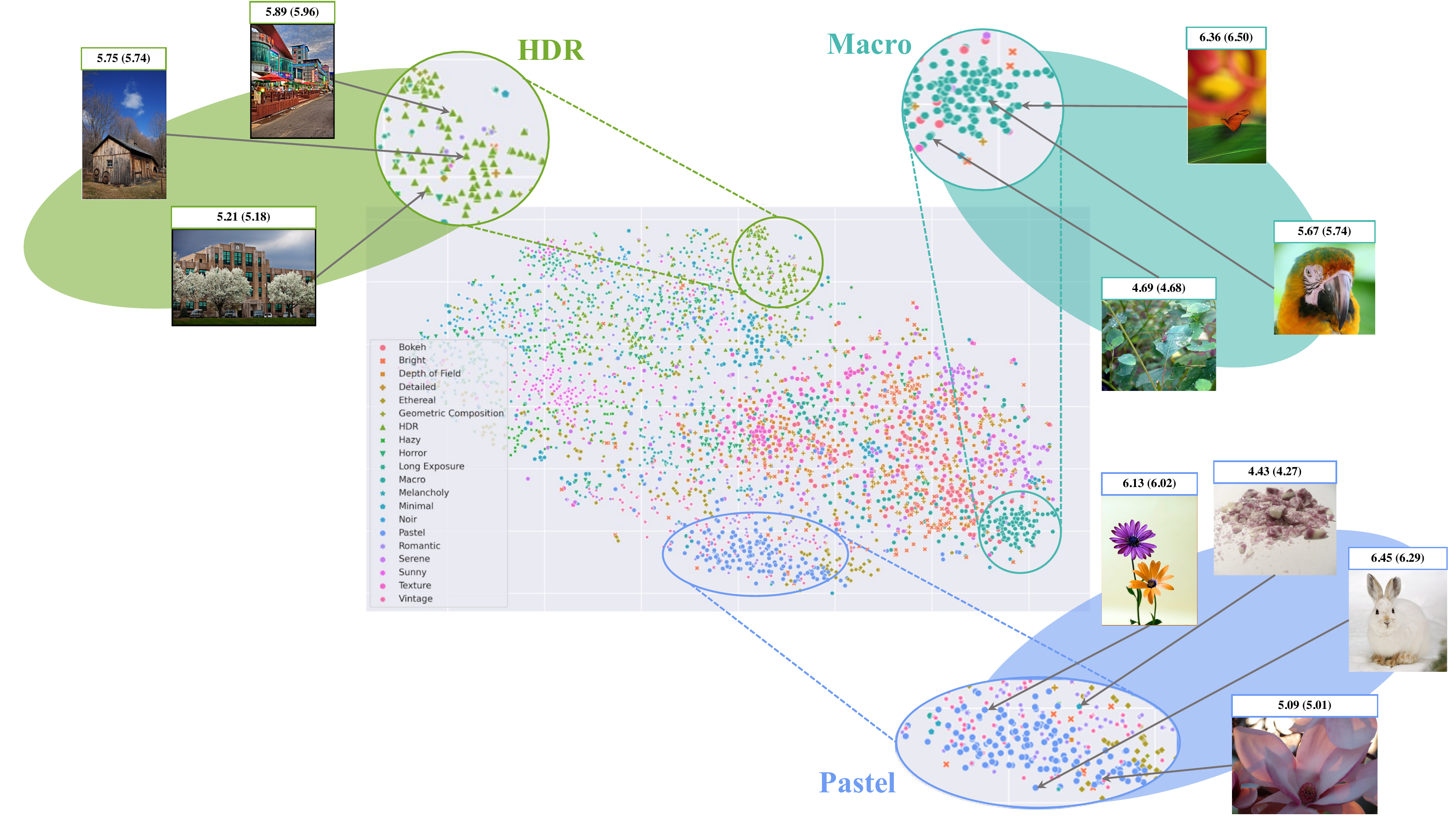}
  \caption{The predicted weights for several images of the AVA test set are plotted in the 2D space after the t-SNE transformation. This figure shows the weights extracted from the last layer of the target network, the weights of the other layers also show a similar distribution. For each of the depicted images, we report the predicted aesthetic score (ground-truth is in brackets).}
  \label{fig:gen-weights}
\end{figure*}

Figure \ref{fig:gen-weights} shows the projection of each selected test image in the two-dimensional space. The weights generated for the other levels of the AestheticNet show a similar distribution. Each point has a different color based on the predicted style category. In the three enlargements, we report a few sample images with the aesthetic scores (predicted and the corresponding ground-truth in brackets).

Several interesting behaviors can be observed from the figure. First, for different images, the generated weights are different. This behavior indicates that our method adopts distinct weights to evaluate the aesthetics of the image in a self-adaptive manner. While for traditional models of automatic aesthetic assessment, the weights are fixed for all input images. Second, images belonging to the same image style (e.g. HDR, pastel, macro) generate weights that are close to each other. This verifies that the training of the AttributeNet is effective. Furthermore, it is noticeable how the information encoded in the embedding produced by the AttributeNet is successfully propagated through the whole model.
%

%
\section{Conclusion}
\label{sec:conclusion}
In this article, we presented a self-adaptive method for image aesthetic assessment. Since the aesthetic evaluation of images is influenced by the content and aesthetic-related attributes, we designed a method that models the aesthetics of the image by explicitly considering semantic content, style, and composition. In particular, the proposed method exploits the side information relating to the aesthetic attributes of an image to build an \emph{ad-hoc} image aesthetics estimator. The parameters of the aesthetic estimator are adaptively generated from a metamodel consisting of an attribute-conditioned hypernetwork. Given an image, the resulting model predicts (i) the style and composition of the image, (ii) the aesthetic score distribution.

Experimental results on three benchmark datasets, namely AADB \cite{kong2016aesthetics}, AVA \cite{murray2012ava}, and Photo.net \cite{datta2006studying} show that the proposed method achieves comparable performance to previous methods for aesthetic quality classification. Instead, it outperforms state-of-the-art methods for both image aesthetic score regression and aesthetic score distribution prediction. Ablation experiments show that aesthetic-attributes, in particular composition rules, allow to obtain an aesthetic evaluation correlating better with human judgments. Furthermore, having attribute-specific aesthetic estimators thanks to the use of the hypernetwork results in better effectiveness of the proposed method with respect to state-of-the-art methods.

Regardless of the obtained results, we believe that the approach integrating attribute-related knowledge in the aesthetic evaluation process represents a step towards a deeper understanding of aesthetic clues. Furthermore, it is highly scalable to new data or aesthetic-related tasks.

\ifCLASSOPTIONcaptionsoff
 \newpage
\fi


\bibliographystyle{IEEEtran}


%

\begin{IEEEbiography}[{\includegraphics[valign=c,width=1in,height=1.25in,clip,keepaspectratio]{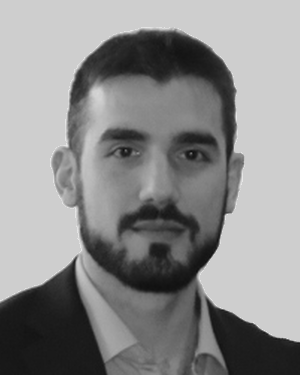}}]{Luigi Celona}
is currently a postdoctoral fellow at DISCo (Department of Informatics, Systems and Communication) of the University of Milano-Bicocca, Italy. In 2018 and 2014, he obtained respectively the PhD and the MSc degree in Computer Science at DISCo. In 2011, he obtained the BSc degree in Computer Science from the University of Messina. His current research interests focus on image analysis and classification, machine learning and face analysis.
\end{IEEEbiography}

\begin{IEEEbiography}[{\includegraphics[valign=c,width=1in,height=1.25in,clip,keepaspectratio]{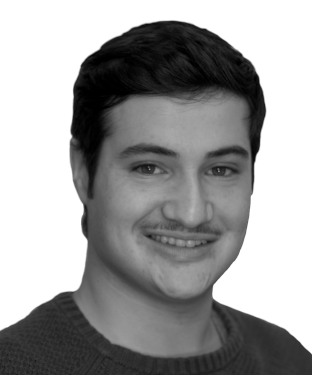}}]{Marco Leonardi}
is a Ph.D. student in Computer Science at DISCo (Department of Informatics, Systems and Communication) of the University of Milano-Bicocca, Italy. In 2018 and 2016, he obtained respectively his Master Degree and Bachelor Degree in Computer Science at DISCo focusing on Image Processing and Computer Vision tasks. The main topics of his current research concern image memorability and machine learning.
\end{IEEEbiography}

\begin{IEEEbiography}[{\includegraphics[valign=c,width=1in,height=1.25in,clip,keepaspectratio]{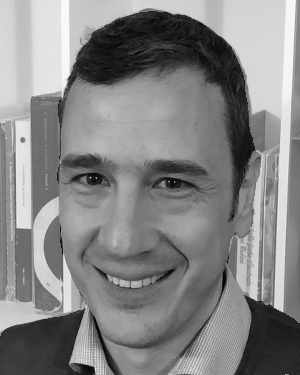}}]{Paolo Napoletano}
is associate professor at the University of Milano-Bicocca (Italy). In 2007, he received a PhD in Information Engineering from the University of Salerno (Italy). In 2003, he received a Master’s degree in Telecommunications Engineering from the University of Naples Federico II. His current research interests focus on signal, image and video analysis and understanding, multimedia information processing and management and machine learning for multi-modal data classification and understanding.
\end{IEEEbiography}

\begin{IEEEbiography}[{\includegraphics[valign=c,width=1in,height=1.25in,clip,keepaspectratio]{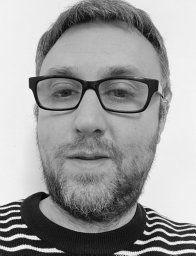}}]{Alessandro Rozza}
is the Chief Scientist of lastminute.com group. In 2011, he received a Doctor of Philosophy degree (PhD) in Computer Science from the Department of Scienze dell’Informazione, Universit\`a degli Studi di Milano. From 2012 to 2014 he was Assistant Professor at Universit\`a degli Studi di Napoli-Parthenope. From 2015 to 2017, he was head of research at Waynaut. His research interests include machine learning and its applications.
\end{IEEEbiography}

\end{document}